\documentclass[twocolumn]{fairmeta}
\usepackage[most]{tcolorbox}

\usepackage{amsmath}
\usepackage{amssymb}

\title{DexGraspFlow: Topology-Aware SE(3) Flow Matching for Cross-Embodiment Dexterous Grasp Generation}

\author[1]{\scalebox{0.94}Bin Zhou}
\author[1,\dagger]{\scalebox{0.94}Boyu Ma}
\author[1]{\scalebox{0.94}Xinyu Zhou}
\author[1]{\scalebox{0.94}Jiahua Dong}
\author[1]{\scalebox{0.94}Chuhao Zhou}
\author[1]{\scalebox{0.94}Yizhou Liu}
\author[2]{\scalebox{0.94}Hao Wan}
\author[1,\dagger]{\scalebox{0.94}Jianfei Yang}

\affiliation[1]{MARS Lab, Nanyang Technological University, Singapore}
\affiliation[2]{ROKAE Robotics}

\contribution[\dagger]{Corresponding authors}

\abstract{
Differences in hand geometry, joint parameterization, and kinematic topology pose challenges for cross-embodiment dexterous grasp generation. However, spatial relations among links alone are insufficient to fully capture the hand's kinematic topology. We propose DexGraspFlow, a conditional link-level $\mathrm{SE}(3)$ flow-matching framework integrating geometric relations and kinematic topology. The framework encodes local object geometry, link geometry, and current poses. A dual-stream DiT models link--link coordination and link--object interaction, while attention biases derived from kinematic-tree distances provide explicit structural information for modeling evolving geometric interactions. The model jointly learns translational and rotational velocity fields, integrates them to generate link poses, and recovers joint configurations through inverse kinematics. Additionally, inference-time palm-orientation guidance enables controllable grasp generation without retraining. Across three robotic hands and 30 object--hand pairs, DexGraspFlow achieves a 95.33\% average simulation success rate, with an average generation latency, including inverse kinematics, approximately 49\% lower than that of a state-of-the-art baseline. Further experiments demonstrate zero-shot adaptability to seven finger-removal variants and palm-orientation control. A model trained on demonstrations for the deployed hand achieves a 91\% success rate over 100 real-robot grasping trials, supporting the framework's applicability to physical grasp execution.
}

\correspondence{Jianfei Yang (\email{jianfei.yang@ntu.edu.sg}); Boyu Ma (\email{boyu.ma@ntu.edu.sg})}

\metadata[Project Page]{\url{https://zhoubin818.github.io/DexGraspFlow/}}

\begin{document}

\maketitle

\begin{figure*}[!t]
    \centering
    \includegraphics[width=0.98\textwidth]{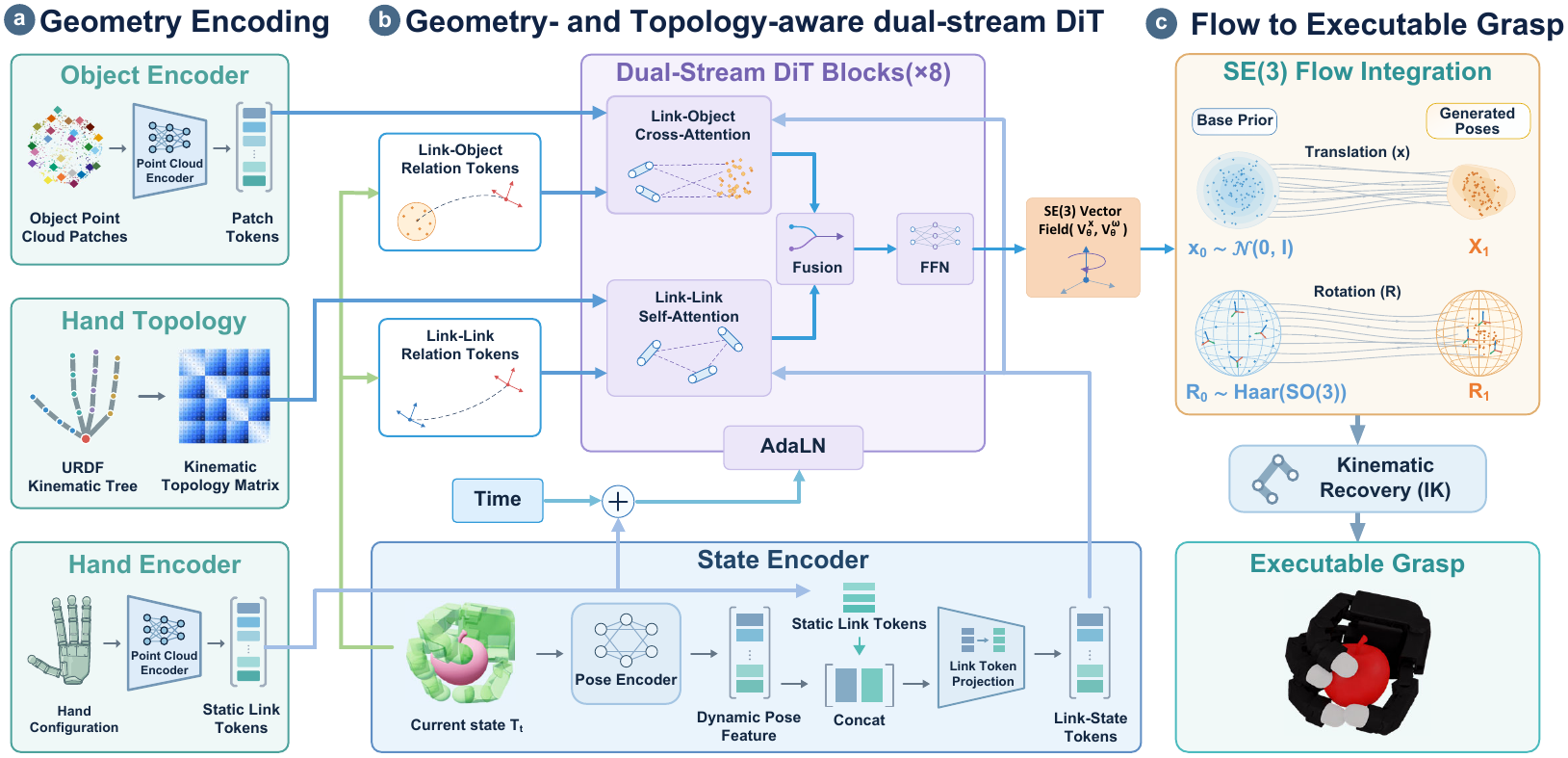}
    \caption{Overview of DexGraspFlow. (a) Geometry and state encoding: Object points and per-link hand point clouds are encoded into object-patch tokens and static link tokens, while the kinematic tree defines the topological distance matrix $\mathbf{D}_{\mathrm{topo}}$. Static link tokens are concatenated with dynamic link-pose features to form link-state tokens. (b) Geometry- and topology-aware dual-stream DiT: Link--object cross-attention and link--link self-attention process state-dependent geometric relation tokens, with kinematic-tree distance biases injected into the link--link stream, to jointly predict translational and rotational velocity fields $v_\theta = (v_\theta^x, v_\theta^\omega)$ with per-link outputs $(\hat{\mathbf{u}}_i^x, \hat{\mathbf{u}}_i^\omega)$. (c) Flow to executable grasp: Continuous velocity fields are numerically integrated from prior distributions to generate target link poses, which are mapped into executable joint configurations via batched inverse kinematics.}
    \label{fig:method_overview}
\end{figure*}

\section{Introduction}
Dexterous hands are robotic end-effectors that interact directly with the physical world, performing the \textit{last centimeter} of robotic manipulation. Through coordinated finger motions, they enable versatile grasping and precise manipulation, supporting the practical deployment of embodied intelligence \citep{dong2026learning,wang2026lifelong}. Reliable dexterous grasping requires jointly accounting for object geometry, hand--object interactions, and kinematic constraints \citep{dexgraspnet}. Differences in link geometry and kinematic topology across hands further complicate cross-embodiment grasp generation. Embodiment-specific joint representations differ in dimensionality and physical meaning, making it difficult to learn a unified model directly from grasp demonstrations collected across different hands \citep{gendexgrasp}. Training a separate model for each hand also increases training and deployment costs. These challenges motivate cross-embodiment dexterous grasp generation methods that share grasping knowledge while accommodating differences in hand morphology and kinematics, facilitating the reuse of grasping capabilities across robotic platforms.

Existing approaches address cross-embodiment grasp synthesis through optimization or shared interaction representations. Optimization-based methods require computationally intensive iterative search for each grasp instance~\citep{dfc}, while contact- and distance-based formulations necessitate auxiliary matching or complex reconstruction to recover executable configurations~\citep{gendexgrasp,dro}. Recently, $\mathcal{T}(\mathcal{R},\mathcal{O})$ Grasp (TRO-Grasp) introduced a transformation-graph diffusion model via object-centric link poses and relative spatial relations without requiring a shared joint parameterization~\citep{tro}. However, relying solely on continuous spatial relations ignores kinematic topology: spatially adjacent links across diverse hands may belong to distant topological branches with fundamentally different kinematic meanings. GraspGraphNet~\citep{graspgraphnet} incorporates kinematic structure through local message passing, but propagating information between distant links along the kinematic tree requires multiple graph hops. Beyond unconditioned synthesis, practical deployment also requires efficient control over grasp approach orientations while maintaining grasp stability across different hand embodiments.

To this end, we propose DexGraspFlow, a cross-embodiment dexterous grasp synthesis framework that unifies continuous $\mathrm{SE}(3)$ flow matching with URDF-derived kinematic topology (Fig.~\ref{fig:method_overview}). To resolve structural ambiguities across heterogeneous hands, DexGraspFlow augments dynamic link states and spatial relations with kinematic-tree shortest-path distances, providing complementary geometric and structural information that explicitly separates spatial arrangement from kinematic connectivity. Grasp synthesis is formulated as link-level flow matching with separate translational reference paths in $\mathbb{R}^3$ and geodesic rotational reference paths on $\mathrm{SO}(3)$~\citep{flow,RFM}. This formulation preserves valid link orientations and supports inference-time palm-orientation guidance through tangent-space velocity corrections without retraining. A geometry- and topology-aware dual-stream DiT~\citep{dit} predicts these coupled velocity fields, followed by batched inverse kinematics~\citep{pyroki} to recover executable joint configurations.

The remainder of this paper is organized as follows. Section~\ref{sec:related_work} reviews related work on cross-embodiment grasp representations and geometric generative models. Section~\ref{sec:framework} presents the grasp representation and geometry- and topology-aware dual-stream DiT architecture of DexGraspFlow. Section~\ref{sec:experiments} evaluates DexGraspFlow through grasp generation on unseen objects, zero-shot generation on finger-removal variants, and palm-orientation-conditioned generation, along with ablation studies and real-robot experiments. Section~\ref{sec:conclusion_limitations} concludes the study and discusses the limitations of the proposed framework and directions for future research.

The main contributions of this work are as follows:
\begin{itemize}
    \item We address the core discrepancy that geometric spatial relations do not reflect kinematic topological connectivity across diverse embodiments, proposing a topology-aware link representation that couples dynamic spatial interactions with kinematic-tree distances to preserve structural coordination under hand topology changes.
    \item We formulate grasp synthesis via link-level $\mathrm{SE}(3)$ flow matching that couples Euclidean translation with Riemannian rotation on $\mathrm{SO}(3)$, and design a geometry- and topology-aware dual-stream DiT, demonstrating superior grasp success with substantially reduced generation latency compared to diffusion baselines in our empirical evaluations.
    \item We propose a training-free directional grasp guidance formulation. By injecting geodesic velocity corrections directly into the tangent space of the continuous flow during inference, DexGraspFlow achieves high-precision, user-specified palm-orientation steering without requiring conditional retraining or auxiliary classifiers.
    \item We conduct extensive empirical evaluations across multi-embodiment simulations, zero-shot generalization to seven finger-removal variants, and 100 real-robot physical trials (91\% success rate), demonstrating the superior adaptability and physical viability of DexGraspFlow.
\end{itemize}

\section{Related work}\label{sec:related_work}
\subsection{Cross-embodiment grasp representations}

GraspIt!~\citep{graspit} supports optimization-based grasp planning, while DFC~\citep{dfc} and DexGraspNet~\citep{dexgraspnet} employ iterative optimization for grasp synthesis and dataset construction, incurring per-instance computational costs. FFHNet~\citep{ffhnet} and UniDexGrasp~\citep{unidexgrasp} represent grasp configurations through global hand poses and joint coordinates, while UniDexGrasp++~\citep{unidexgraspplus} learns grasping policies for a hand-specific state and action space. However, differences in joint dimensionality and semantics complicate sharing these configurations across embodiments, motivating hand--object interaction representations that do not require a common joint parameterization.

GenDexGrasp~\citep{gendexgrasp} uses hand-agnostic contact maps and recovers hand configurations through contact matching and optimization. $\mathcal{D}(\mathcal{R},\mathcal{O})$ Grasp (DRO-Grasp)~\citep{dro} predicts robot--object point-to-point distances using a configuration-invariant pretrained hand encoder, followed by link-pose reconstruction and joint-configuration recovery. Its dense pairwise representation incurs memory overhead, while encoder pretraining adds a separate training stage. TRO-Grasp~\citep{tro} models hands via relative spatial transformations on fully connected link--object and link--link edges. However, spatial link proximity alone is insufficient across diverse embodiments, as spatially adjacent links can belong to disparate topological branches that convey fundamentally different kinematic semantics. Alternatively, GraspGraphNet~\citep{graspgraphnet} encodes articulated hand states using a kinematic-tree MPNN and incorporates object information through cross-attention and dynamic interaction edges. Within its kinematic-tree encoder, information exchange between distant links relies on successive local message-passing steps. DexGraspFlow instead uses link--link self-attention to directly model interactions between all valid link pairs, with kinematic-tree distance biases providing structural information alongside state-dependent geometric relations. This design combines direct long-range link interaction with explicit kinematic proximity encoding.

\subsection{Geometric generative models}

Generative grasp synthesis has been explored through adversarial and diffusion models~\citep{dexgangrasp,dexdiffuser,dexgraspnet2}, drawing on broader denoising and score-based formulations~\citep{ddpm,scoresde}. Flow-based approaches learn continuous transport dynamics in Euclidean and non-Euclidean spaces~\citep{flow,rectified_flow,RFM}. Complementary advances in equivariant geometric modeling~\citep{se3transformer,egnn}, diffusion-based grasp and motion optimization~\citep{se3diffusionfields}, and Transformer generative architectures~\citep{dit} provide foundations for geometry-aware generation.

For cross-embodiment grasping, these formulations differ in their conditioning requirements and the computational path to executable grasps. DRO-Grasp~\citep{dro} combines CVAE latent variables with features of an initial hand configuration and the object to predict grasp distances. Although latent sampling enables diversity, generation remains conditioned on the initial hand state and requires subsequent pose reconstruction. TRO-Grasp~\citep{tro} instead samples link poses from noise through graph diffusion in Lie-algebra coordinates. Its denoiser processes node and edge features through multiple graph layers, with repeated evaluations accumulating sampling costs. DexGraspFlow adopts link-level \(\mathrm{SE}(3)\) flow matching, jointly learning translational and rotational velocities from Euclidean and geodesic reference paths, respectively. Geometric interactions and kinematic topology condition the velocity field within a unified formulation supporting both unconditioned and condition-guided grasp generation. Our experiments demonstrate lower generation latency than DRO-Grasp and TRO-Grasp while maintaining competitive grasp success.

\section{DexGraspFlow framework}\label{sec:framework}
DexGraspFlow generates cross-embodiment dexterous grasps by conditioning continuous $\mathrm{SE}(3)$ flow matching on object geometry, hand morphology, and kinematic topology (Fig.~\ref{fig:method_overview}). First, the object and hand encoders extract object-patch tokens and static link tokens from object and local link point clouds, respectively (Sec.~\ref{sec:geometry_encoding}). Next, grasp generation is formulated as link-level $\mathrm{SE}(3)$ flow matching (Sec.~\ref{sec:link_flow_matching}), where a geometry- and topology-aware dual-stream DiT jointly predicts translational and rotational velocities via parallel link--link and link--object attention (Sec.~\ref{sec:dual_stream_network}). Finally, flow integration can be steered via training-free velocity corrections for palm-orientation guidance, after which batched inverse kinematics resolves the generated link poses into executable joint configurations (Sec.~\ref{sec:guided_sampling_recovery}).

\subsection{Problem formulation and geometry encoding}\label{sec:geometry_encoding}
Given an object point cloud $P^O\in\mathbb{R}^{N\times3}$ and a hand described by its URDF and local link point clouds $\{P_i^H\}_{i=1}^{L_H}$, we formulate grasp generation over the poses of the hand's $L_H$ valid physical links. The point clouds describe object and link geometry, while the URDF specifies the hand's kinematic tree and link connectivity~\citep{fktree}. This link-level representation accommodates different hand embodiments without requiring a shared joint parameterization. Object and link geometry are encoded separately to condition the generative model.

We express link poses in a normalized object-centric coordinate frame. The object point cloud and link translations are centered at the object centroid $\mathbf{c}^O$ and divided by the maximum half-extent $s^O$, while rotations remain unchanged. Each normalized link pose is denoted by $(\mathbf{x}_i,\mathbf{R}_i)\in\mathbb{R}^3\times \mathrm{SO}(3)$~\citep{lie}. For batched processing across hands, states are padded to $L_{\max}$ links:
\begin{equation}
    \mathcal{T}
    =
    \{(\mathbf{x}_i,\mathbf{R}_i)\}_{i=1}^{L_{\max}}
    \in
    \left(\mathbb{R}^3\times \mathrm{SO}(3)\right)^{L_{\max}}.
\end{equation}

The validity mask $\mathbf{m}^H=(m_i)_{i=1}^{L_{\max}}$ assigns $m_i=1$ to valid links and $m_i=0$ to padded entries.

\textbf{Object encoder.}
From the normalized object point cloud $\hat{P}^O$, we select $N^O$ patch centers using farthest point sampling (FPS)~\citep{pointnet} and group their $K_{\mathrm{nn}}$ nearest neighbors~\citep{knn}. Each patch is centered at its centroid $\mathbf{c}_j^O$ and normalized by its radius $r_j^O$. A shared point-cloud encoder extracts a local geometric feature $\mathbf{f}_j^O\in\mathbb{R}^{d_f^O}$, which is concatenated with patch location and scale information:
\begin{equation}
\begin{aligned}
    \mathbf{e}_j^O
    &=
    \operatorname{concat}
    \left(\mathbf{f}_j^O,\mathbf{c}_j^O,\rho_j^O,r_j^O,s^O\right),\\
    \rho_j^O
    &=
    \frac{r_j^O}{\frac{1}{N^O}\sum_{k=1}^{N^O}r_k^O}.
\end{aligned}
\end{equation}

The patch radius $\rho_j^O$ is defined relative to the mean patch radius, while $s^O$ retains the original object scale. The resulting representation is $\mathbf{E}^O=\{\mathbf{e}_j^O\}_{j=1}^{N^O}\in\mathbb{R}^{N^O\times d_e^O}$, with $d_e^O=d_f^O+6$.

\textbf{Hand encoder.}
For each valid link, FPS samples points from its URDF-associated mesh in the link's local coordinate frame. The point cloud is centered at its local centroid $\mathbf{c}_i^H$ and normalized by its radius $r_i^H$. A shared residual point-cloud encoder produces a geometric feature $\mathbf{f}_i^H\in\mathbb{R}^{d_f^H}$. Let $\mathbf{p}_i^{\mathrm{ref}}$ and $\mathbf{R}_i^{\mathrm{ref}}$ denote the link's position and orientation in the palm frame at a fixed hand-specific reference configuration. We concatenate these quantities into $\mathbf{z}_i^H$ and encode the static link token $\mathbf{e}_i^H\in\mathbb{R}^{d_s^H}$:
\begin{equation}
\begin{aligned}
    \mathbf{z}_i^H
    &= \operatorname{concat}\left(
    \mathbf{f}_i^H,\mathbf{c}_i^H,r_i^H,
    \mathbf{p}_i^{\mathrm{ref}},
    \operatorname{Rot6D}(\mathbf{R}_i^{\mathrm{ref}})
    \right),\\
    \mathbf{e}_i^H
    &= \operatorname{MLP}_{\mathrm{link}}\left(
    \operatorname{Proj}_{d_f^H+13\rightarrow d_s^H}
    (\mathbf{z}_i^H)
    \right).
\end{aligned}
\end{equation}
The operator $\operatorname{Rot6D}$ denotes the continuous six-dimensional rotation representation~\citep{6drotation}.

We use $\operatorname{Proj}$ to denote learnable projections, with subscripts indicating their roles or input and output dimensions. The hand representation $\mathbf{E}^H=\{\mathbf{e}_i^H\}_{i=1}^{L_H}$ is padded to $\tilde{\mathbf{E}}^H\in\mathbb{R}^{L_{\max}\times d_s^H}$ using the validity mask defined above, with $\tilde{\mathbf{e}}_i^H$ denoting its $i$-th token. These tokens encode local link geometry and palm-relative reference poses. Their integration with the current link poses is described in Sec.~\ref{sec:dual_stream_network}.

\begin{figure*}[!t]
    \centering
    \includegraphics[width=\textwidth]{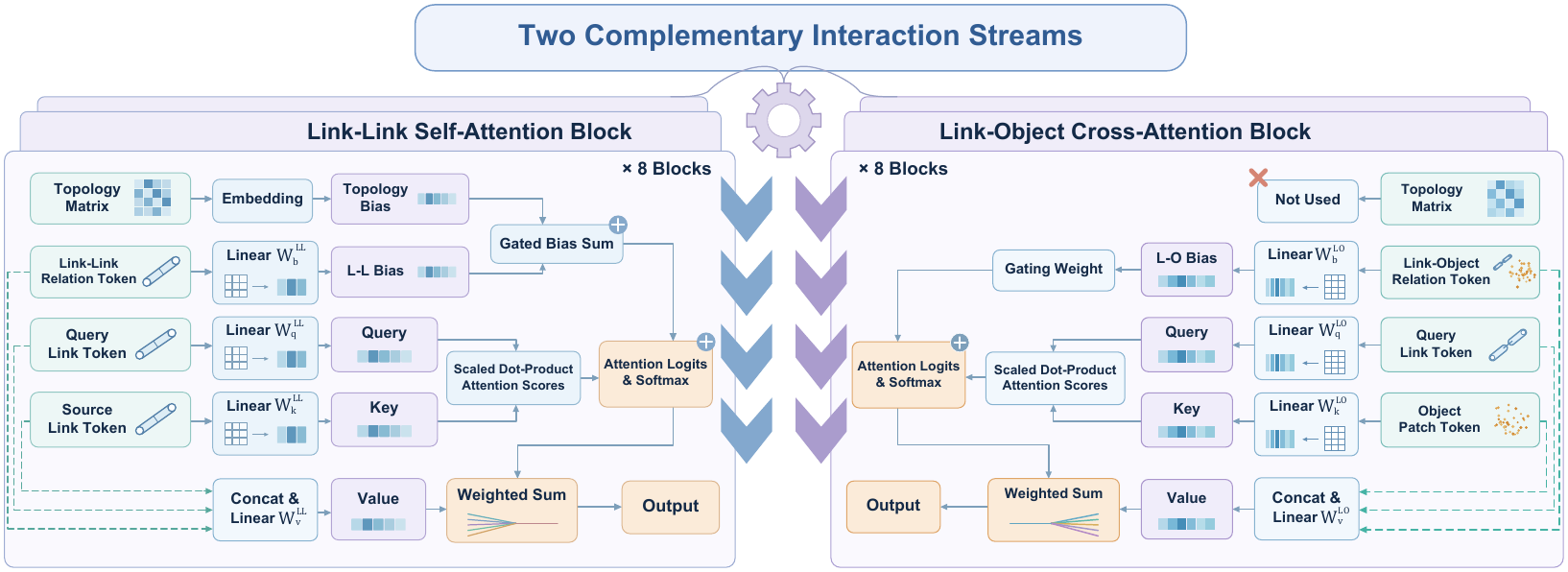}
    \caption{Geometry-aware dual-stream attention in the DiT. The left branch performs link--link self-attention with physical-relation and topology biases. The right branch performs link--object cross-attention from link queries to object patch tokens with a gated physical-relation bias; topology bias is not used in this branch. In both branches, relation-aware pairwise features are projected to value messages before weighted sum.}
    \label{fig:dual_stream_attention}
\end{figure*}

\subsection{Link-level SE(3) flow matching}\label{sec:link_flow_matching}
Conditioned on the object representation $\mathbf{E}^O$ and hand representation $\tilde{\mathbf{E}}^H$, flow matching learns a time-dependent vector field that transports a source distribution $p_0$ to a conditional grasp-data distribution~\citep{flow,rectified_flow,RFM}. Let $p_{\mathrm{data}}(\mathcal{T}\mid\mathbf{E}^O,\tilde{\mathbf{E}}^H)$ denote the empirical distribution of link poses obtained from grasp demonstrations through forward kinematics. During training, we sample a target state $\mathcal{T}_1$ from this distribution and a source state $\mathcal{T}_0\sim p_0$, with standard Gaussian translations and Haar-uniform rotations for each valid link.

\textbf{Reference paths and velocities.}
For a source--target pair, the intermediate state at $t\in[0,1]$ follows a linear translational path and a geodesic rotational path:
\begin{equation}
\begin{aligned}
    \mathbf{x}_{t,i}
    &=(1-t)\mathbf{x}_{0,i}+t\mathbf{x}_{1,i},\\
    \mathbf{R}_{t,i}
    &=\mathbf{R}_{0,i}\operatorname{Exp}
    \left(t\operatorname{Log}(\mathbf{R}_{0,i}^{\top}\mathbf{R}_{1,i})\right).
\end{aligned}
\end{equation}

Here, $\operatorname{Exp}$ and $\operatorname{Log}$ denote the exponential and logarithm maps on $\mathrm{SO}(3)$, expressed using three-dimensional rotation vectors. The corresponding reference velocities are
\begin{equation}
    \mathbf{u}_i^x=\mathbf{x}_{1,i}-\mathbf{x}_{0,i},
    \qquad
    \mathbf{u}_i^\omega=
    \operatorname{Log}(\mathbf{R}_{0,i}^{\top}\mathbf{R}_{1,i}).
\end{equation}

Translational and rotational velocities are expressed in normalized coordinate units and radians per unit flow time, respectively. The translational reference path lies in $\mathbb{R}^3$, while the rotational reference path remains on $\mathrm{SO}(3)$. This preserves valid link orientations for relative-pose computation, while the tangent-space rotational velocity provides a natural interface for inference-time orientation correction.

\textbf{Training objective.}
The network $v_\theta$, with learnable parameters $\theta$, jointly predicts the translational and rotational velocity components. For brevity, the fixed hand-topology condition and validity mask are omitted from the conditioning arguments:
\begin{equation}
    v_\theta
    \left(\mathcal{T}_t,t\mid\mathbf{E}^O,\tilde{\mathbf{E}}^H\right)
    =\{\hat{\mathbf{u}}_i\}_{i=1}^{L_{\max}}.
\end{equation}
where $\hat{\mathbf{u}}_i=[\hat{\mathbf{u}}_i^x,\hat{\mathbf{u}}_i^\omega]\in\mathbb{R}^6$ concatenates the predicted translational and rotational velocities for link $i$. For sampled endpoints and $t\sim\mathcal{U}(0,1)$, we supervise the prediction using the masked flow-matching loss:
\begin{equation}
\begin{aligned}
    \mathcal{L}_{\mathrm{FM}}
    =
    \frac{1}{\sum_{i=1}^{L_{\max}}m_i}
    \sum_{i=1}^{L_{\max}}m_i
    \Bigl(
    &\lambda_x\|\hat{\mathbf{u}}_i^x-\mathbf{u}_i^x\|_2^2\\
    +{}&\lambda_\omega
    \|\hat{\mathbf{u}}_i^\omega-\mathbf{u}_i^\omega\|_2^2
    \Bigr).
\end{aligned}
\end{equation}

The coefficients $\lambda_x$ and $\lambda_\omega$ balance the translational and rotational losses, while $m_i$ excludes padded links from supervision. The vector field is parameterized by the geometry- and topology-aware dual-stream network described in Sec.~\ref{sec:dual_stream_network}.

\textbf{Sampling.}\label{sec:unconditioned_inference}
At inference, we adjust the source initialization through temperature-controlled sampling, conditioned only on the object and hand. For each valid link, we draw $\mathbf{x}_{0,i}\sim\mathcal{N}(\mathbf{0},\tau_x^2\mathbf{I})$ and a Haar-uniform rotation $\widetilde{\mathbf{R}}_{0,i}$, and construct
\begin{equation}
    \mathbf{R}_{0,i}
    =
    \mathbf{R}_i^{\mathrm{init}}
    \operatorname{Exp}
    \left(
    \tau_R\operatorname{Log}
    \left((\mathbf{R}_i^{\mathrm{init}})^\top
    \widetilde{\mathbf{R}}_{0,i}\right)
    \right).
\end{equation}

Here, $\mathbf{R}_i^{\mathrm{init}}$ is obtained by forward kinematics from a randomly sampled initial hand configuration and serves as the center of rotational sampling. The temperatures $\tau_x$ and $\tau_R$ control the translational and rotational initialization spread; $\tau_R=1$ recovers Haar-uniform rotations.

Starting from $\mathcal{T}_0$, we integrate the learned field over $S$ steps with $\Delta t=1/S$. Writing $\mathbf{x}_{k,i}=\mathbf{x}_{t_k,i}$ and $\mathbf{R}_{k,i}=\mathbf{R}_{t_k,i}$ at $t_k=k\Delta t$, the updates are
\begin{equation}
\begin{aligned}
    \mathbf{x}_{k+1,i}
    &=\mathbf{x}_{k,i}+\Delta t\,\hat{\mathbf{u}}_{k,i}^x,\\
    \mathbf{R}_{k+1,i}
    &=\mathbf{R}_{k,i}
    \operatorname{Exp}\left(\Delta t\,\hat{\mathbf{u}}_{k,i}^\omega\right).
\end{aligned}
\end{equation}
where $\hat{\mathbf{u}}_{k,i}$ is the network prediction at $(\mathcal{T}_{t_k},t_k)$. The rotational velocity is expressed in the local tangent coordinates of link $i$, and the exponential update preserves $\mathrm{SO}(3)$. Padded links remain inactive throughout integration. After integration, the generated link translations are de-normalized, and the resulting target poses are converted into executable joint configurations through the kinematic recovery procedure in Sec.~\ref{sec:guided_sampling_recovery}.

\subsection{Geometry- and topology-aware dual-stream network}\label{sec:dual_stream_network}
We parameterize the conditional vector field with a geometry- and topology-aware dual-stream DiT~\citep{dit}. Link--link (LL) self-attention models interactions among hand links, while link--object (LO) cross-attention incorporates object geometry into the evolving link states. Both streams use geometric relation tokens to construct attention biases and value messages, with an additional topology bias in the LL stream. The attention mechanisms are illustrated in Fig.~\ref{fig:dual_stream_attention}.

\textbf{Link-state tokens.}
The object representations defined in Sec.~\ref{sec:geometry_encoding} are projected to the hidden dimension $d_h$, yielding object tokens $\mathbf{o}_j=\operatorname{Proj}_O(\mathbf{e}_j^O)\in\mathbb{R}^{d_h}$. For each hand link, we encode its current pose and combine it with its static link token:
\begin{equation}
\begin{aligned}
\mathbf{d}_{t,i}^H
&=\operatorname{MLP}_{\mathrm{dyn}}
\left(\operatorname{concat}(\mathbf{x}_{t,i},\mathbf{r}_{t,i}^{6D})\right),\\
\mathbf{h}_{t,i}^{(0)}
&=\operatorname{Proj}_H
\left(\operatorname{concat}(\tilde{\mathbf{e}}_i^H,\mathbf{d}_{t,i}^H)\right).
\end{aligned}
\end{equation}

The current rotation is encoded as $\mathbf{r}_{t,i}^{6D}=\operatorname{Rot6D}(\mathbf{R}_{t,i})$. The dynamic encoder consists of a linear layer, layer normalization, and GELU activation, producing $\mathbf{d}_{t,i}^H\in\mathbb{R}^{d_d^H}$. The dynamic pose-feature dimension is denoted by $d_d^H$. The resulting link-state tokens are $\mathbf{H}_t^{(0)}=\{\mathbf{h}_{t,i}^{(0)}\}_{i=1}^{L_{\max}}$, with $\mathbf{h}_{t,i}^{(0)}\in\mathbb{R}^{d_h}$. They combine static hand information with the current link poses and provide the input to the DiT stack.

\textbf{Time and morphology conditioning.}
In addition to their role in link-state tokens, static hand tokens provide a global morphology condition. We combine their masked mean with a Gaussian Fourier time embedding~\citep{fourier}:
\begin{equation}
\begin{aligned}
\gamma(t)
&=\operatorname{concat}\left(
\sin(2\pi t\mathbf{W}^{\top}),
\cos(2\pi t\mathbf{W}^{\top})\right),\\
\mathbf{c}
&=\operatorname{MLP}_t(\gamma(t))
+\operatorname{Proj}_G
\left(\frac{\sum_i m_i\tilde{\mathbf{e}}_i^H}{\sum_i m_i}\right).
\end{aligned}
\end{equation}

The entries of $\mathbf{W}\in\mathbb{R}^{(d_h/2)\times1}$ are sampled from $\mathcal{N}(0,\sigma^2)$ at initialization and remain fixed, with $\sigma=16$. The time MLP consists of two linear layers with a SiLU activation between them. The resulting condition $\mathbf{c}\in\mathbb{R}^{d_h}$ generates block-specific AdaLN shifts, scales, and residual gates, together with per-head geometric-bias gates. Dynamic pose features enter the link-state pathway, whereas the global hand condition is derived from the static link tokens.

\textbf{Geometric relations.}
The current poses also determine the relations used by attention. The index $i$ denotes the query link, while $j$ denotes a source link for LL relations or an object patch for LO relations:
\begin{equation}
\begin{aligned}
\mathbf{r}_{ij}^{LO}
&=\operatorname{concat}
\left(\mathbf{x}_{t,i}-\mathbf{c}_j^O,\mathbf{r}_{t,i}^{6D}\right),\\
\mathbf{r}_{ij}^{LL}
&=\operatorname{concat}\Bigl(
\mathbf{R}_{t,i}^{\top}(\mathbf{x}_{t,j}-\mathbf{x}_{t,i}),  \operatorname{Rot6D}(\mathbf{R}_{t,i}^{\top}\mathbf{R}_{t,j})
\Bigr).
\end{aligned}
\end{equation}

The LL relation represents the source link pose relative to the query link, whereas the LO relation combines the object-frame link--patch displacement with the query link orientation. Separate linear projections $\phi^s$ map these nine-dimensional features to relation tokens $\mathbf{s}_{ij}^s=\phi^s(\mathbf{r}_{ij}^s)\in\mathbb{R}^{d_r}$ for stream $s\in\{LL,LO\}$. Thus, dynamic poses influence the network through both link-state tokens and geometric relations. Relation tokens are recomputed at each vector-field evaluation and reused unchanged across the DiT blocks; each block applies its own projections to incorporate them into attention biases and value messages.

\textbf{Topology encoding.}
Kinematic structure is supplied separately through a topology-bias pathway. We construct an undirected physical-link graph $\mathcal{G}^H=(\mathcal{V}^H,\mathcal{A}^H)$ from the URDF by retaining valid physical links and contracting auxiliary nodes. The sets $\mathcal{V}^H$ and $\mathcal{A}^H$ denote the valid physical links and their undirected kinematic connections, respectively. Structural proximity is represented by shortest-path distances $d_{ij}^{\mathrm{topo}}=\operatorname{dist}_{\mathcal{G}^H}(i,j)$, collected into a symmetric, zero-diagonal topological distance matrix $\mathbf{D}_{\mathrm{topo}}=[d_{ij}^{\mathrm{topo}}]_{i,j=1}^{L_H}\in\mathbb{N}_0^{L_H\times L_H}$.

\begin{figure}[!t]
    \centering
    \includegraphics[width=\columnwidth]{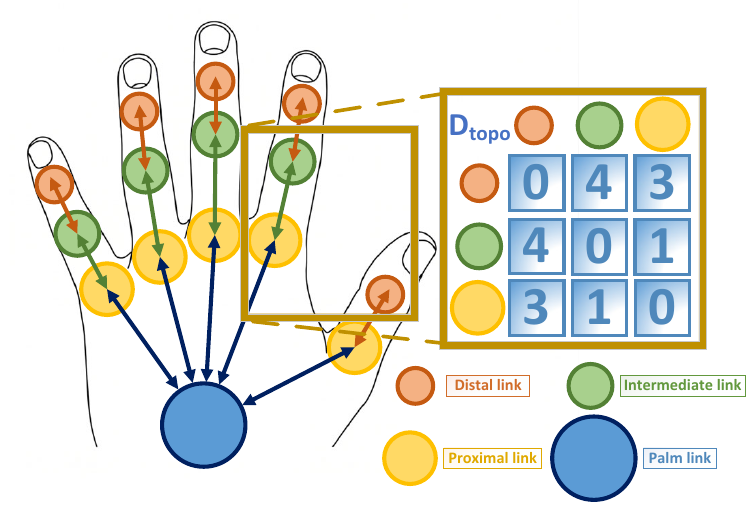}
    \caption{Illustration of topology-distance construction from the
    URDF-derived physical-link graph. Each valid hand link is treated as a
    vertex in the undirected graph, and the distance between two links is the
    shortest-path length. The right panel shows an illustrative $3\times3$
    submatrix of $\mathbf{D}_{\mathrm{topo}}$; the colors indicate schematic link
    categories.}
    \label{fig:topology_distance}
\end{figure}

This matrix is computed offline for each hand, as illustrated in Fig.~\ref{fig:topology_distance}. Within each block, distances are clipped at $d_{\mathrm{clip}}$ and mapped to per-head biases by a learnable embedding:
\begin{equation}
\mathbf{b}_{ij}^{\mathrm{topo}}
=\mathbf{E}_{\mathrm{topo}}
\left(\operatorname{clip}(d_{ij}^{\mathrm{topo}},0,d_{\mathrm{clip}})\right)
\in\mathbb{R}^{n_h}.
\end{equation}
where $n_h$ is the number of attention heads and $\mathbf{E}_{\mathrm{topo}}\in\mathbb{R}^{(d_{\mathrm{clip}}+1)\times n_h}$. Invalid pairs use the padding index $d_{\mathrm{clip}}$. The distance matrix is fixed for a given hand, while each block learns its own embedding and per-head topology gate.

\textbf{Dual-stream attention and output.}
To describe a single DiT block at a fixed flow time, we omit the time and block indices and denote the block input by $\mathbf{h}_i$. Bias terms in linear projections are omitted for clarity. Both streams operate on the same input link tokens, using separate AdaLN modulation:
\begin{equation}
    \bar{\mathbf{h}}_i^s = (1 + \operatorname{scale}^s(\mathbf{c})) \odot \operatorname{LN}(\mathbf{h}_i) + \operatorname{shift}^s(\mathbf{c}),
\end{equation}
In this expression, $s \in \{LL, LO\}$ indexes the interaction stream, $\bar{\mathbf{h}}_i^s$ denotes the modulated link token, $\operatorname{LN}$ denotes layer normalization, and $\odot$ denotes element-wise multiplication. The scale and shift are feature-wise vectors generated from $\mathbf{c}$. Queries and keys are constructed as
\begin{equation}
\begin{aligned}
\mathbf{q}_i^{LL}&=\mathbf{W}_Q^{LL}\bar{\mathbf{h}}_i^{LL},
&\mathbf{k}_j^{LL}&=\mathbf{W}_K^{LL}\bar{\mathbf{h}}_j^{LL},\\
\mathbf{q}_i^{LO}&=\mathbf{W}_Q^{LO}\bar{\mathbf{h}}_i^{LO},
&\mathbf{k}_j^{LO}&=\mathbf{W}_K^{LO}\mathbf{o}_j.
\end{aligned}
\end{equation}

Each stream constructs relation-conditioned value messages by linearly projecting the concatenated query, source, and relation features:
\begin{equation}
\begin{aligned}
\mathbf{v}_{ij}^{LL}
&=\mathbf{W}_V^{LL}\operatorname{concat}
\left(\bar{\mathbf{h}}_i^{LL},\bar{\mathbf{h}}_j^{LL},\mathbf{s}_{ij}^{LL}\right),\\
\mathbf{v}_{ij}^{LO}
&=\mathbf{W}_V^{LO}\operatorname{concat}
\left(\bar{\mathbf{h}}_i^{LO},\mathbf{o}_j,\mathbf{s}_{ij}^{LO}\right).
\end{aligned}
\end{equation}

The matrices $\mathbf{W}_Q^s$, $\mathbf{W}_K^s$, and $\mathbf{W}_V^s$ are learnable query, key, and value projections specific to stream $s$. The value-projection input has dimension $2d_h+d_r$. Separate linear projections produce geometric biases $\mathbf{b}_{ij}^{s}=\mathbf{W}_b^s\mathbf{s}_{ij}^{s}\in\mathbb{R}^{n_h}$, with $\mathbf{W}_b^s\in\mathbb{R}^{n_h\times d_r}$. The additional index $a$ identifies an attention head, with $\mathbf{q}_{i,a}^s$, $\mathbf{k}_{j,a}^s$, and $\mathbf{v}_{ij,a}^s$ denoting the $d_a=d_h/n_h$-dimensional head components, while $b_{ij,a}^s$ is the corresponding scalar bias entry for head $a$. The attention logits are
\begin{equation}
\begin{aligned}
\ell_{ij,a}^{LL}
&=\frac{(\mathbf{q}_{i,a}^{LL})^\top\mathbf{k}_{j,a}^{LL}}{\sqrt{d_a}}
+\mu_a^{LL}(\mathbf{c})b_{ij,a}^{LL}+g_a^{\mathrm{topo}}b_{ij,a}^{\mathrm{topo}},\\
\ell_{ij,a}^{LO}
&=\frac{(\mathbf{q}_{i,a}^{LO})^\top\mathbf{k}_{j,a}^{LO}}{\sqrt{d_a}}
+\mu_a^{LO}(\mathbf{c})b_{ij,a}^{LO}.
\end{aligned}
\end{equation}

The geometric-bias gate $\mu_a^s(\mathbf{c})$ is condition-dependent, whereas $g_a^{\mathrm{topo}}$ is a learnable per-head topology gate. In the LL stream, relative link poses contribute to the attention logits through the geometric bias, while the topology bias provides kinematic-tree proximity information. The LO stream combines link-dependent query--key scores with an additive geometric bias and relation-conditioned value messages. Under linear relation projections, query-only terms cancel out under softmax, so the effective LO bias acts directly on object-patch positions modulated by condition $\mathbf{c}$, injecting explicit spatial context into cross-attention matching.

Invalid sources are masked before softmax, and LL self-attention excludes the query link itself. For each stream,
\begin{equation}
\begin{aligned}
\alpha_{ij,a}^{s}
&=\operatorname{softmax}_{j\in\mathcal{J}_i^s}(\ell_{ij,a}^{s}),\\
\mathbf{y}_{i,a}^{s}
&=\sum_{j\in\mathcal{J}_i^{s}}
\alpha_{ij,a}^{s}\mathbf{v}_{ij,a}^{s}.
\end{aligned}
\end{equation}
The resulting weight $\alpha_{ij,a}^s$ is assigned to source $j$ for query link $i$ in head $a$ of stream $s$, and $\mathbf{y}_{i,a}^s$ is the corresponding aggregated value vector. The source set $\mathcal{J}_i^{LL}$ contains valid links other than $i$, while $\mathcal{J}_i^{LO}$ contains the object patches.

Within each stream, the head outputs are concatenated and projected to the hidden dimension. The two stream outputs are added to the input link tokens through separate condition-dependent residual gates, followed by a gated feed-forward residual update. After the DiT stack, a final layer normalization and a shared linear prediction head produce the translational and rotational velocities defined in Sec.~\ref{sec:link_flow_matching}. Relation tokens associated with padded pairs are zeroed before entering the stack, and padded link states and predictions are masked.

\subsection{Guided sampling and kinematic recovery}\label{sec:guided_sampling_recovery}

\textbf{Palm-orientation guidance.}\label{sec:conditioned_inference}
Following guided flow matching~\citep{flow_guidance}, we steer grasp generation toward a desired palm orientation through an inference-time velocity correction. Sampling starts at $t=0$, with the translational initialization described in Sec.~\ref{sec:link_flow_matching}. For rotational initialization, we align the palm of an initial hand configuration with the requested orientation and obtain the link-wise sampling centers through forward kinematics, while retaining the finger joint configuration. Temperature-controlled perturbations are then applied around these link orientations.

Let $p$ denote the palm link and $\mathbf{R}_p^c\in \mathrm{SO}(3)$ its desired orientation in the object-centric frame. The geodesic orientation residual is defined as
\begin{equation}
    \boldsymbol{\omega}_t^c
    =\operatorname{Log}(\mathbf{R}_p^c\mathbf{R}_{t,p}^{\top}).
\end{equation}
This residual gives the gradient-descent direction of the orientation energy $\mathcal{E}_c(\mathbf{R}_{t,p})=\frac12\|\boldsymbol{\omega}_t^c\|_2^2$ in the object frame. We broadcast this palm-derived correction to all valid links and map it into each link's local coordinates:
\begin{equation}
\begin{aligned}
\tilde{\mathbf{u}}_i^x
&=\hat{\mathbf{u}}_i^x,\\
\tilde{\mathbf{u}}_i^\omega
&=\hat{\mathbf{u}}_i^\omega+
m_i\mathbf{R}_{t,i}^{\top}
\left(s(t)\lambda_c\boldsymbol{\omega}_t^c\right).
\end{aligned}
\end{equation}

\begin{figure}[!t]
    \centering
    \includegraphics[width=\columnwidth]{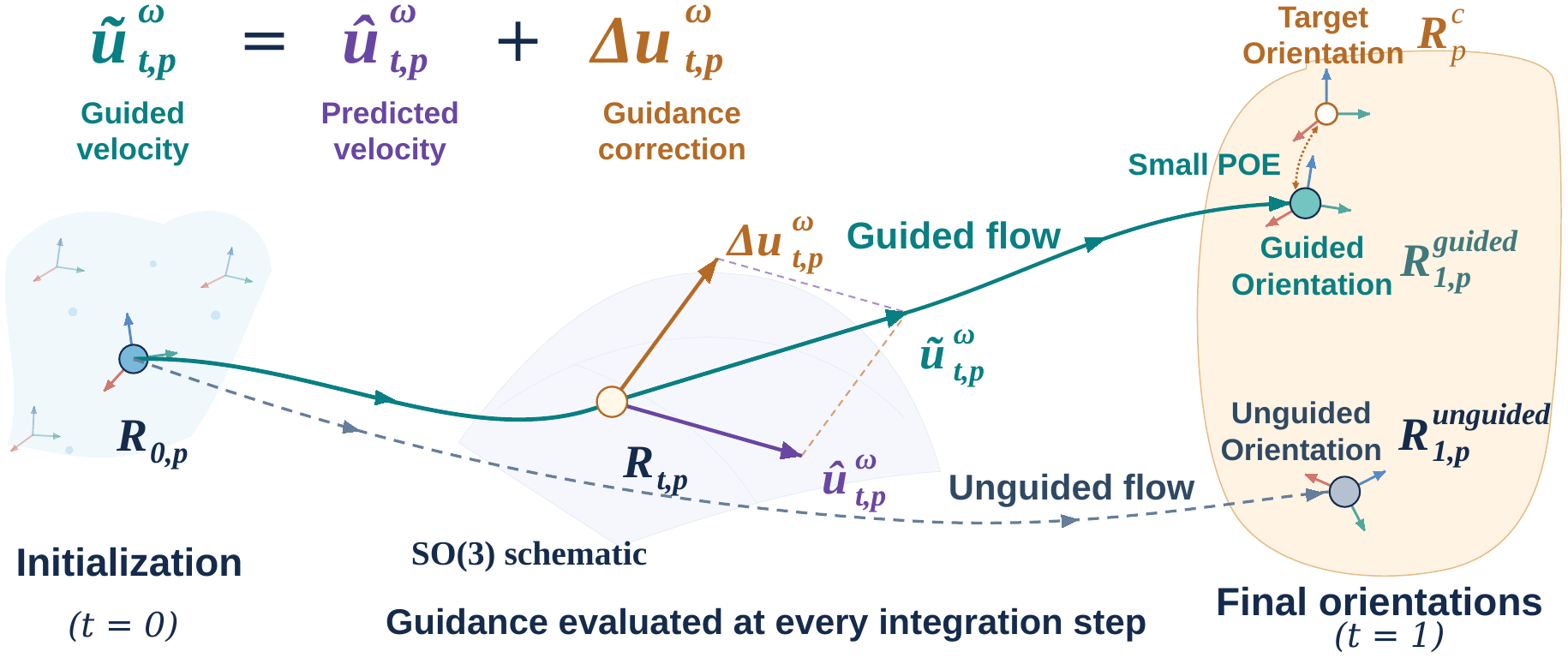}
    \caption{Inference-time palm-orientation guidance. The parallelogram depicts $\tilde{\mathbf{u}}_{t,p}^{\omega}=\hat{\mathbf{u}}_{t,p}^{\omega}+\Delta\mathbf{u}_{t,p}^{\omega}$, where $\Delta\mathbf{u}_{t,p}^{\omega}=s(t)\lambda_c\mathbf{R}_{t,p}^{\top}\boldsymbol{\omega}_t^c$. Solid and dashed curves illustrate sampling with and without the correction from the same initial state, with endpoints shown before inverse kinematics. ``Small POE'' indicates a small orientation error relative to the target $\mathbf{R}_p^c$. Only the palm trajectory is shown, although all valid links receive the correction.}
    \label{fig:palm_orientation_guidance}
\end{figure}

Here, $s(t)=s_0\sin(\pi t/2)$ increases the guidance strength over time, with maximum value $s_0$, while $\lambda_c$ scales the correction. At each integration step, the correction is recomputed from the current palm orientation and added to the predicted rotational velocities, leaving the predicted translational velocities unchanged. The resulting velocities are integrated using the updates in Sec.~\ref{sec:link_flow_matching}, without retraining the model or back-propagating through the network. Figure~\ref{fig:palm_orientation_guidance} illustrates the velocity correction and schematically compares guided and unguided palm-orientation trajectories.

\textbf{Kinematic recovery.}
Both unguided and guided sampling produce target link poses that are converted into executable joint configurations through inverse kinematics. Let $(\hat{\mathbf{x}}_i,\hat{\mathbf{R}}_i )$ denote the generated endpoint pose. Before recovery, we restore translations to the original object coordinate frame:
\begin{equation}
    \hat{\mathbf{x}}_i^{\,\mathrm{orig}}
    =s^O\hat{\mathbf{x}}_i+\mathbf{c}^O.
    \label{eq:world_restoration}
\end{equation}

The link-pose fitting objective is formulated as:
\begin{equation}
\begin{aligned}
(\mathbf{T}_w^*,\mathbf{q}^*)
=\arg\min_{\mathbf{T}_w,\mathbf{q}}\;
&\sum_{i=1}^{L_{\max}}m_i
\Bigl(
\|w_{x,i}\mathbf{e}_i^x(\mathbf{T}_w,\mathbf{q})\|_2^2\\
&\qquad+
\|w_{R,i}\mathbf{e}_i^R(\mathbf{T}_w,\mathbf{q})\|_2^2
\Bigr).
\end{aligned}
\end{equation}

The variables $\mathbf{T}_w$ and $\mathbf{q}$ denote the wrist pose
and finger joint angles, respectively.
The residuals $\mathbf{e}_i^x$ and $\mathbf{e}_i^R$ measure the
translational and rotational discrepancies between the generated
target and the forward-kinematics pose of link $i$.
The nonnegative weights $w_{x,i}$ and $w_{R,i}$ specify the
corresponding link-wise fitting priorities.
We perform batched kinematic optimization using
Pyroki~\citep{pyroki}.

\begin{table*}[!t]
    \centering
    \caption{Comparison of unconditioned grasp success rates and single-sample end-to-end generation latency across three hand embodiments.}
    \label{tab:unconditioned_generation}
    \footnotesize
    \setlength{\tabcolsep}{3.5pt}
    \begin{tabular}{l|cccc|cccc}
        \hline
        \multirow{2}{*}{Method}
        & \multicolumn{4}{c|}{Success Rate (\%)}
        & \multicolumn{4}{c}{Efficiency (s) $\downarrow$} \\
        \cline{2-9}
        & Barrett & Allegro & ShadowHand & Avg.
        & Barrett & Allegro & ShadowHand & Avg. \\
        \hline
        DFC~\citep{dfc}
        & 86.30 & 76.21 & 58.80 & 73.77
        & $>1800$ & $>1800$ & $>1800$ & $>1800$ \\

        GenDexGrasp~\citep{gendexgrasp}
        & 57.40 & 43.10 & 42.30 & 47.60
        & 12.933 & 18.347 & 16.540 & 15.940 \\

        DRO-Grasp~\citep{dro}
        & 86.60 & 91.20 & 81.60 & 86.47
        & 0.523 & 0.485 & 0.856 & 0.621 \\

        TRO-Grasp~\citep{tro}
        & 92.40 & 93.30 & 98.10 & 94.60
        & 0.254 & 0.247 & 0.269 & 0.257 \\

        {\bfseries DexGraspFlow (Ours)}
        & {\bfseries 92.10} & {\bfseries 98.40} & {\bfseries 95.50} & {\bfseries 95.33}
        & {\bfseries 0.126} & {\bfseries 0.134} & {\bfseries 0.136} & {\bfseries 0.132} \\
        \hline
    \end{tabular}
\end{table*}

\section{Experiments}\label{sec:experiments}

We evaluate DexGraspFlow in terms of grasp quality and generation efficiency, zero-shot generalization to topology-modified hands, and palm-orientation-conditioned generation. Ablation studies examine the effects of attention biases, integration steps, and guidance strength. Real-robot experiments further evaluate the execution of generated grasps on a physical robotic platform.

\subsection{Experimental setup}\label{sec:experimental_setup}

\textbf{Dataset.}
We use the filtered CMapDataset~\citep{gendexgrasp} split adopted by DRO-Grasp~\citep{dro}. The training set contains 14,011 grasp demonstrations involving 48 objects from ContactDB~\citep{contactdb} and YCB~\citep{ycb}, covering Barrett, Allegro, and ShadowHand. Ten additional objects are held out to evaluate generalization to unseen object geometries: six from ContactDB (apple, camera, medium cylinder, door knob, rubber duck, and water bottle) and four from YCB (baseball, pear, potted meat can, and tomato soup can). Annotated joint configurations are converted through forward kinematics into object-centric link poses for flow-matching supervision.

\textbf{Evaluation protocol and metrics.}
Following the standard benchmark protocol in DRO-Grasp~\citep{dro} and TRO-Grasp~\citep{tro}, we generate 100 grasp candidates per object--hand pair and evaluate them in Isaac Gym. External forces are sequentially applied along the \(+x,+y,+z,-x,-y,\) and \(-z\) directions, each for one second with a magnitude of 0.5~N per kilogram of object mass. A grasp is successful if the Euclidean displacement of the object between the beginning and end of the six-force sequence does not exceed \(2\,\mathrm{cm}\). Success rate (SR) is the percentage of candidates satisfying this criterion.

Generation latency encompasses complete model inference and Pyroki inverse kinematics without physics simulation. To ensure a consistent execution benchmark, all runtime evaluations are conducted on an NVIDIA GeForce RTX 5090 GPU.

\textbf{Implementation and training details.}
For each object, 65,536 mesh surface points are pre-sampled to form a surface-point reservoir. The object encoder receives \(N=512\) points, selects \(N^O=25\) patch centers using farthest point sampling, and groups \(K_{\mathrm{nn}}=24\) neighboring points around each center to extract $d_f^O=64$-dimensional local geometric features. The hand encoder combines local link geometry, local centroids and radii, and palm-relative reference poses to produce $d_s^H=128$-dimensional static link tokens. The dynamic pose branch produces $d_d^H=64$-dimensional features, which are combined with the static tokens before projection into the DiT. The dual-stream DiT uses a hidden dimension of $d_h=384$, eight Transformer blocks, $n_h=12$ attention heads, and $d_r=64$-dimensional relation tokens. Hand representations are padded to \(L_{\max}=26\) links, with validity masks excluding padded entries. Topology distances are clipped at \(d_{\mathrm{clip}}=15\).

We train the model for 400 epochs using AdamW with a batch size of 16, a learning rate of \(5\times10^{-4}\), and a weight decay of \(10^{-4}\). The learning rate follows a cosine schedule with a 20-epoch warm-up. Gradient clipping uses a maximum norm of 0.2, and an exponential moving average is maintained with a decay of 0.999. The flow-matching loss weights are set to \(\lambda_x=\lambda_\omega=1.0\). The final velocity-prediction head is initialized with zero weights and biases. Unless otherwise stated, simulation results use the final checkpoint saved at epoch 400. During training, object and hand observations are augmented via surface-point resampling, link-level geometric perturbations, and token masking.

\textbf{Inference details.}
Unless otherwise specified, we use an $S=10$-step Euler solver with translation and rotation sampling temperatures set to $\tau_x=\tau_R=0.2$. 

\begin{figure}[!t]
    \centering
    \footnotesize
    \setlength{\tabcolsep}{0.3pt}
    \renewcommand{\arraystretch}{0.90}
    \begin{tabular}{@{}c@{}ccccc|c@{}}
        \raisebox{2.20em}[0pt][0pt]{\parbox[c]{0.10\columnwidth}{\centering\scriptsize\textit{Allegro}}} &
        \includegraphics[width=0.145\columnwidth]{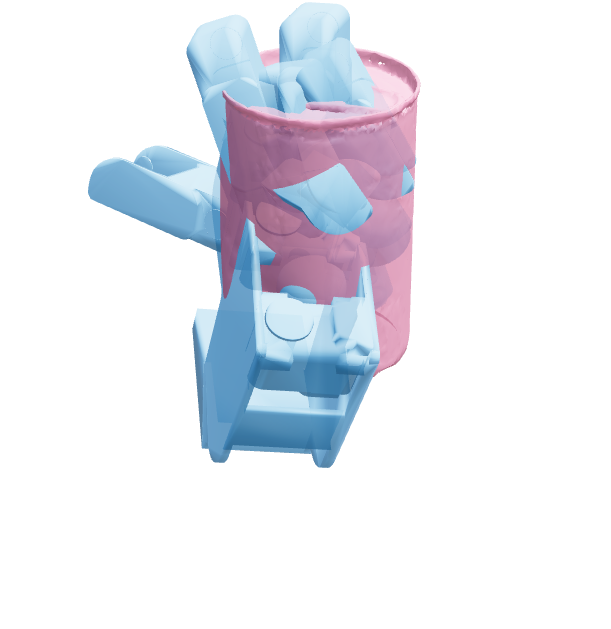} &
        \includegraphics[width=0.145\columnwidth]{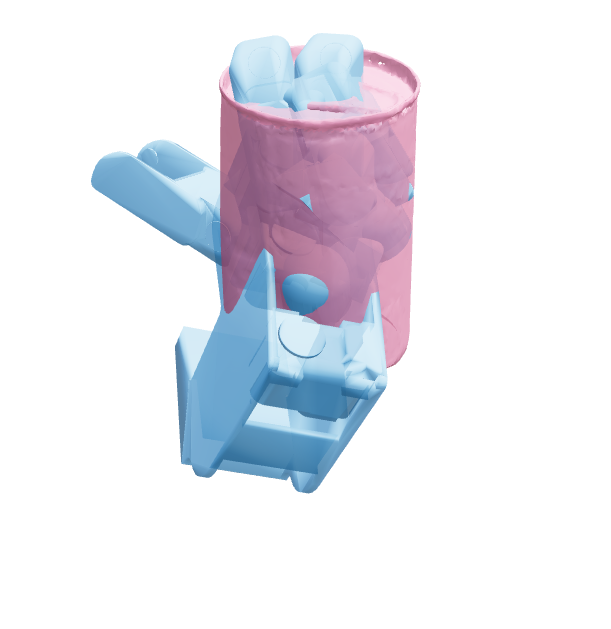} &
        \includegraphics[width=0.145\columnwidth]{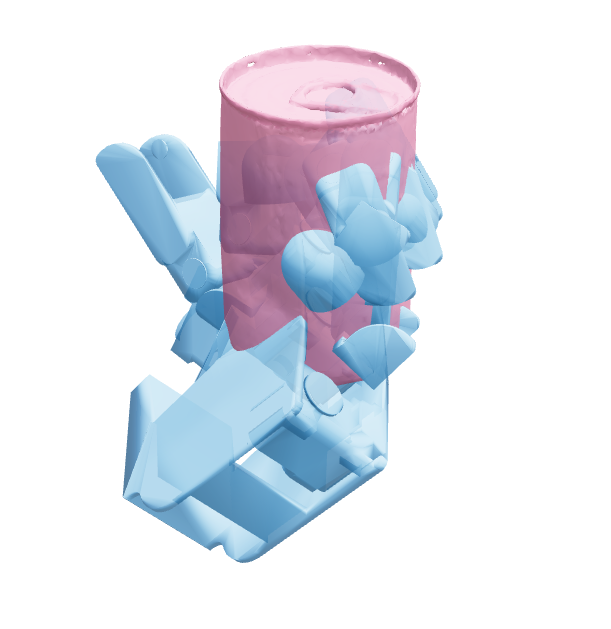} &
        \includegraphics[width=0.145\columnwidth]{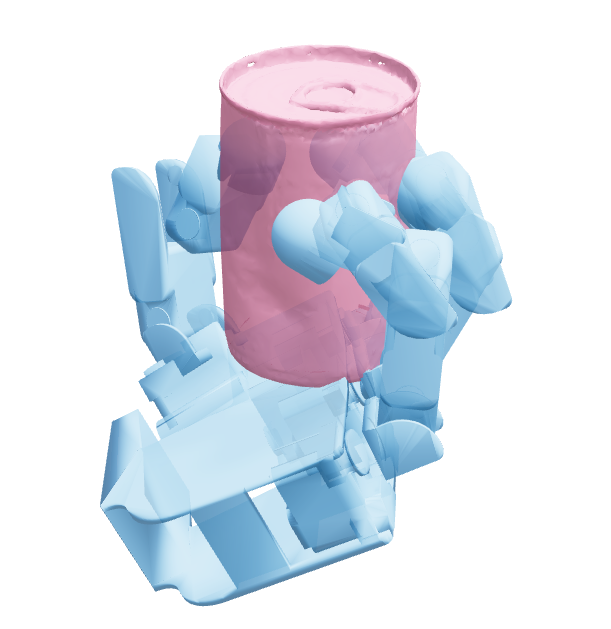} &
        \includegraphics[width=0.145\columnwidth]{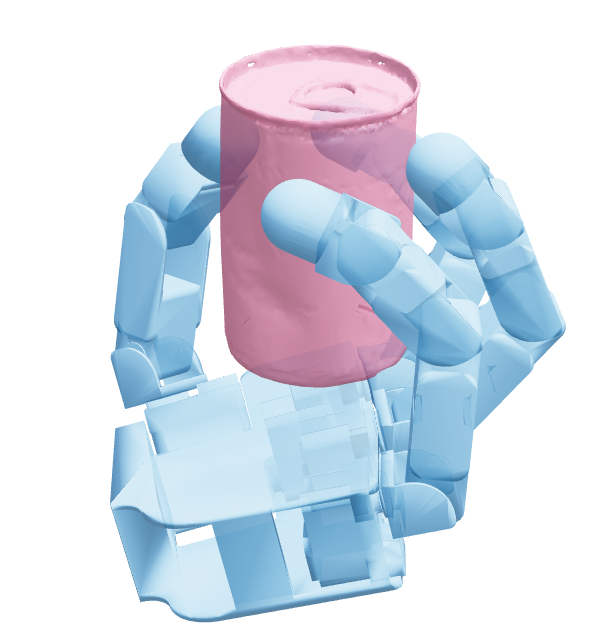} &
        \includegraphics[width=0.145\columnwidth]{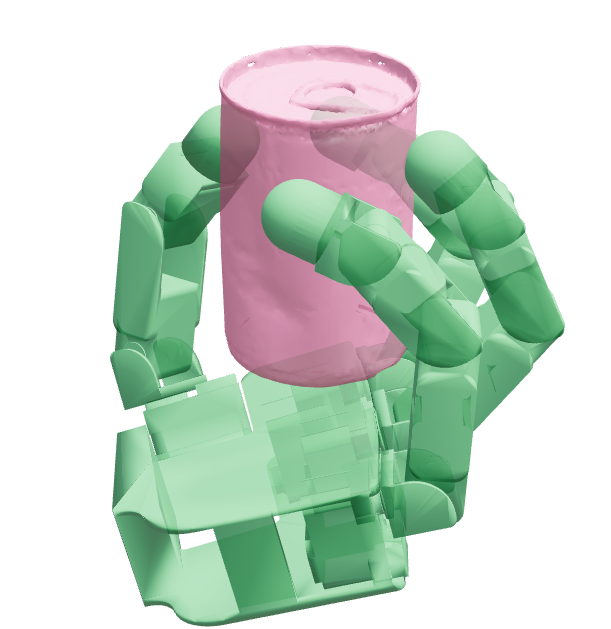} \\
        \raisebox{2.20em}[0pt][0pt]{\parbox[c]{0.10\columnwidth}{\centering\scriptsize\textit{Barrett}}} &
        \includegraphics[width=0.145\columnwidth]{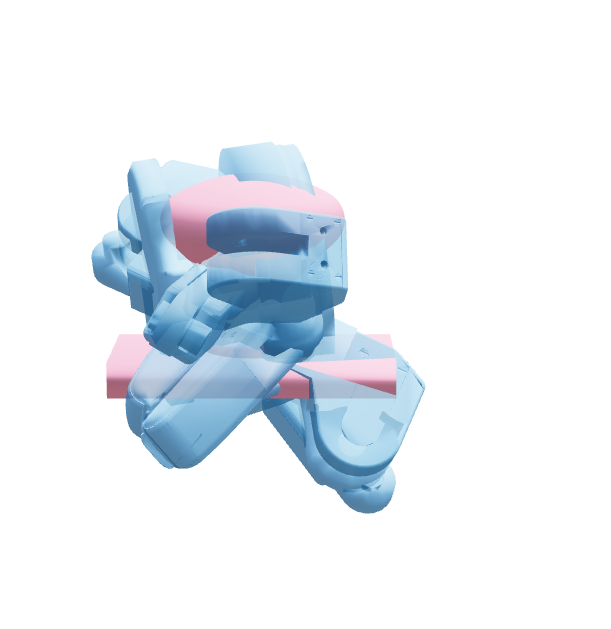} &
        \includegraphics[width=0.145\columnwidth]{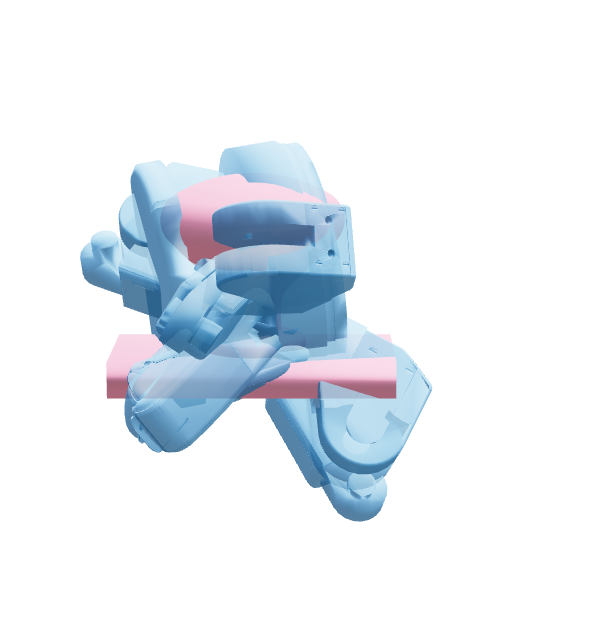} &
        \includegraphics[width=0.145\columnwidth]{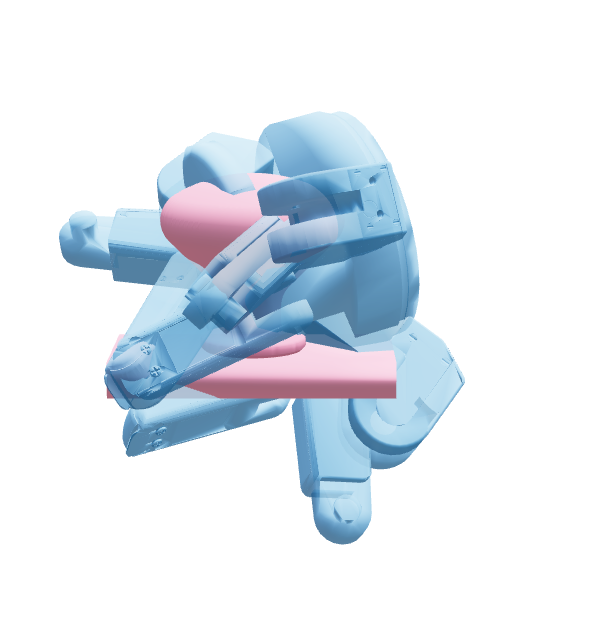} &
        \includegraphics[width=0.145\columnwidth]{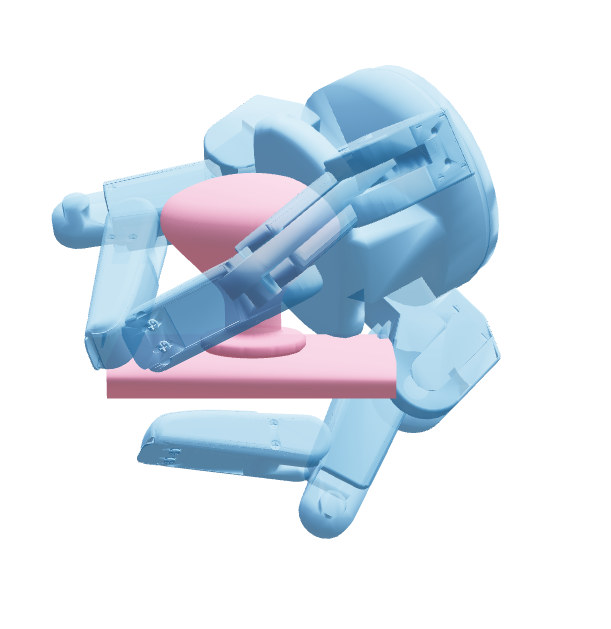} &
        \includegraphics[width=0.145\columnwidth]{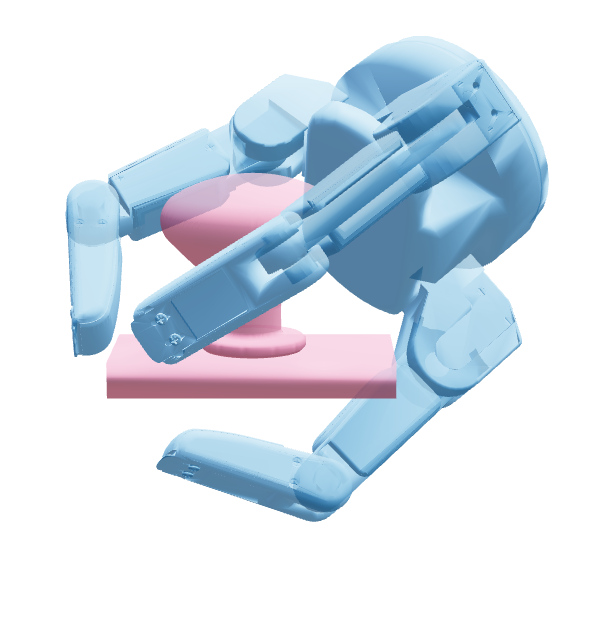} &
        \includegraphics[width=0.145\columnwidth]{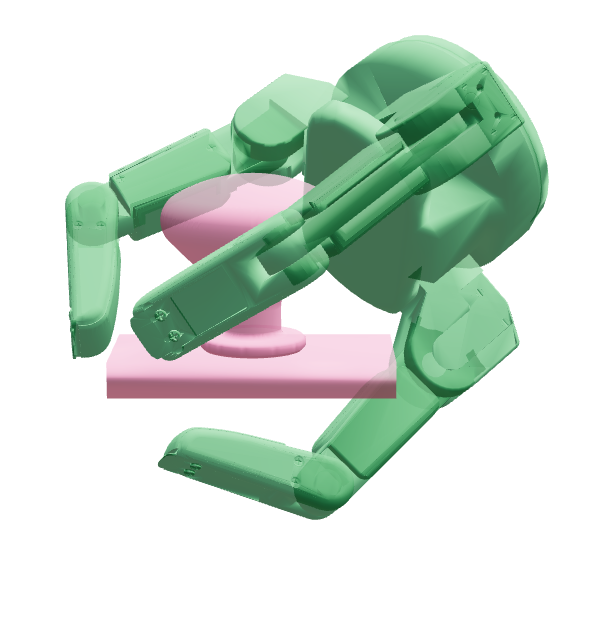} \\
        \raisebox{1.80em}[0pt][0pt]{\parbox[c]{0.10\columnwidth}{\centering\scriptsize\textit{Shadow}\\[-1pt]\textit{hand}}} &
        \includegraphics[width=0.145\columnwidth]{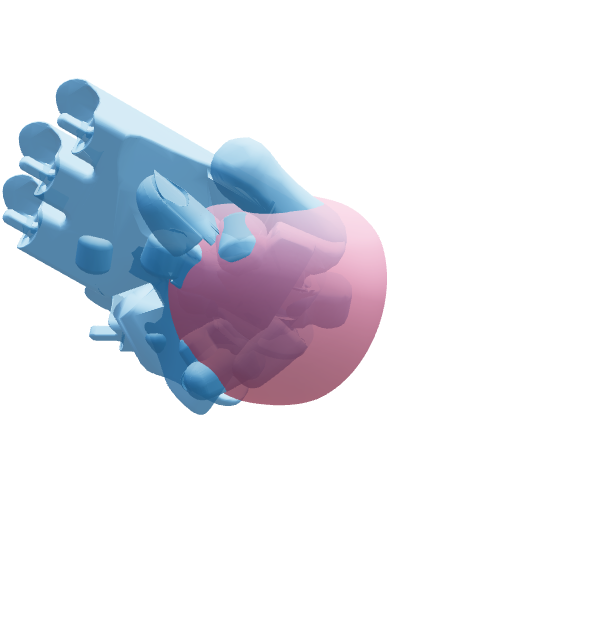} &
        \includegraphics[width=0.145\columnwidth]{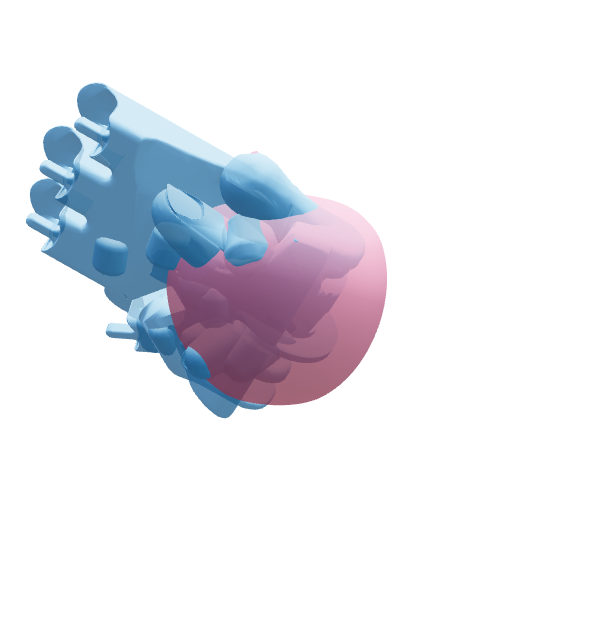} &
        \includegraphics[width=0.145\columnwidth]{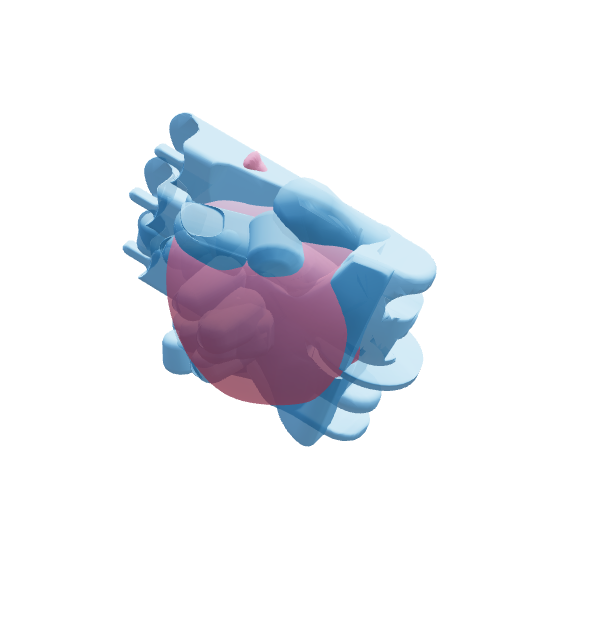} &
        \includegraphics[width=0.145\columnwidth]{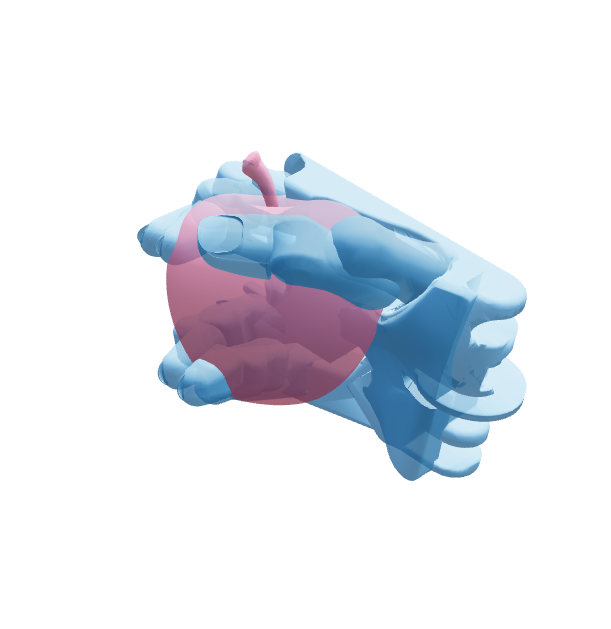} &
        \includegraphics[width=0.145\columnwidth]{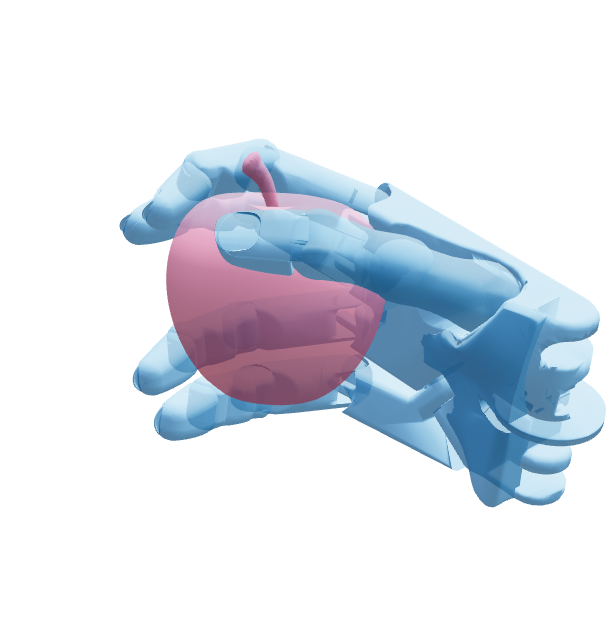} &
        \includegraphics[width=0.145\columnwidth]{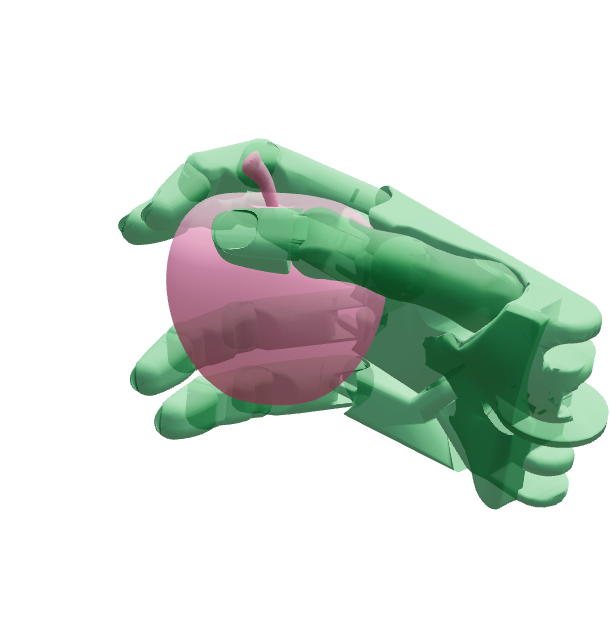} \\[-3pt]
        & \hspace{-0.80em}\scriptsize \(t=0\) & \hspace{-0.80em}\scriptsize \(t=0.2\)
        & \scriptsize \(t=0.5\) & \scriptsize \(t=0.8\)
        & \scriptsize \(t=1.0\) & \scriptsize After IK
    \end{tabular}
    \caption{Representative flow-matching trajectories for three hand--object pairs. The rows correspond to Allegro, Barrett, and ShadowHand; the associated objects are a YCB tomato soup can, a ContactDB door knob, and a ContactDB apple, respectively. Blue images show selected states along the trajectory from \(t=0\) to \(t=1\), while the final column shows the executable grasp obtained after Pyroki inverse kinematics.}
    \label{fig:flow_trajectory}
\end{figure}

\subsection{Grasp quality and efficiency}\label{sec:unconditioned_generation}

\textbf{Evaluation protocol.}
We evaluate unconditioned grasp generation on all 30 held-out object--hand pairs without orientation guidance, yielding 3,000 simulated trials (100 candidates per pair). All learning-based baselines (GenDexGrasp, DRO-Grasp, and TRO-Grasp) and DexGraspFlow are re-evaluated locally using official implementations. Following~\citep{dro,tro}, single-sample latency is measured using a pre-compiled Pyroki IK solver. For optimization-based DFC~\citep{dfc}, we strictly follow DRO-Grasp~\citep{dro} by evaluating pre-generated grasps from CMapDataset due to its extensive optimization runtime.

\textbf{Overall performance.}
Table~\ref{tab:unconditioned_generation} compares grasp success rates and single-sample latency across three hand embodiments. DexGraspFlow achieves the highest average success rate of 95.33\%, exceeding the strongest baseline TRO-Grasp by 0.73 percentage points. In particular, it demonstrates the most pronounced gain on Allegro with a 98.40\% success rate (outperforming TRO-Grasp by 5.10 percentage points), while maintaining highly competitive performance across the remaining hand embodiments.

\begin{figure}[!t]
    \centering
    \includegraphics[width=\columnwidth]{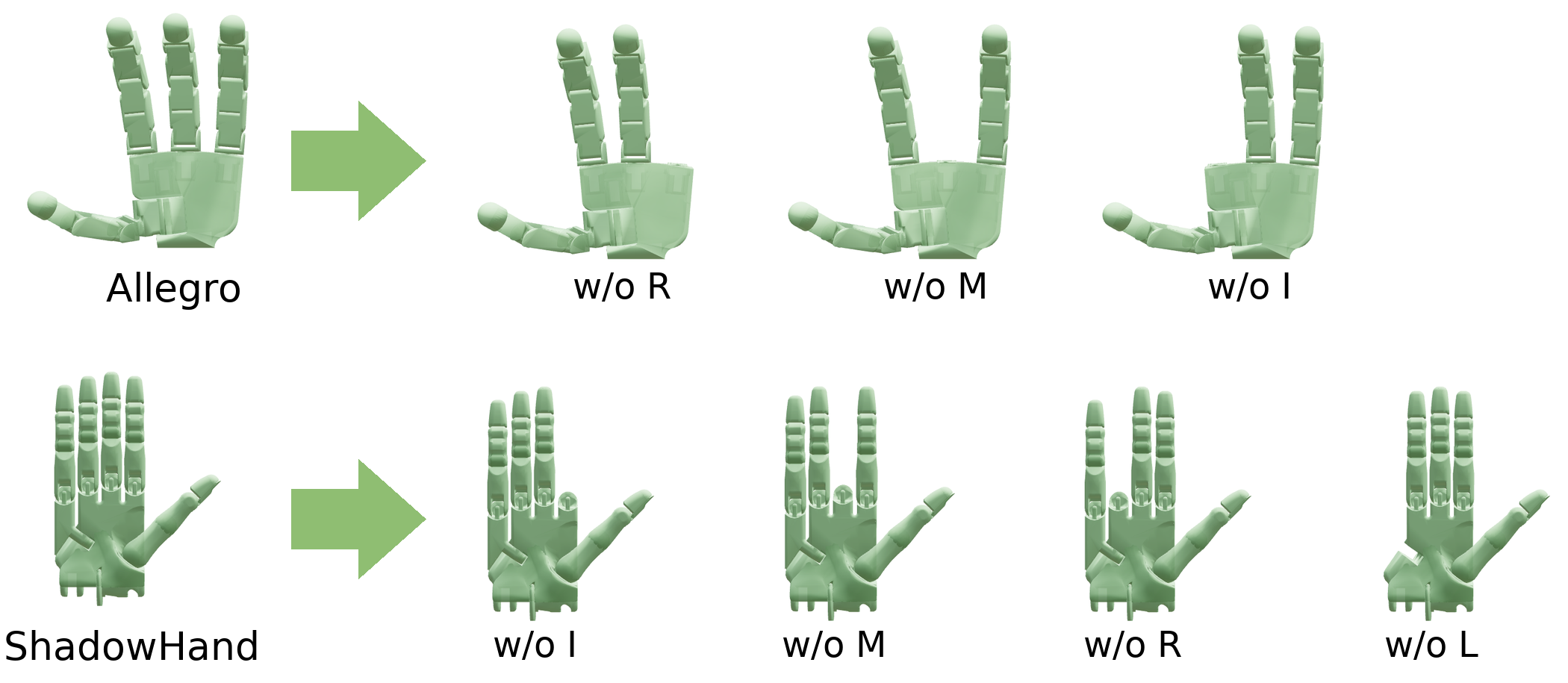}
    \caption{Intact and topology-modified hand embodiments used for zero-shot
    evaluation. The arrows indicate the transition from each intact hand to
    the corresponding missing-finger variants. Here, I, M, R, and L denote the
    index, middle, ring, and little fingers, respectively.}
    \label{fig:topology_variants}
\end{figure}

\begin{figure*}[!t]
    \centering
    {\scriptsize
    \setlength{\tabcolsep}{3pt}
    \def\zerogridroot{fig}
    \def\zerogridwidth{0.125\textwidth}
    \def\zerogridimagewidth{0.115\textwidth}
    \def\zerogridcell#1{%
        \parbox[c][\zerogridwidth][c]{\zerogridwidth}{%
            \centering
            \includegraphics[width=\zerogridimagewidth,height=\zerogridimagewidth,keepaspectratio]{#1}}}
    \begin{tabular}{@{}ccc@{\hspace{5pt}}|@{\hspace{5pt}}cccc@{}}
        \multicolumn{3}{c}{\textbf{Allegro}} &
        \multicolumn{4}{c}{\textbf{ShadowHand}} \\
        w/o I & w/o M & w/o R & w/o I & w/o M & w/o R & w/o L \\
        \zerogridcell{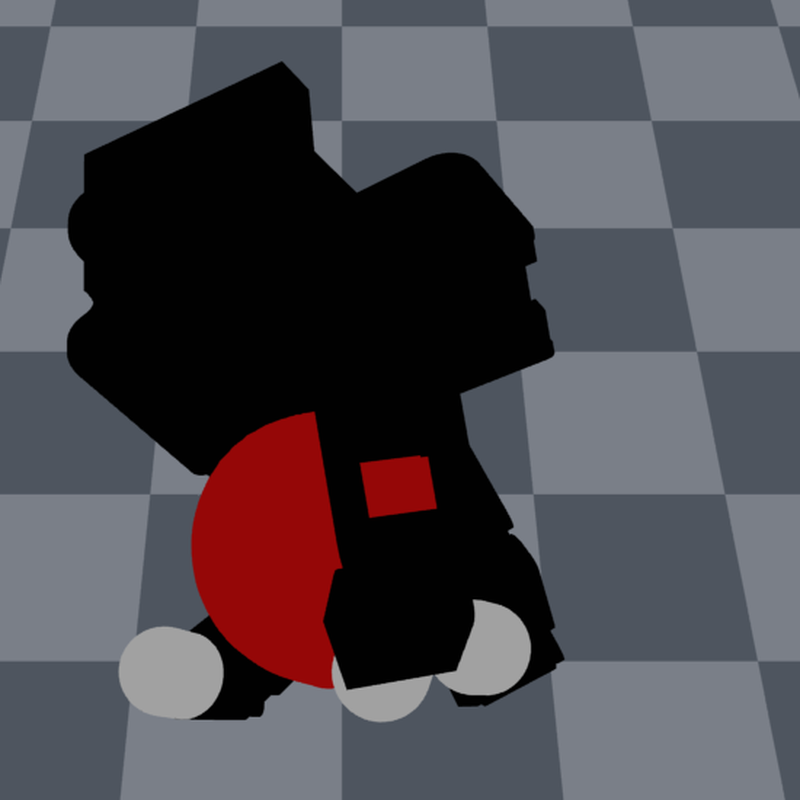} &
        \zerogridcell{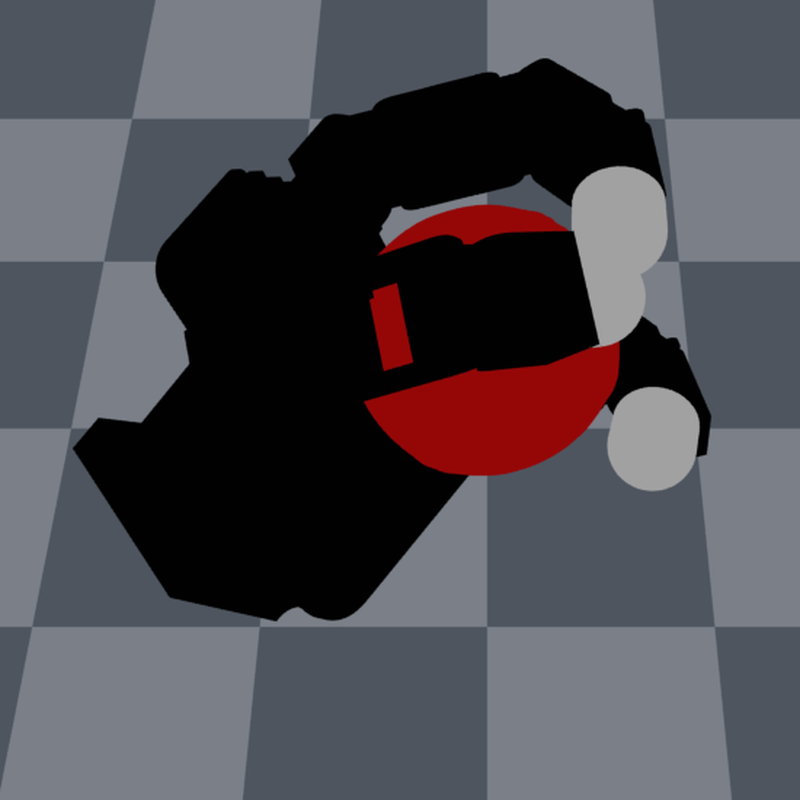} &
        \zerogridcell{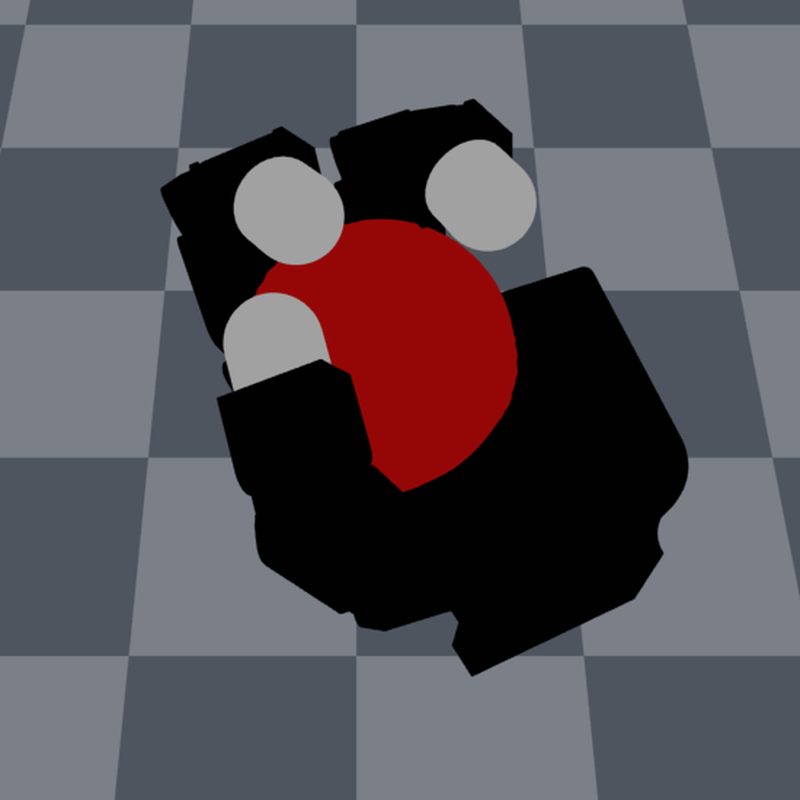} &
        \zerogridcell{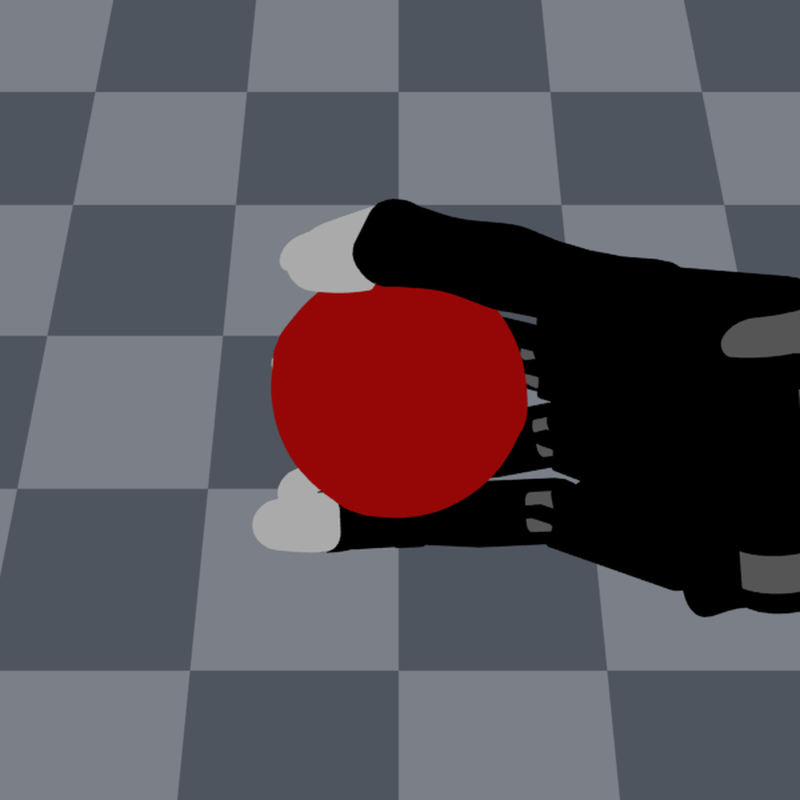} &
        \zerogridcell{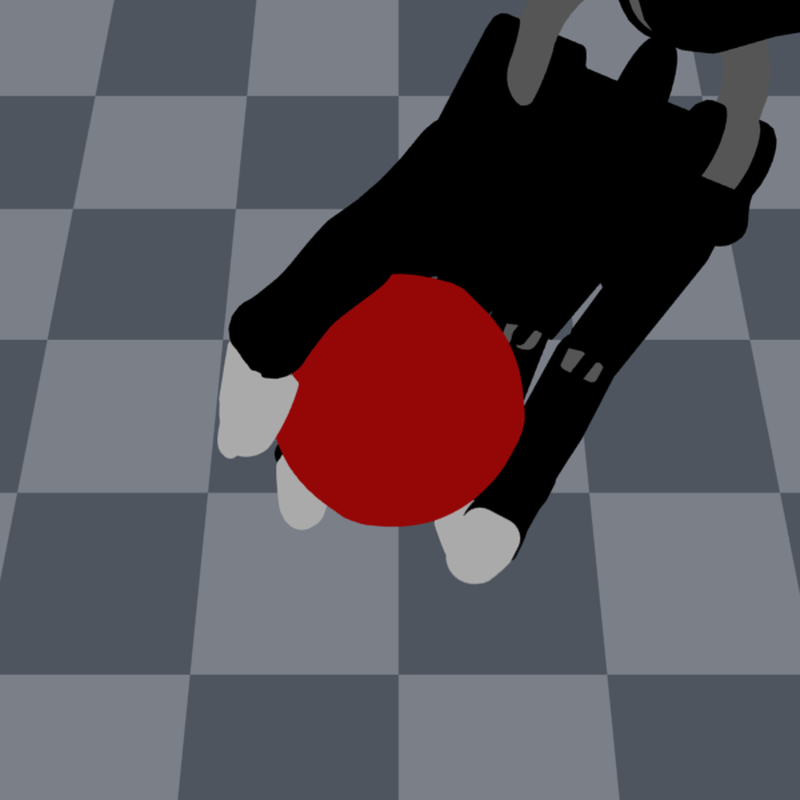} &
        \zerogridcell{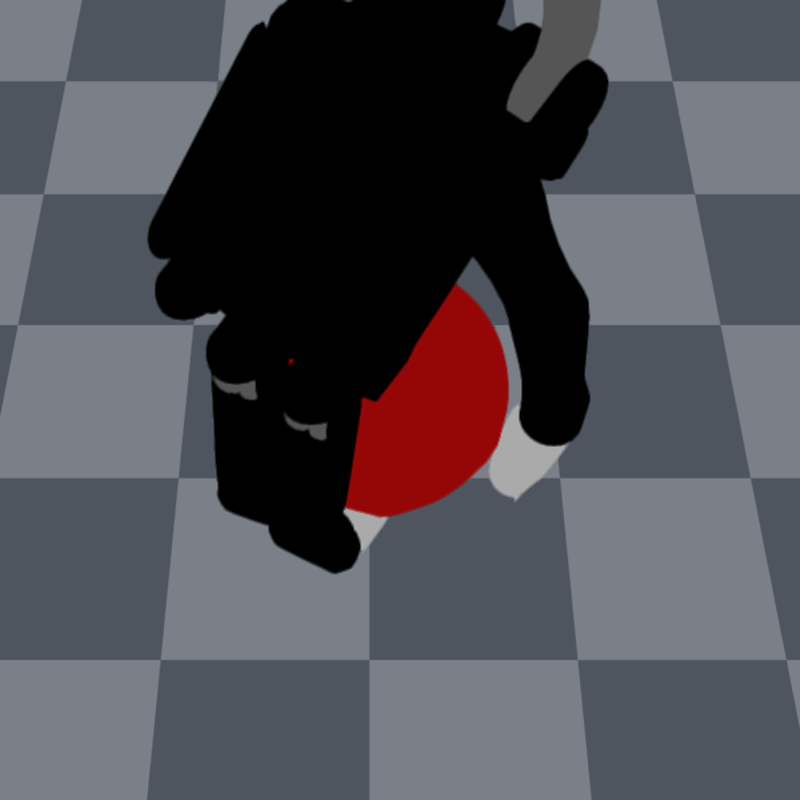} &
        \zerogridcell{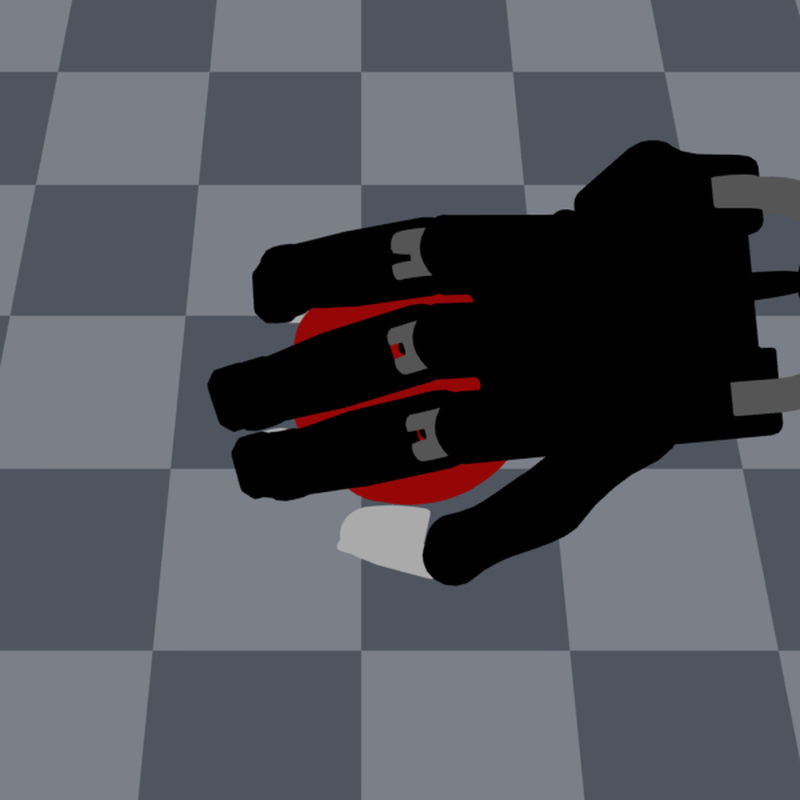} \\[4mm]
        \zerogridcell{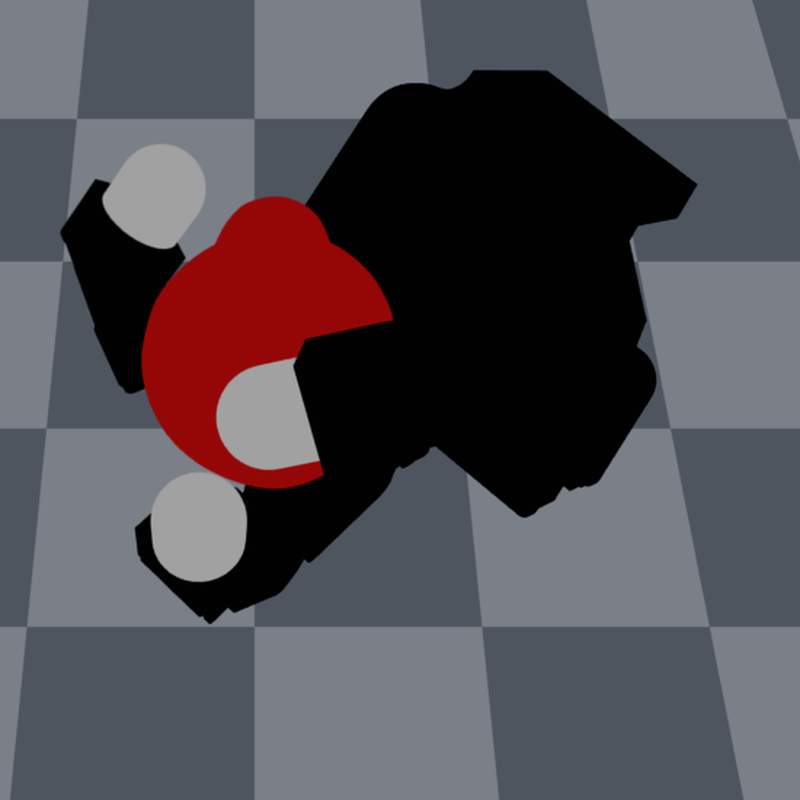} &
        \zerogridcell{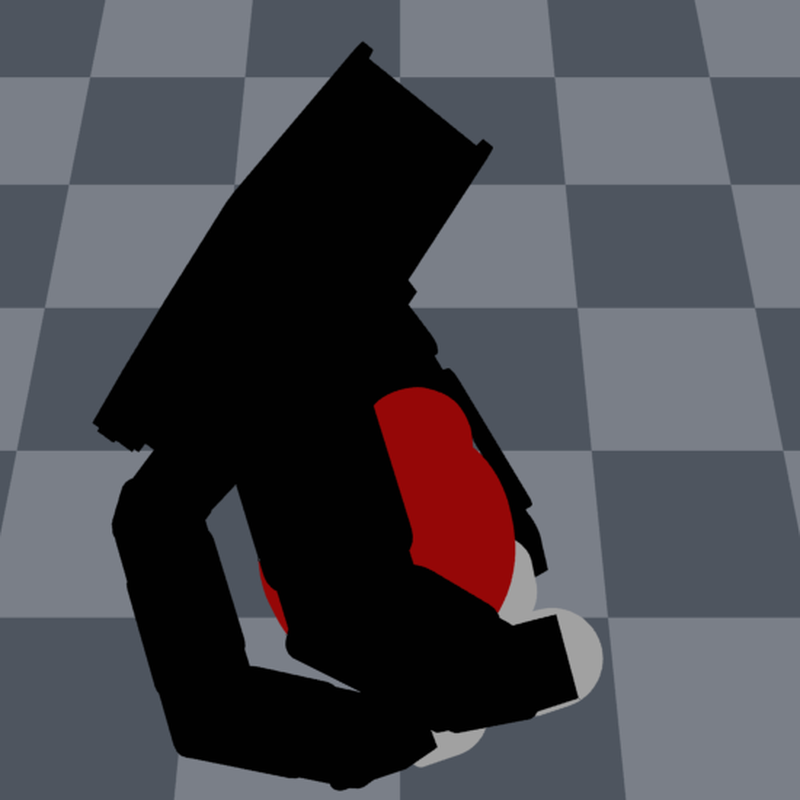} &
        \zerogridcell{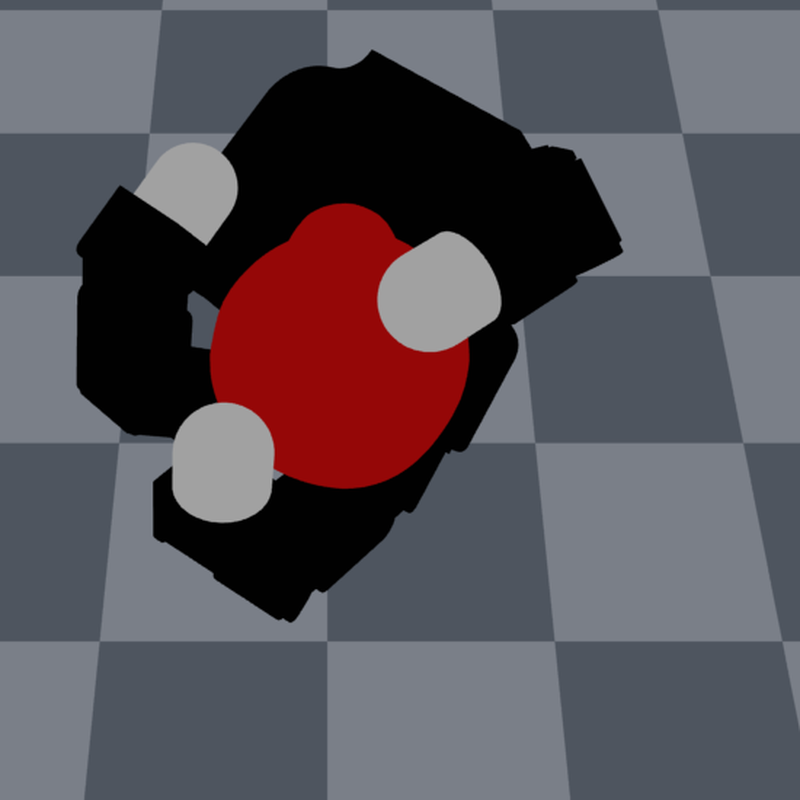} &
        \zerogridcell{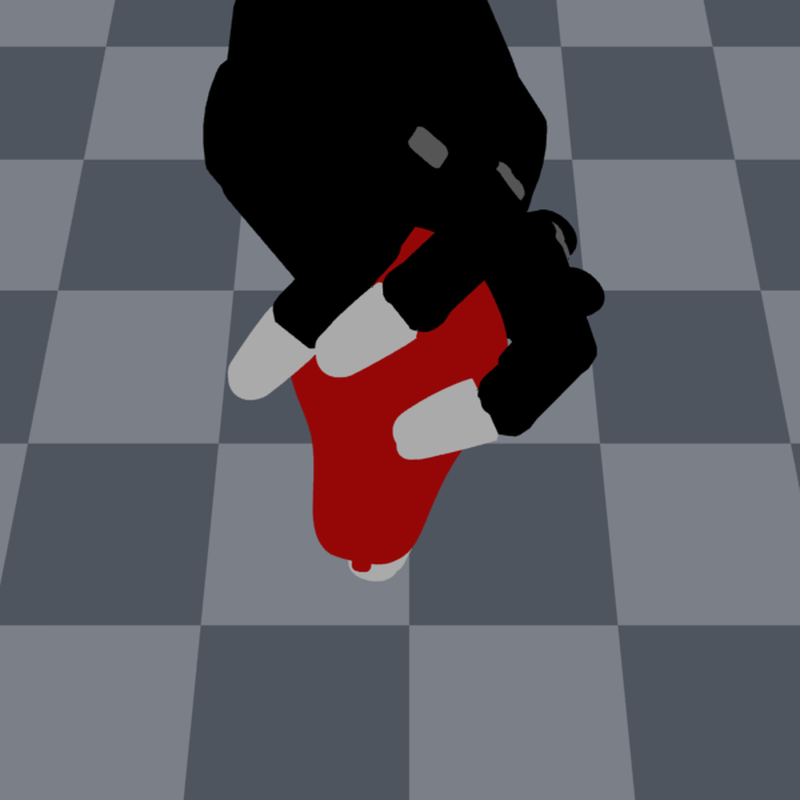} &
        \zerogridcell{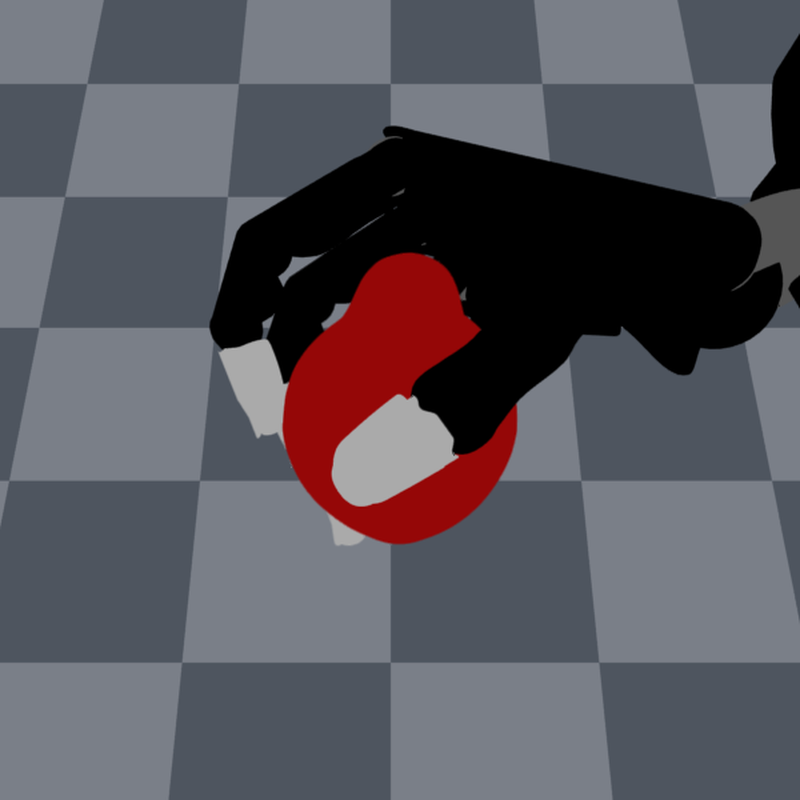} &
        \zerogridcell{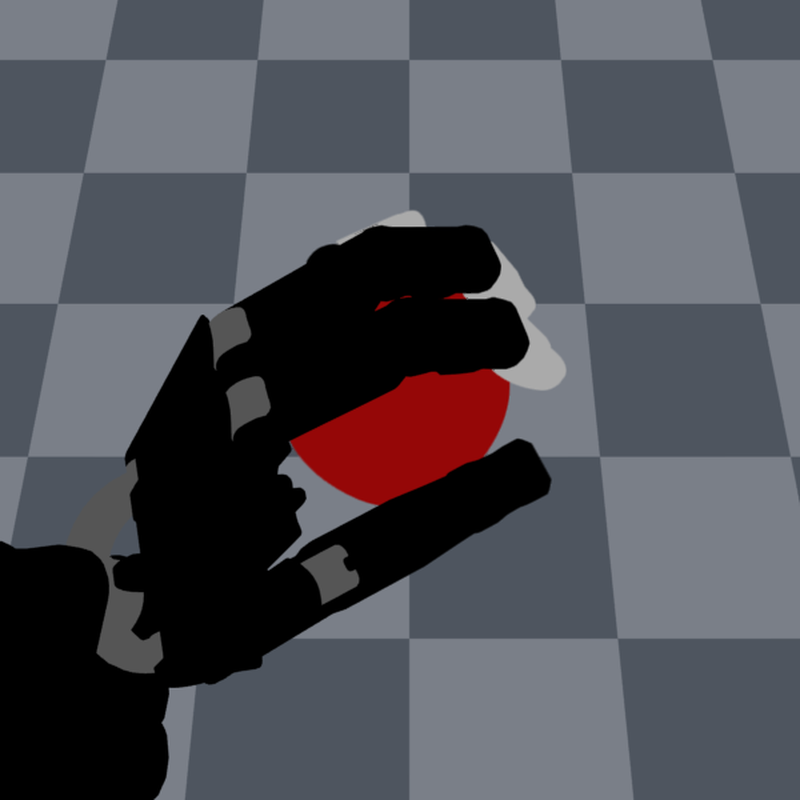} &
        \zerogridcell{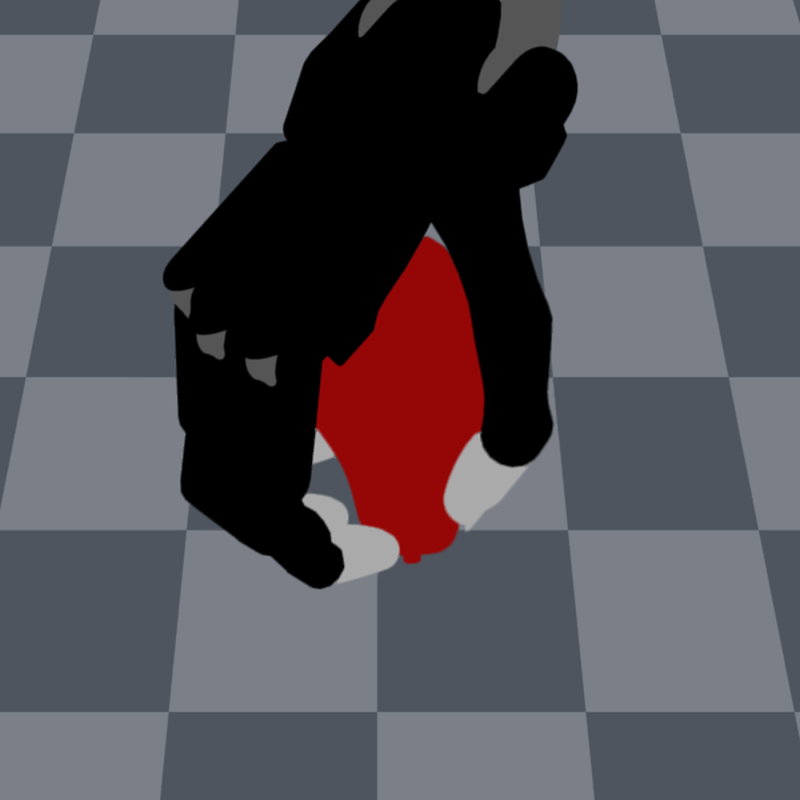} \\[4mm]
        \zerogridcell{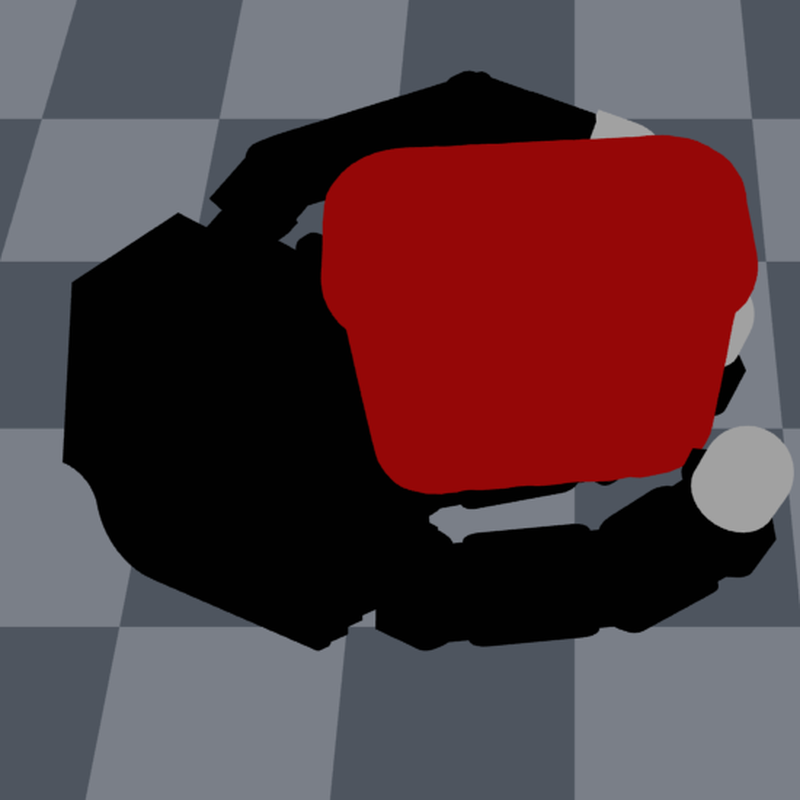} &
        \zerogridcell{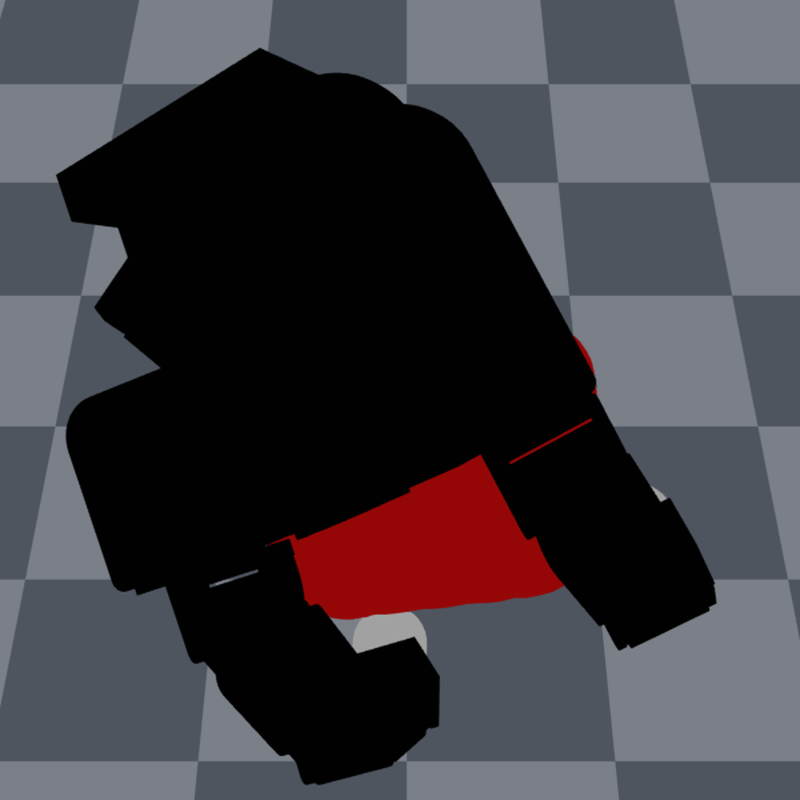} &
        \zerogridcell{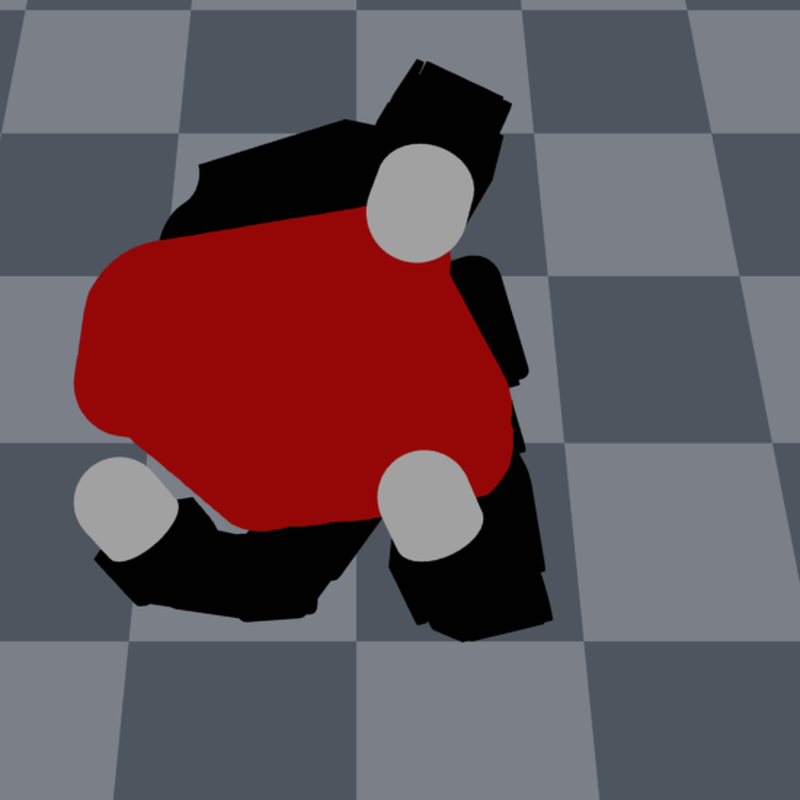} &
        \zerogridcell{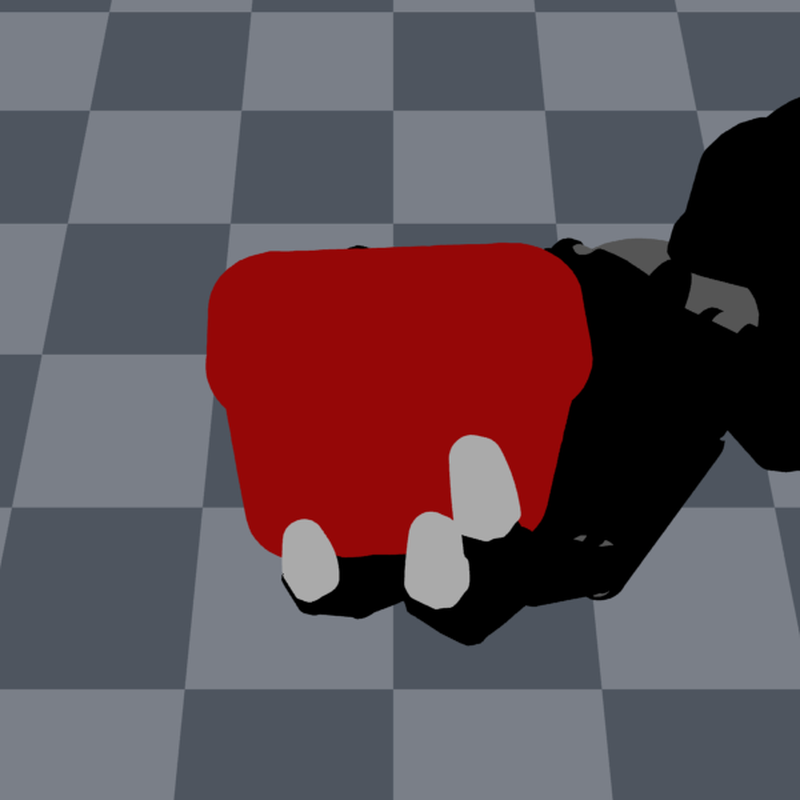} &
        \zerogridcell{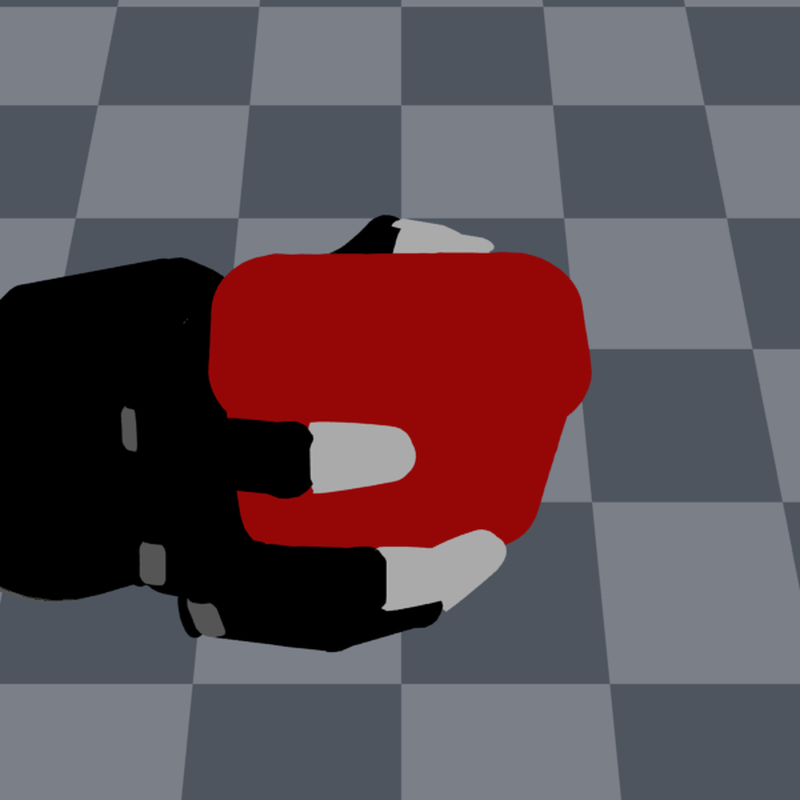} &
        \zerogridcell{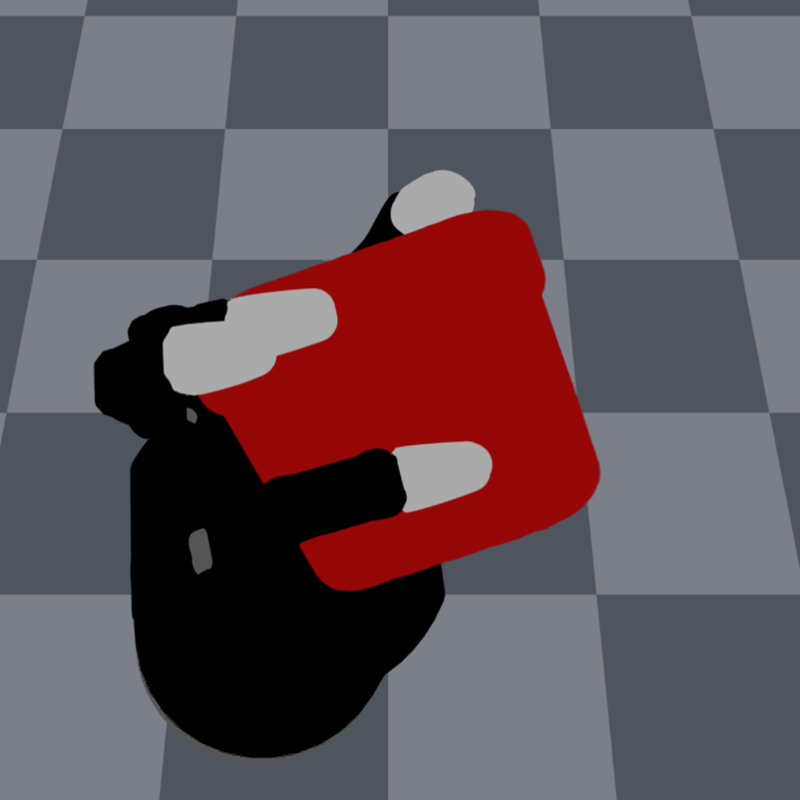} &
        \zerogridcell{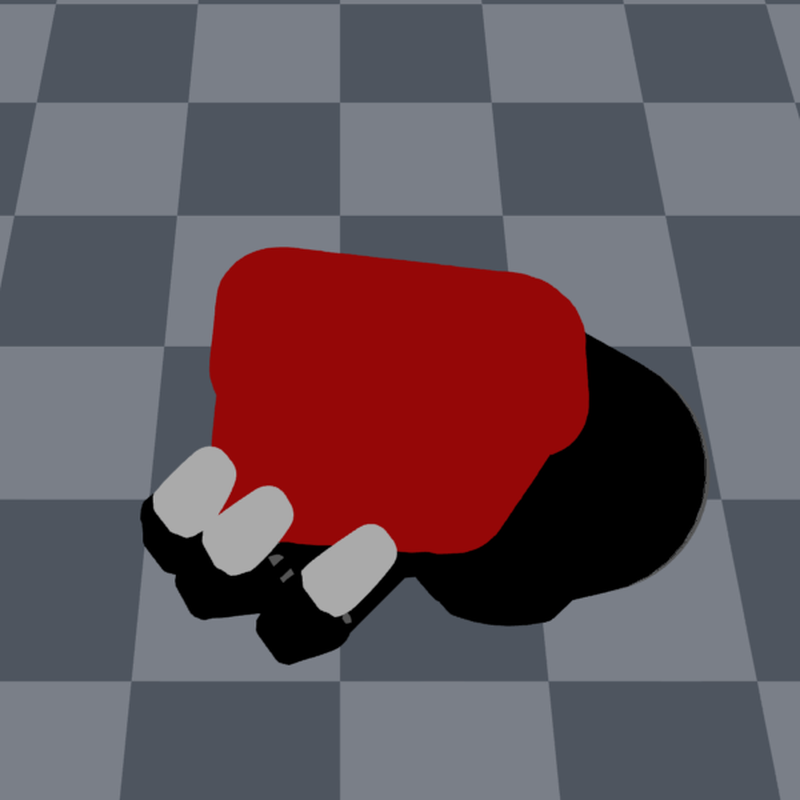} \\[4mm]
        \zerogridcell{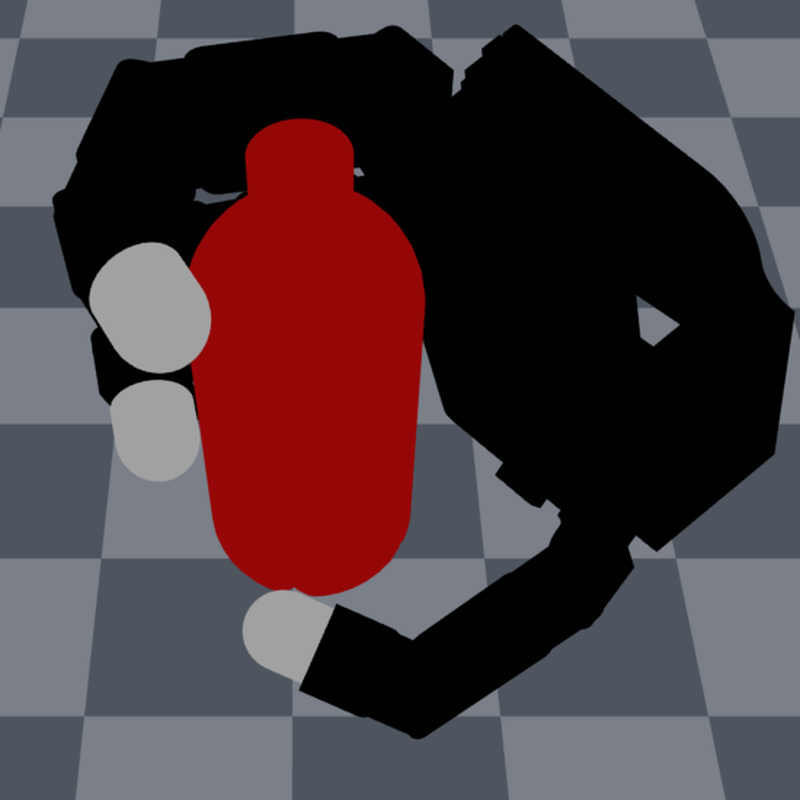} &
        \zerogridcell{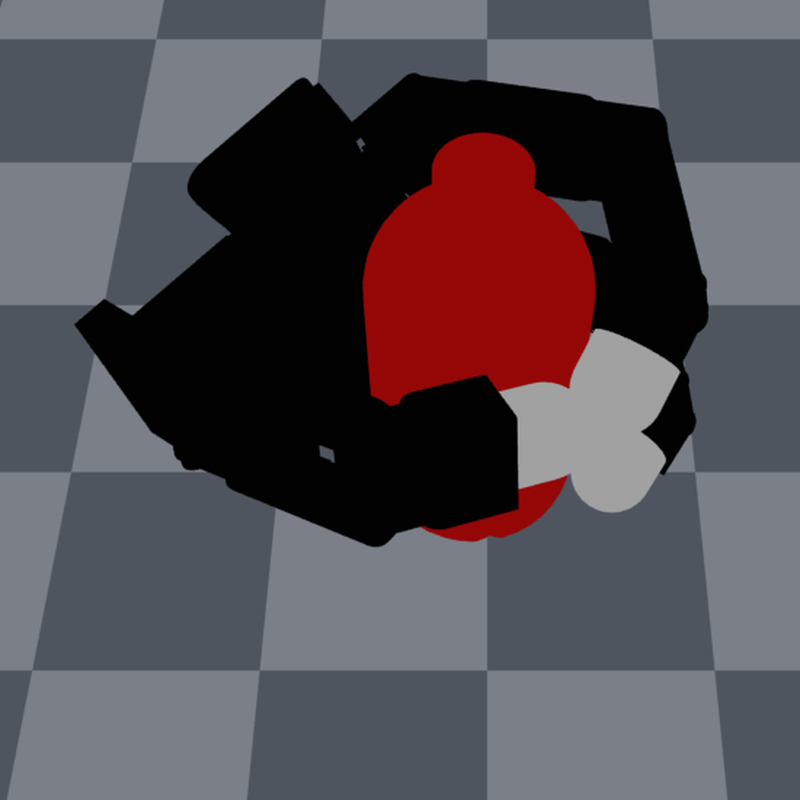} &
        \zerogridcell{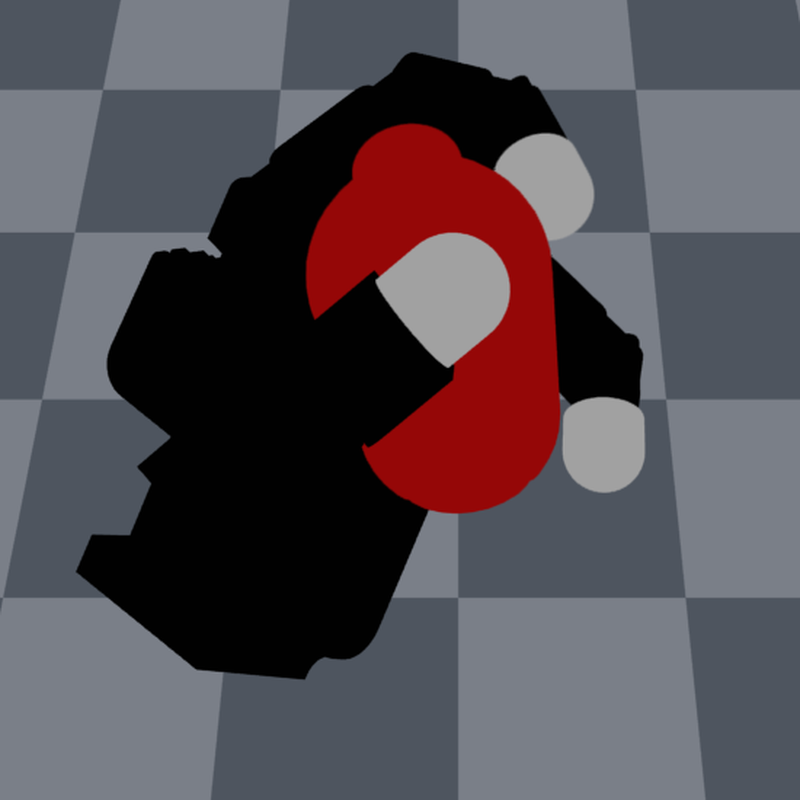} &
        \zerogridcell{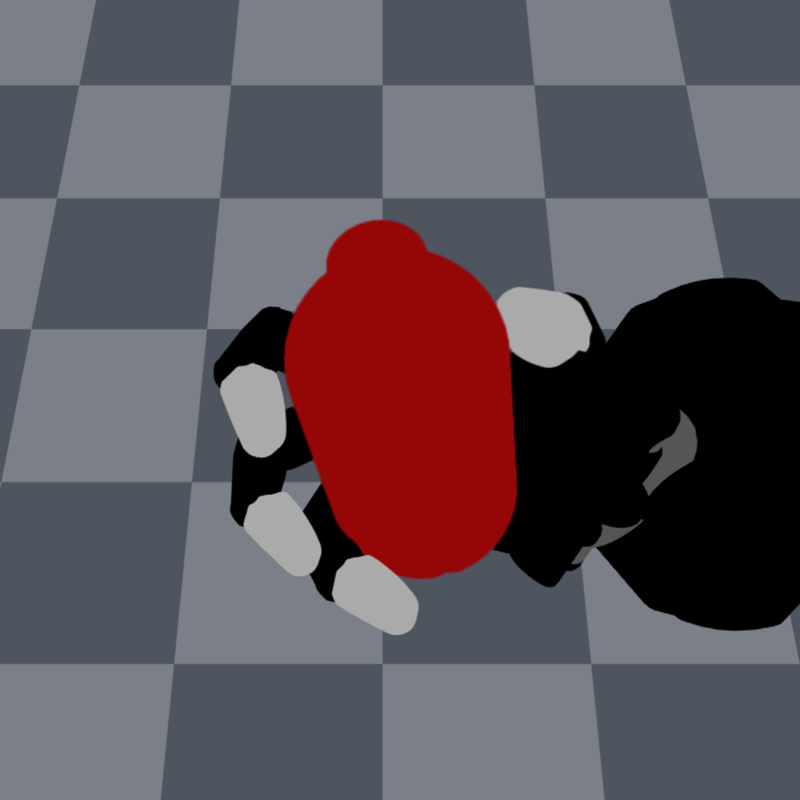} &
        \zerogridcell{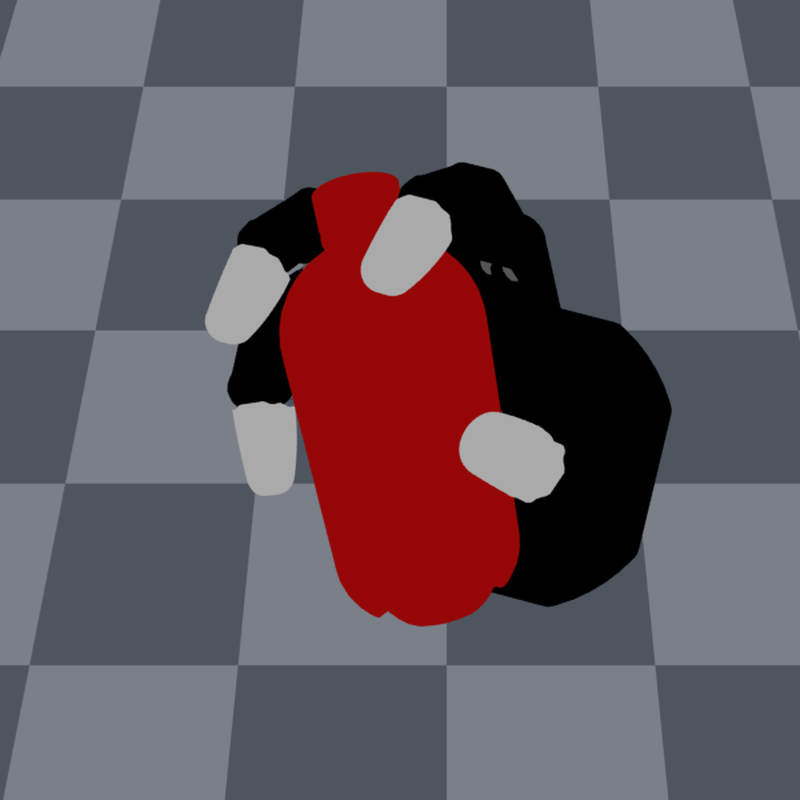} &
        \zerogridcell{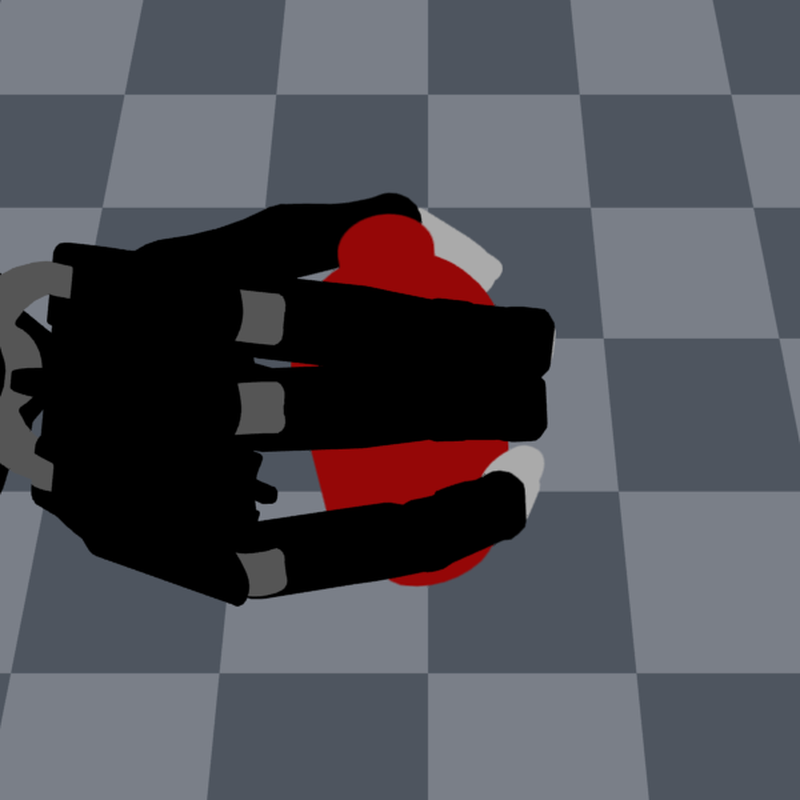} &
        \zerogridcell{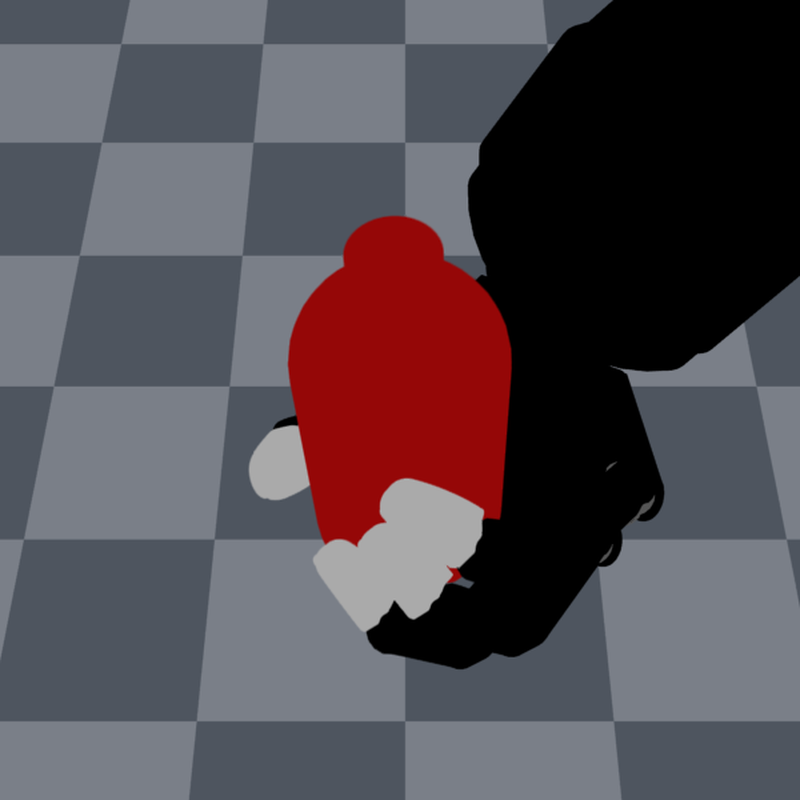}
    \end{tabular}}
    \caption{Representative zero-shot grasp examples on topology-modified
    hands. The columns correspond to Allegro with the index, middle, or ring
    finger removed and ShadowHand with the index, middle, ring, or little finger
    removed. From top to bottom, the rows show a baseball, pear, potted meat can,
    and water bottle.}
    \label{fig:zero_shot_qualitative}
\end{figure*}

Across all three embodiments, DexGraspFlow produces an executable grasp in 0.126--0.136~s (averaging 0.132~s including IK), reducing latency by approximately 49\% relative to TRO-Grasp (0.257~s) and requiring substantially less time than all competing baselines. DexGraspFlow generates link poses using only 10 flow-integration steps before recovering joint configurations through batched inverse kinematics. This pipeline avoids a separate contact-map or distance-based pose-reconstruction stage and achieves low generation latency while maintaining high grasp success. Fig.~\ref{fig:flow_trajectory} visualizes this process, showing smooth link-pose evolution from $t=0$ to $t=1$ before being resolved into executable joints via batched IK.

\subsection{Zero-shot generalization to topology-modified hands}\label{sec:zero_shot_generalization}

\textbf{Evaluation protocol.}
We evaluate seven hand variants absent from training: three Allegro variants with the index, middle, or ring finger removed, and four ShadowHand variants with the index, middle, ring, or little finger removed. Each variant is constructed by removing the complete kinematic chain and associated joints of the selected finger, yielding unseen hand configurations with altered link sets, actuated degrees of freedom, and kinematic-tree connectivity. For each modified hand, we recompute the link features and validity mask and rebuild the physical-link graph and topology-distance matrix from its modified URDF.

\begin{table}[!t]
    \centering
    \caption{Zero-shot unconditioned grasp success rates (\%)
    on topology-modified hands. I, M, R, and L denote the
    index, middle, ring, and little fingers, respectively.}
    \label{tab:zero_shot_topology}

    \footnotesize
    \setlength{\tabcolsep}{3pt}
    \renewcommand{\arraystretch}{1.15}

    \begin{tabular}{@{}lccc@{}}
        \toprule
        Hand variant & DRO-Grasp & TRO-Grasp & Ours \\
        \midrule
        Allegro w/o I
        & 53.60 & 11.80 & {\bfseries 97.00} \\
        Allegro w/o M
        & 61.40 & 45.80 & {\bfseries 75.10} \\
        Allegro w/o R
        & 49.90 & 65.40 & {\bfseries 94.50} \\
        \midrule
        ShadowHand w/o I
        & 47.80 & 2.50 & {\bfseries 89.00} \\
        ShadowHand w/o M
        & 55.40 & 4.00 & {\bfseries 90.60} \\
        ShadowHand w/o R
        & 57.90 & 7.20 & {\bfseries 82.80} \\
        ShadowHand w/o L
        & 45.80 & 6.60 & {\bfseries 92.20} \\
        \midrule
        Average
        & 53.11 & 20.47 & {\bfseries 88.74} \\
        \bottomrule
    \end{tabular}
\end{table}

All evaluated models are applied directly to these variants without retraining, fine-tuning, or embodiment-specific adaptation. Each variant is evaluated on the 10 held-out objects using 100 candidates per object, resulting in 7,000 simulated grasp trials per method. Generation is unconditioned, and DexGraspFlow uses the default sampling and Pyroki IK settings. All grasps are evaluated using the same Isaac Gym success criterion.

Figure~\ref{fig:topology_variants} shows the intact and modified hand embodiments. The abbreviations I, M, R, and L denote the index, middle, ring, and little fingers, respectively, and are used consistently in Table~\ref{tab:zero_shot_topology}. Figure~\ref{fig:zero_shot_qualitative} presents representative grasps on the modified hands.

\textbf{Overall performance.}
As shown in Table~\ref{tab:zero_shot_topology}, DexGraspFlow achieves the highest success rate on all seven variants, with an average of 88.74\%, compared with 53.11\% for DRO-Grasp and 20.47\% for TRO-Grasp. Its success rates range from 75.10\% to 97.00\%, indicating that performance varies with the removed finger but remains higher than the evaluated baselines across all configurations. The link-level formulation represents the remaining links using the same token construction, while the recomputed topology distances describe their kinematic relationships in the modified hand. These inputs allow the model to incorporate changes in both the link set and structural connectivity without changing the representation scheme. The results support zero-shot grasp generation on the evaluated topology-modified hands.

\subsection{Palm-orientation-conditioned grasp generation}\label{sec:conditioned_generation}

\textbf{Evaluation protocol.}
We evaluate palm-orientation control using the guided sampling procedure in Sec.~\ref{sec:conditioned_inference}. Target orientations are constructed from six canonical directions in the object-centric frame: \(+x,-x,+y,-y,+z,\) and \(-z\), referred to as front, back, left, right, up, and down, respectively. For each direction, the target palm approach axis is aligned with the specified direction. Following TRO-Grasp~\citep{tro}, target orientations \(\mathbf{R}_p^c \in \mathrm{SO}(3)\) are canonically constructed as full rigid rotations to ensure an identical protocol comparison. Sampling starts at \(t=0\), using the same translational initialization as unconditioned generation and the direction-aligned, link-wise rotational initialization described in Sec.~\ref{sec:conditioned_inference}. Sampling is guided toward the target orientation through inference-time correction without providing a reference grasp pose.

\begin{table}[!t]
    \centering
    \caption{Six-direction conditioned grasp generation results. TRO and Ours refer to TRO-Grasp and DexGraspFlow, respectively. SR and POE denote success rate and final palm-orientation error, respectively. $\Delta$SR denotes the success-rate difference in percentage points, while $\Delta\mathrm{POE}=\mathrm{POE}_{\mathrm{TRO}}-\mathrm{POE}_{\mathrm{Ours}}$ denotes the absolute POE reduction in degrees; positive values favor DexGraspFlow.}
    \label{tab:conditioned_generation}
    \renewcommand{\arraystretch}{1.15}
    \resizebox{\columnwidth}{!}{%
    \setlength{\tabcolsep}{3.5pt}
    \begin{tabular}{l|ccc|ccc}
        \hline
        \multirow{2}{*}{Setting}
        & \multicolumn{3}{c|}{Success Rate (\%)}
        & \multicolumn{3}{c}{POE (deg.)} \\
        \cline{2-7}
        & TRO & Ours & $\Delta$SR$\uparrow$ & TRO & Ours & $\Delta\mathrm{POE}$ \\
        \hline
        \multicolumn{7}{l}{\textit{Hand-wise results}} \\
        Allegro    & 80.42 & {\bfseries 95.17} & {\bfseries +14.75} & 38.25 & {\bfseries 15.18} & {\bfseries +23.07} \\
        Barrett    & 77.17 & {\bfseries 83.50} & {\bfseries +6.33}  & 41.40 & {\bfseries 9.12}  & {\bfseries +32.28} \\
        ShadowHand & 63.67 & {\bfseries 81.42} & {\bfseries +17.75} & 35.51 & {\bfseries 17.46} & {\bfseries +18.05} \\
        Average    & 73.75 & {\bfseries 86.69} & {\bfseries +12.94} & 38.38 & {\bfseries 13.92} & {\bfseries +24.46} \\
        \hline
        \multicolumn{7}{l}{\textit{Direction-wise results averaged across hand embodiments}} \\
        Front & 71.33 & {\bfseries 90.50} & {\bfseries +19.17} & 39.44 & {\bfseries 13.95} & {\bfseries +25.49} \\
        Back  & 73.67 & {\bfseries 82.67} & {\bfseries +9.00}  & 37.95 & {\bfseries 14.16} & {\bfseries +23.79} \\
        Left  & 73.67 & {\bfseries 88.00} & {\bfseries +14.33} & 37.88 & {\bfseries 14.24} & {\bfseries +23.64} \\
        Right & 75.50 & {\bfseries 83.17} & {\bfseries +7.67}  & 37.66 & {\bfseries 14.64} & {\bfseries +23.02} \\
        Up    & 80.00 & {\bfseries 91.50} & {\bfseries +11.50} & 32.37 & {\bfseries 13.24} & {\bfseries +19.13} \\
        Down  & 68.33 & {\bfseries 84.33} & {\bfseries +16.00} & 45.00 & {\bfseries 13.29} & {\bfseries +31.71} \\
        \hline
    \end{tabular}%
    }
\end{table}

We generate 20 candidates per direction for each object--hand pair, yielding 120 candidates per pair, 1,200 per hand, and 3,600 in total. DexGraspFlow uses the default conditioned inference settings with \(s_0=1.0\) and \(\lambda_c=20\). TRO-Grasp is evaluated using its palm-conditioned inference procedure with the same target directions, test objects, random seed, candidate counts, and physical success criterion.

In addition to SR, we report palm-orientation error in degrees, defined as $\mathrm{POE} = \frac{180}{\pi} \|\operatorname{Log}(\mathbf{R}_{p}^{c}\hat{\mathbf{R}}_{p}^{\top})\|_2$, where \(\mathbf{R}_p^c\) is the target palm orientation and \(\hat{\mathbf{R}}_p\) is obtained by forward kinematics from the final executable joint configuration.

\begin{figure*}[!t]
    \centering
    \footnotesize
    \setlength{\tabcolsep}{3pt}
    \def\conimagewidth{0.145\textwidth}
    \def\conlabelwidth{0.065\textwidth}
    \setlength{\tabcolsep}{2pt}
    \begin{tabular}{@{}p{\conlabelwidth}@{\hspace{2pt}}cccccc@{}}
        & \scriptsize Front & \scriptsize Back & \scriptsize Left
        & \scriptsize Right & \scriptsize Up & \scriptsize Down \\
        \raisebox{0.065\textwidth}[0pt][0pt]{\makebox[\conlabelwidth][r]{\scriptsize\textbf{Allegro}}} &
        \includegraphics[width=\conimagewidth,height=\conimagewidth,keepaspectratio]{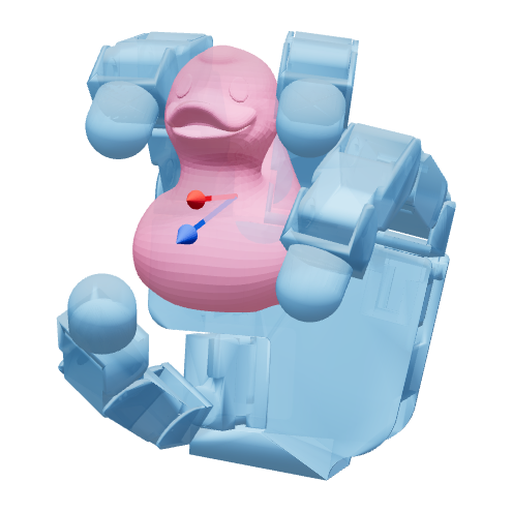} &
        \includegraphics[width=\conimagewidth,height=\conimagewidth,keepaspectratio]{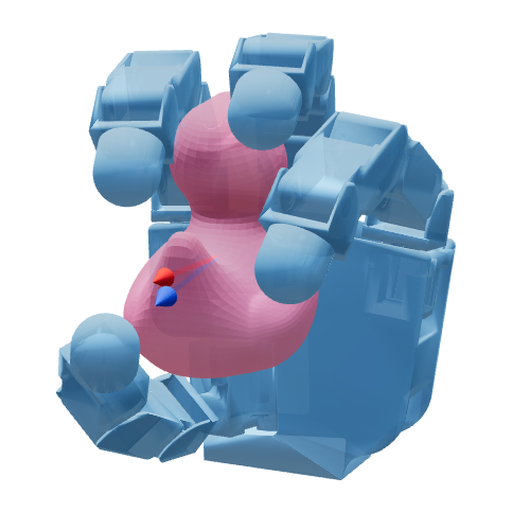} &
        \includegraphics[width=\conimagewidth,height=\conimagewidth,keepaspectratio]{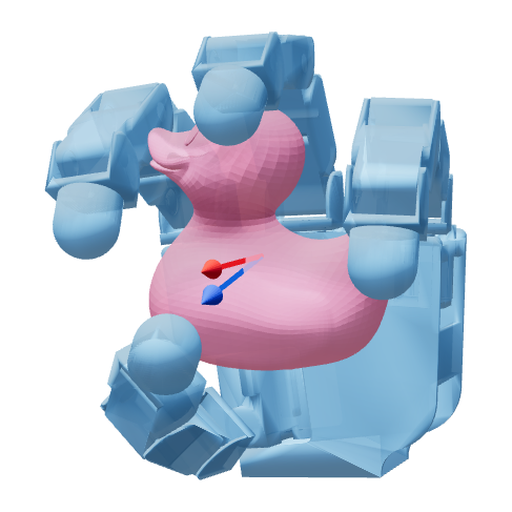} &
        \includegraphics[width=\conimagewidth,height=\conimagewidth,keepaspectratio]{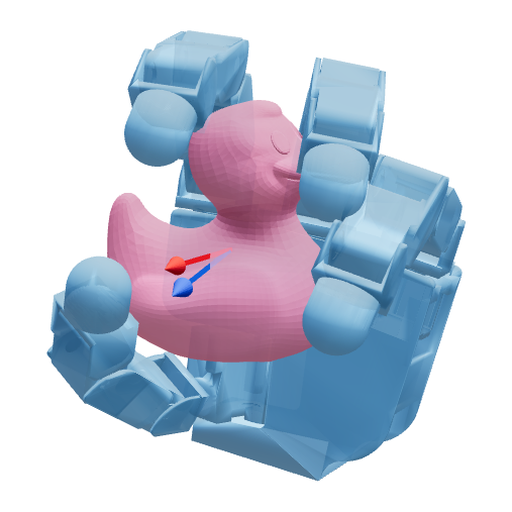} &
        \includegraphics[width=\conimagewidth,height=\conimagewidth,keepaspectratio]{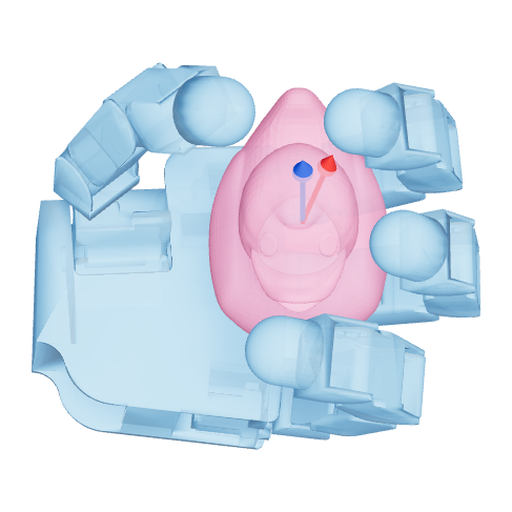} &
        \includegraphics[width=\conimagewidth,height=\conimagewidth,keepaspectratio]{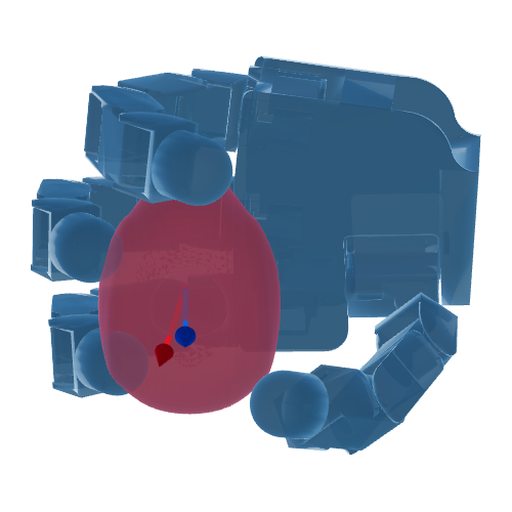} \\[1.5mm]
        \raisebox{0.065\textwidth}[0pt][0pt]{\makebox[\conlabelwidth][r]{\scriptsize\textbf{Barrett}}} &
        \includegraphics[width=\conimagewidth,height=\conimagewidth,keepaspectratio]{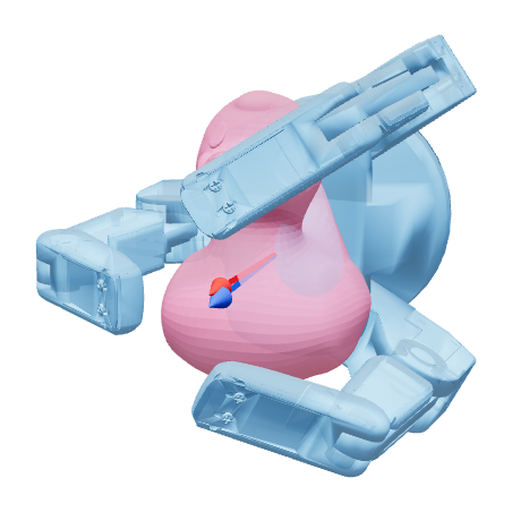} &
        \includegraphics[width=\conimagewidth,height=\conimagewidth,keepaspectratio]{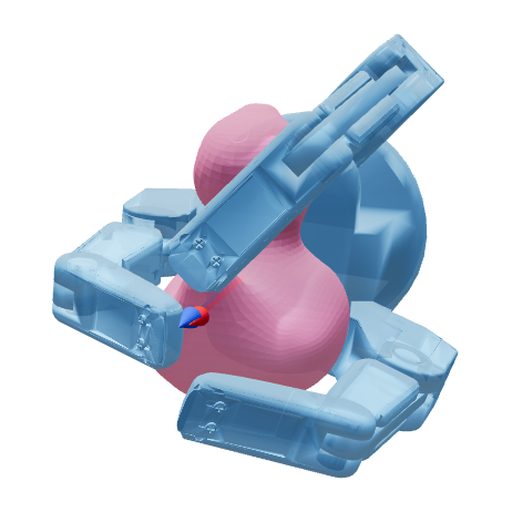} &
        \includegraphics[width=\conimagewidth,height=\conimagewidth,keepaspectratio]{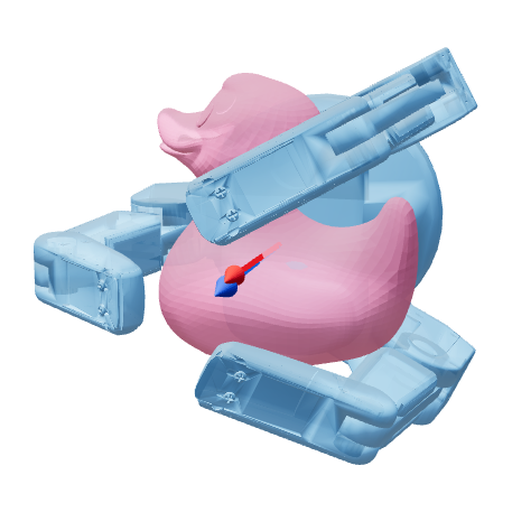} &
        \includegraphics[width=\conimagewidth,height=\conimagewidth,keepaspectratio]{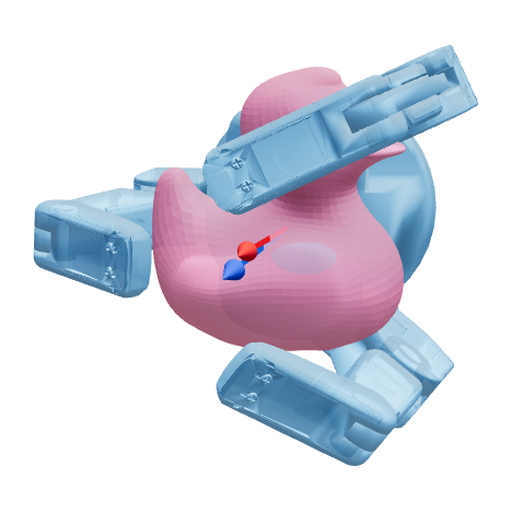} &
        \includegraphics[width=\conimagewidth,height=\conimagewidth,keepaspectratio]{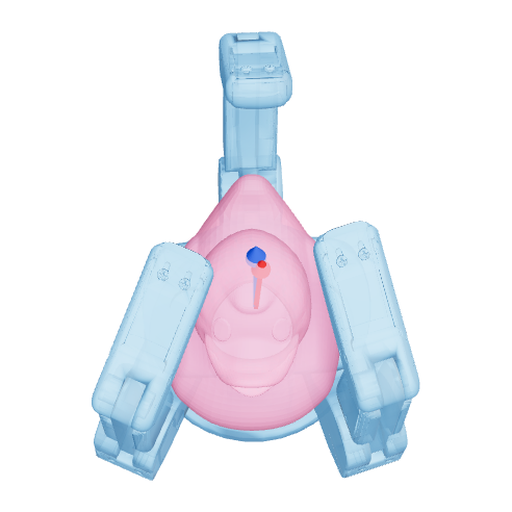} &
        \includegraphics[width=\conimagewidth,height=\conimagewidth,keepaspectratio]{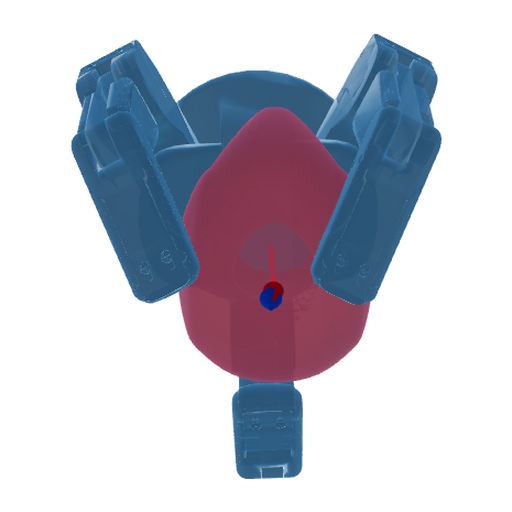} \\[1.5mm]
        \raisebox{0.060\textwidth}[0pt][0pt]{\makebox[\conlabelwidth][r]{\scriptsize\shortstack[r]{\textbf{Shadow}\\[1pt]\textbf{Hand}}}} &
        \includegraphics[width=\conimagewidth,height=\conimagewidth,keepaspectratio]{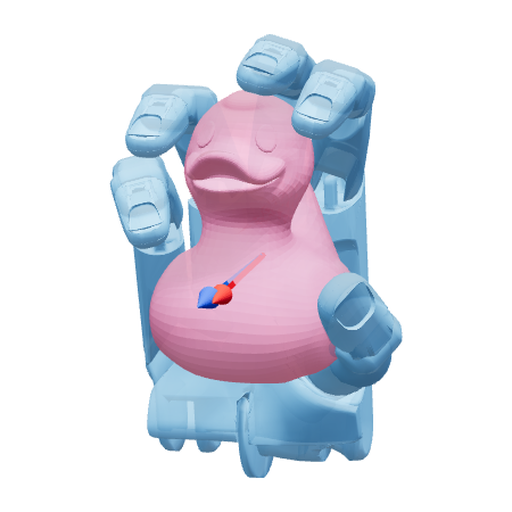} &
        \includegraphics[width=\conimagewidth,height=\conimagewidth,keepaspectratio]{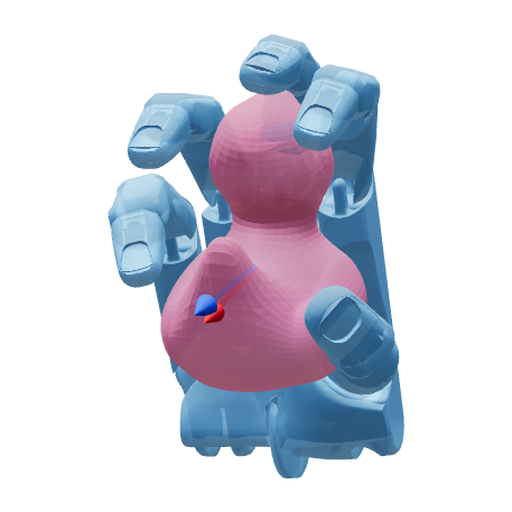} &
        \includegraphics[width=\conimagewidth,height=\conimagewidth,keepaspectratio]{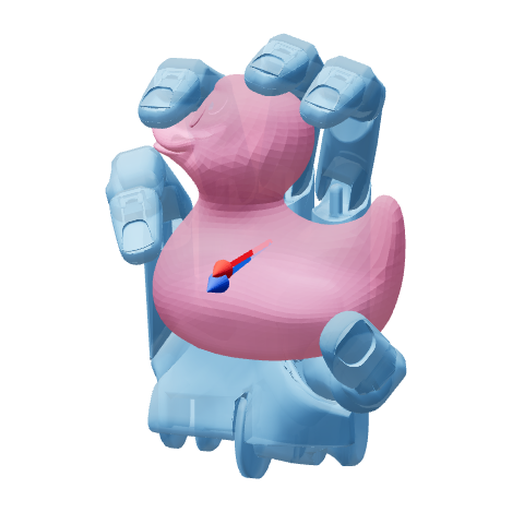} &
        \includegraphics[width=\conimagewidth,height=\conimagewidth,keepaspectratio]{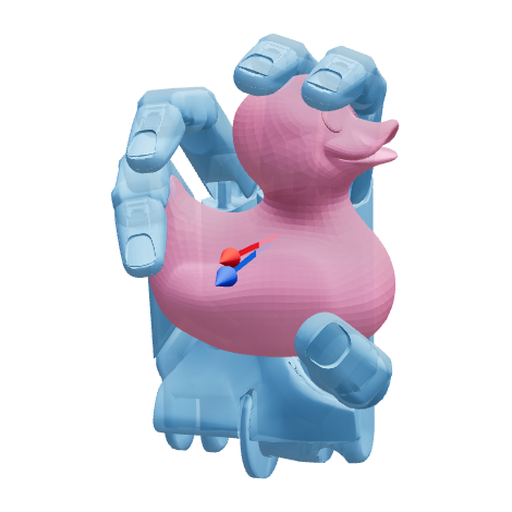} &
        \includegraphics[width=\conimagewidth,height=\conimagewidth,keepaspectratio]{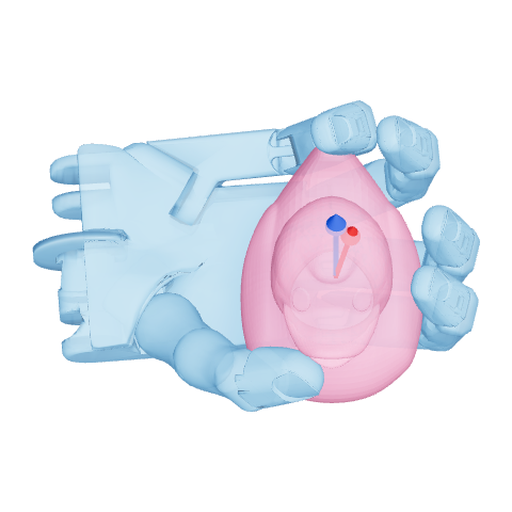} &
        \includegraphics[width=\conimagewidth,height=\conimagewidth,keepaspectratio]{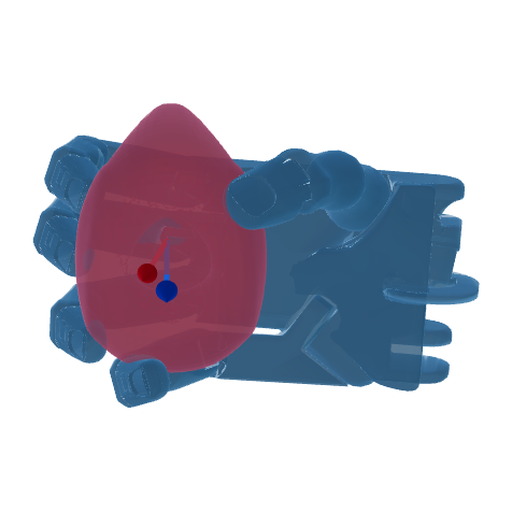}
    \end{tabular}

    \caption{Palm-orientation-conditioned grasp examples for a representative
    held-out rubber duck. The rows correspond to Allegro, Barrett, and
    ShadowHand, while the columns show the front, back, left, right, up, and
    down target directions. Blue and red arrows denote the target and post-IK
    actual palm approach directions, respectively.}
    \label{fig:conditioned_grasps}
\end{figure*}

\textbf{Overall performance.}
Table~\ref{tab:conditioned_generation} reports hand-wise and direction-wise results. DexGraspFlow improves the average SR from 73.75\% to 86.69\% while reducing mean final POE from 38.38 to 13.92 degrees. Improvements in both metrics are observed across all three hand embodiments and all six target directions.

Direction-wise POE remains within 13.24--14.64 degrees for DexGraspFlow, whereas TRO-Grasp exhibits larger variation and reaches 45.00 degrees for the downward direction. The orientation correction steers sampling toward the requested palm pose, while the learned vector field continues to update the link states using hand and object information. The improvements in both SR and POE show that the guided generation procedure can satisfy the orientation requirement more accurately while retaining higher grasp success under the evaluated protocol. Figure~\ref{fig:conditioned_grasps} illustrates representative grasps for the held-out rubber duck, with rows corresponding to hand embodiments and columns to target directions.

\subsection{Ablation studies}\label{sec:ablation_study}

We examine the effects of attention biases, ODE integration steps, and palm-orientation guidance strength. The first two studies use unconditioned generation, while the guidance-strength study follows the conditioned protocol. Unless otherwise specified, all experiments use the settings in Sec.~\ref{sec:experimental_setup}.

\begin{table}[!t]
    \centering
    \caption{Ablation of structural attention biases under unconditioned generation. All values denote success rate (\%).}
    \label{tab:attention_bias_ablation}
    \footnotesize
    \setlength{\tabcolsep}{1.6pt}
    \begin{tabular}{lcccc}
        \hline
        Bias Configuration & Barrett & Allegro & ShadowHand & Avg. \\
        \hline
        No Attention Bias   & 86.10 & 98.30 & 75.50 & 86.63 \\
        Geometry Bias Only  & 87.50 & 99.50 & 82.10 & 89.70 \\
        Topology Bias Only  & 88.70 & 99.50 & 90.00 & 92.73 \\
        Geometry + Topology & 92.10 & 98.40 & 95.50 & 95.33 \\
        \hline
    \end{tabular}
\end{table}

\textbf{Effect of attention biases.}
We evaluate four configurations by enabling or disabling the geometric biases in both attention streams and the topology bias in the link--link stream. \textit{No Attention Bias} disables both types; \textit{Geometry Bias Only} retains geometric biases in both streams; \textit{Topology Bias Only} retains the link--link topology bias; and \textit{Geometry + Topology} uses the complete model. Relation-conditioned value messages and the gates of active bias branches are retained in all configurations, isolating the contribution of attention-logit biases.

Table~\ref{tab:attention_bias_ablation} shows that removing both biases reduces average SR to 86.63\%. Geometry bias alone raises it to 89.70\%, while topology bias alone reaches 92.73\%. The topology-bias gain is particularly pronounced on ShadowHand, where SR increases from 75.50\% to 90.00\%.

Combining both biases yields the highest average SR of 95.33\% and the best results on Barrett and ShadowHand. Although the single-bias variants perform slightly better on Allegro, the combined configuration provides the best overall performance across embodiments. Geometric biases provide spatial context for link--link and link--object interactions, whereas topology bias distinguishes links according to their relationships in the kinematic tree. Spatial proximity alone does not determine these structural relationships. The highest average SR obtained with both biases therefore supports the benefit of combining geometric and kinematic information during grasp generation.

\begin{table}[!t]
    \centering
    \caption{Effect of ODE integration steps on unconditioned grasp success rate and batch-amortized latency.}
    \label{tab:ode_step_ablation}
    \footnotesize
    \setlength{\tabcolsep}{3.0pt}
    \begin{tabular}{c|cccc|c}
        \hline
        \multirow{2}{*}{Steps}
        & \multicolumn{4}{c|}{Success Rate (\%)}
        & \multirow{2}{*}{Time (s) $\downarrow$} \\
        \cline{2-5}
        & Barrett & Allegro & ShadowHand & Avg. & \\
        \hline
        1  & 50.90 & 60.30 & 76.20 & 62.47 & 0.0130 \\
        5  & 93.40 & 98.10 & 92.60 & 94.70 & 0.0205 \\
        10 & 92.10 & 98.40 & 95.50 & 95.33 & 0.0257 \\
        15 & 91.70 & 97.80 & 93.50 & 94.33 & 0.0321 \\
        20 & 92.90 & 97.90 & 94.80 & 95.20 & 0.0361 \\
        \hline
    \end{tabular}
\end{table}

\textbf{Effect of ODE integration steps.}
We vary the number of Euler integration steps $S$ over \(\{1,5,10,15,20\}\), using the same trained checkpoint and identical initial samples. Table~\ref{tab:ode_step_ablation} reports SR and batch-amortized end-to-end latency with a batch size of 100, including model inference and Pyroki IK.

Increasing the number of steps from one to five raises average SR from 62.47\% to 94.70\%. Multiple integration steps update the velocity prediction along the evolving state, providing a finer approximation of the learned flow trajectory. However, the small and non-monotonic SR differences beyond five steps show that finer integration does not necessarily produce more successful grasps, while latency continues to increase. The 10-step setting achieves the highest average SR of 95.33\% at 0.0257~s per grasp, whereas 20 steps yield 95.20\% at 0.0361~s. We therefore adopt 10 steps as the default balance between grasp quality and generation cost.

\textbf{Effect of palm-orientation guidance strength.}
We vary \(\lambda_c\) over \(\{5,10,20,30\}\), keeping the checkpoint, initial sampling, target directions, candidate counts, ODE solver, \(s_0=1.0\), and IK settings fixed. Table~\ref{tab:guidance_scale} reports SR and mean final POE over all candidates.

\begin{table}[!t]
    \centering
    \caption{Sensitivity to rotational guidance multiplier $\lambda_c$ under six-direction conditioning. Bold indicates default setting.}
    \label{tab:guidance_scale}
    \footnotesize
    \setlength{\tabcolsep}{3.0pt}
    \begin{tabular}{c|cc}
        \hline
        Rotational Guidance $\lambda_c$
        & Avg. SR (\%) & Avg. POE (deg.) \\
        \hline
        $5$             & 89.08 & 22.135 \\
        $10$            & 87.33 & 17.552 \\
        $20$            & {\bfseries 86.69} & {\bfseries 13.921} \\
        $30$            & 83.50 & 12.225 \\
        \hline
    \end{tabular}
\end{table}

Increasing $\lambda_c$ gives greater weight to orientation correction over learned rotational velocity, reducing mean POE from 22.135 to 12.225 degrees while lowering average SR from 89.08\% to 83.50\% (Table~\ref{tab:guidance_scale}). This highlights a trade-off between condition satisfaction and grasp stability, leading us to select $\lambda_c=20$ as the default balance.

\subsection{Real-robot experiments}\label{sec:real_robot}

\begin{figure}[!t]
    \centering
    \subfloat[Robot workspace]{%
        \includegraphics[width=0.48\columnwidth,trim=35 170 20 70,clip]{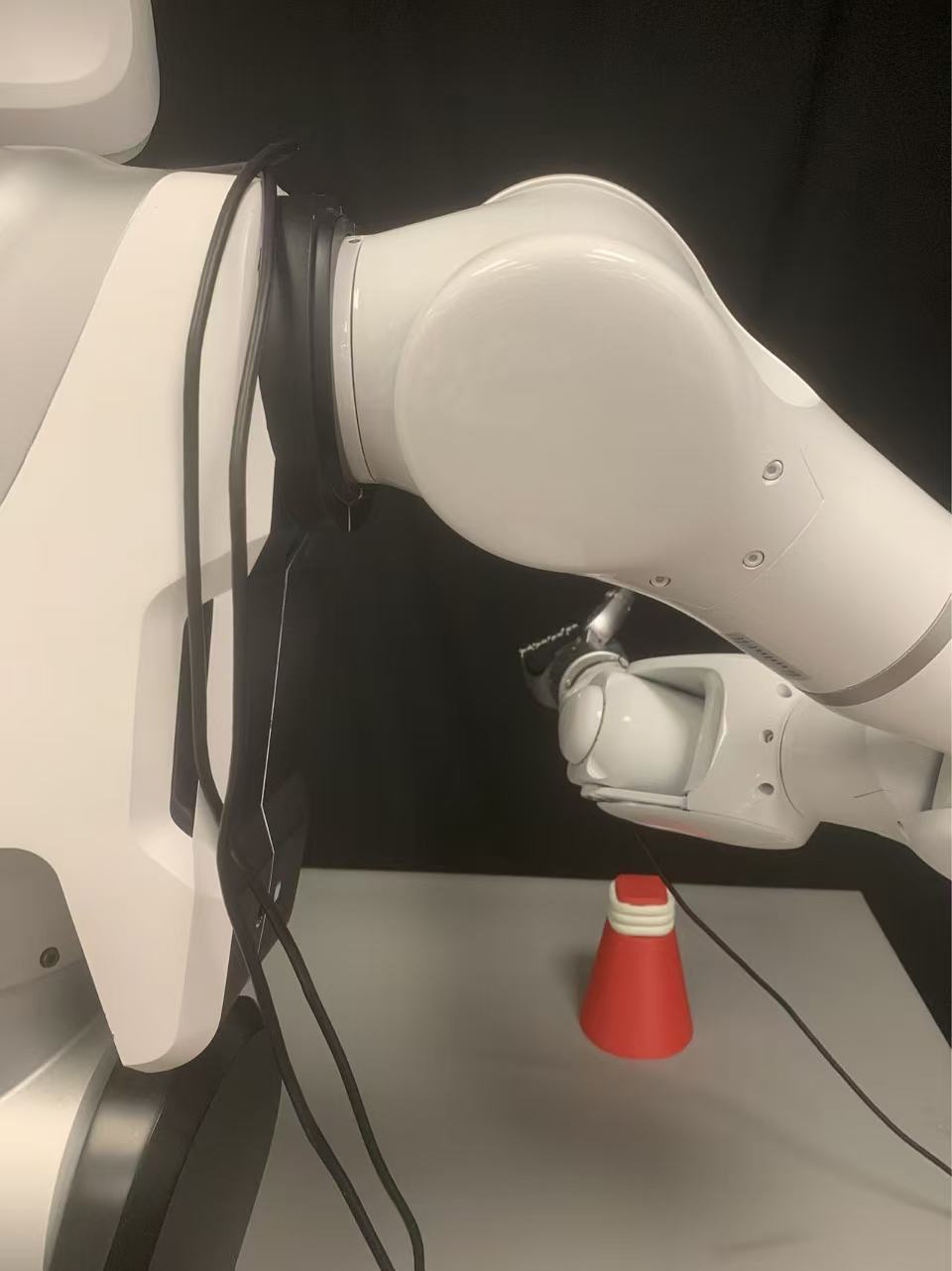}}\hfill
    \subfloat[Hand--object interaction]{%
        \includegraphics[width=0.48\columnwidth,trim=35 170 20 70,clip]{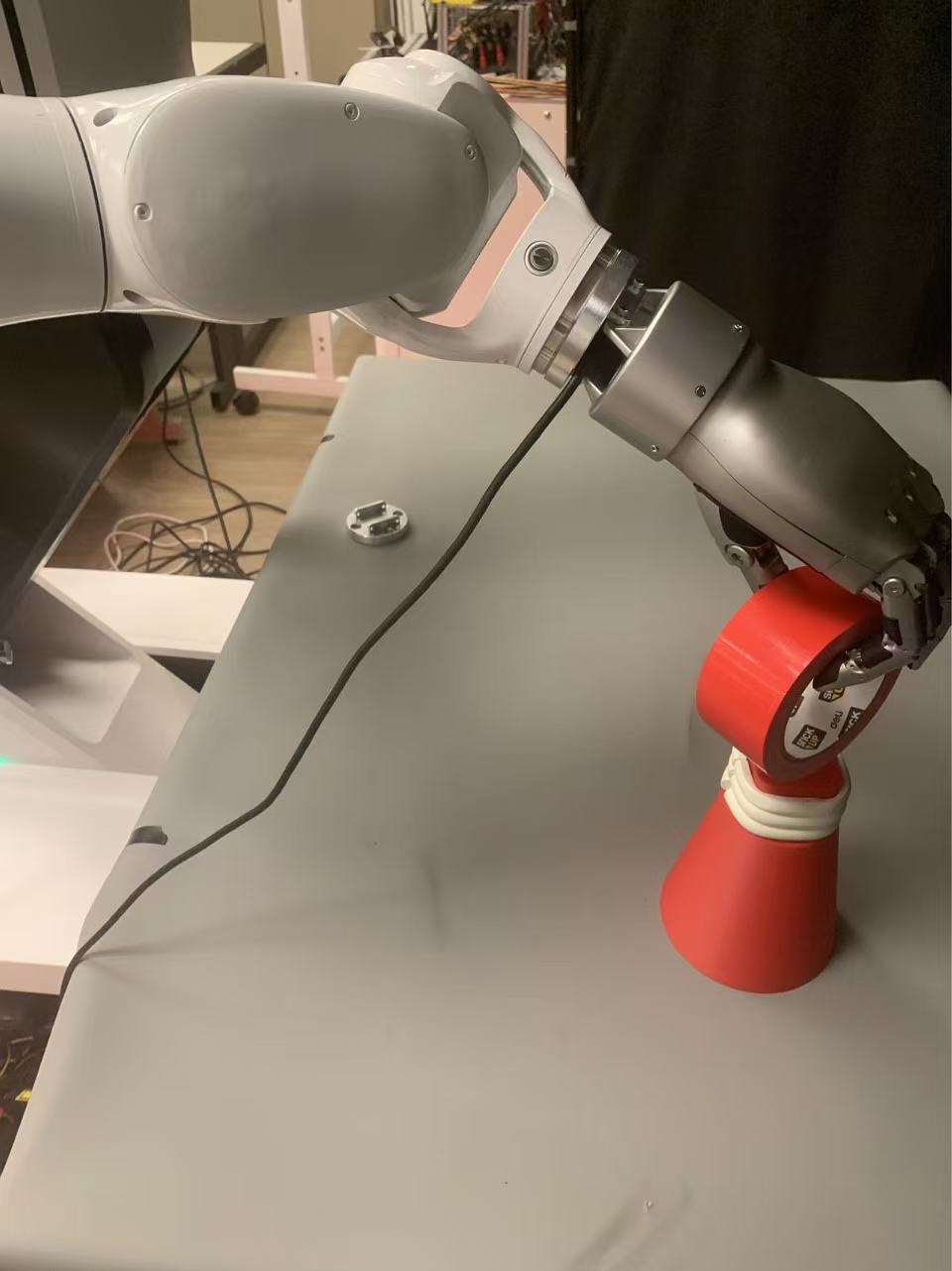}}
    \caption{Real-robot experimental setup. (a) Robot workspace view. (b) Close-up of the hand--object interaction.}
    \label{fig:real_robot_setup}
\end{figure}

\textbf{Experimental setup and protocol.}
We conduct real-robot experiments using the right arm of a ROKAE Wheeled Dual-Arm Robot equipped with a Linker Hand L20Lite, as shown in Fig.~\ref{fig:real_robot_setup}. An Intel RealSense D455 captures RGB-D observations and object point clouds, while FoundationPose~\citep{foundationpose} estimates target-object poses. DexGraspFlow is trained on 13,566 Linker Hand L20Lite grasp demonstrations and evaluated on 10 physical objects unseen during training.

\begin{figure*}[!t]
    \centering
    \subfloat[Can]{%
        \includegraphics[width=0.19\textwidth]{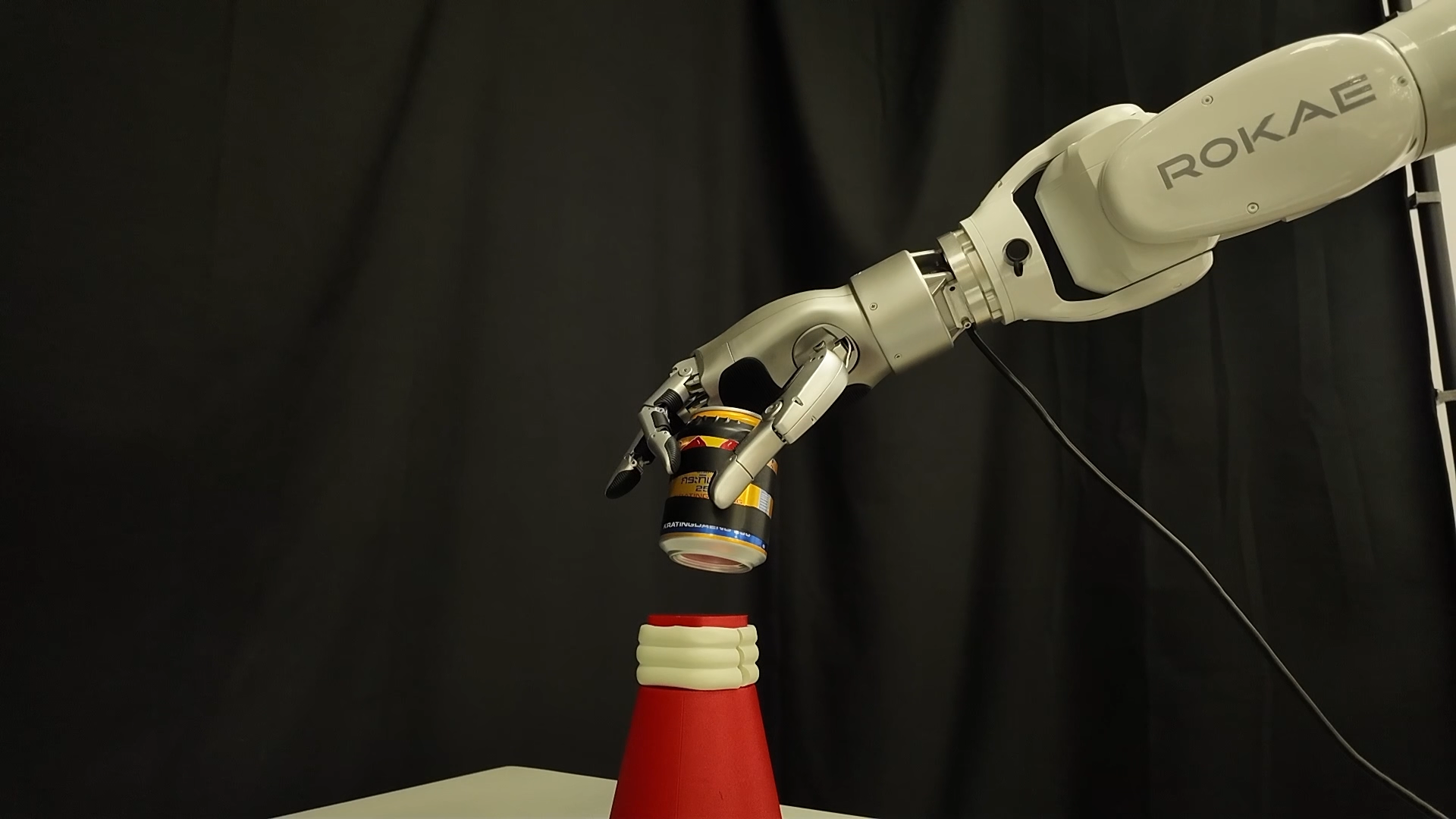}}\hfill
    \subfloat[Coil]{%
        \includegraphics[width=0.19\textwidth]{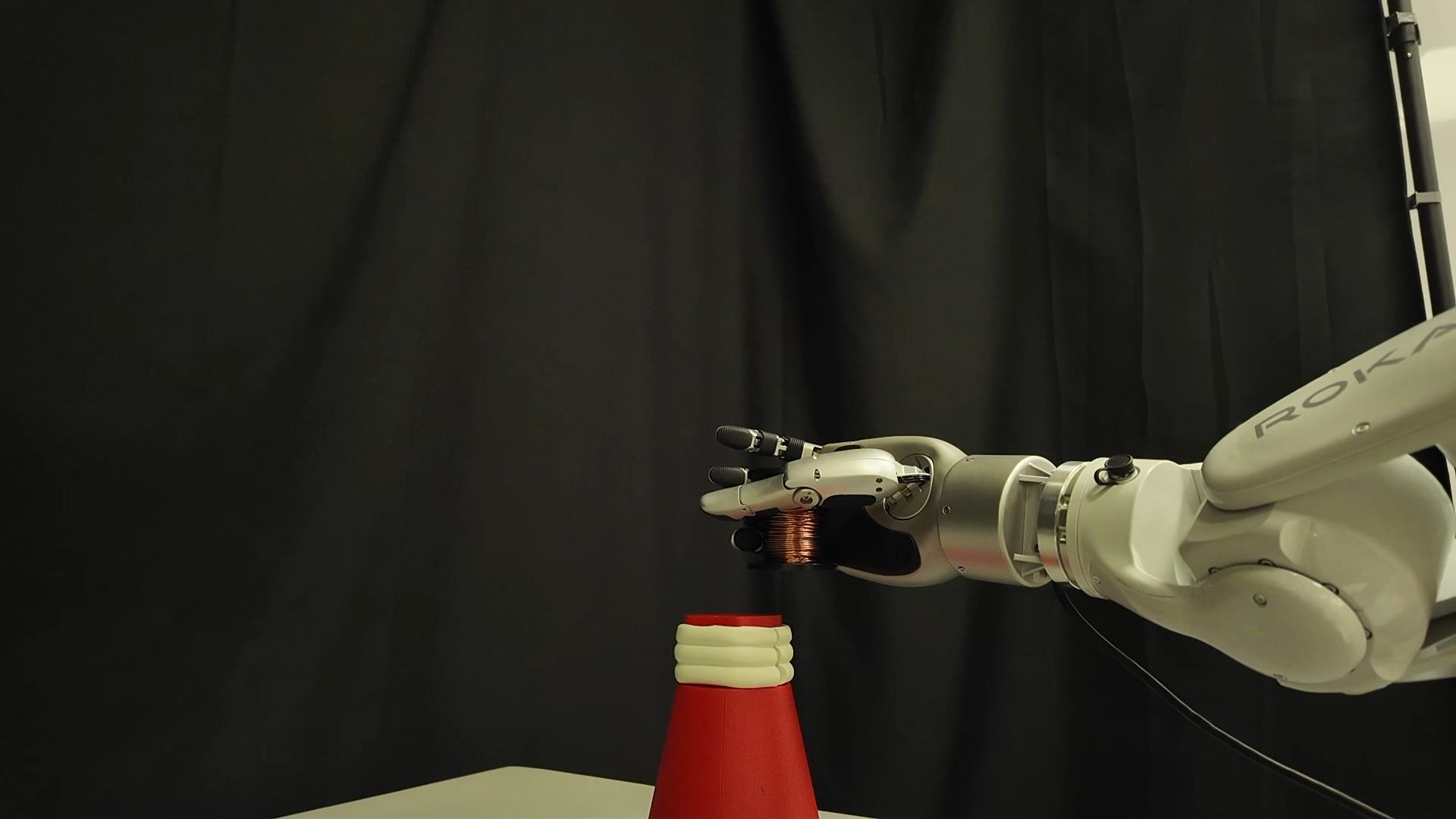}}\hfill
    \subfloat[Toy]{%
        \includegraphics[width=0.19\textwidth]{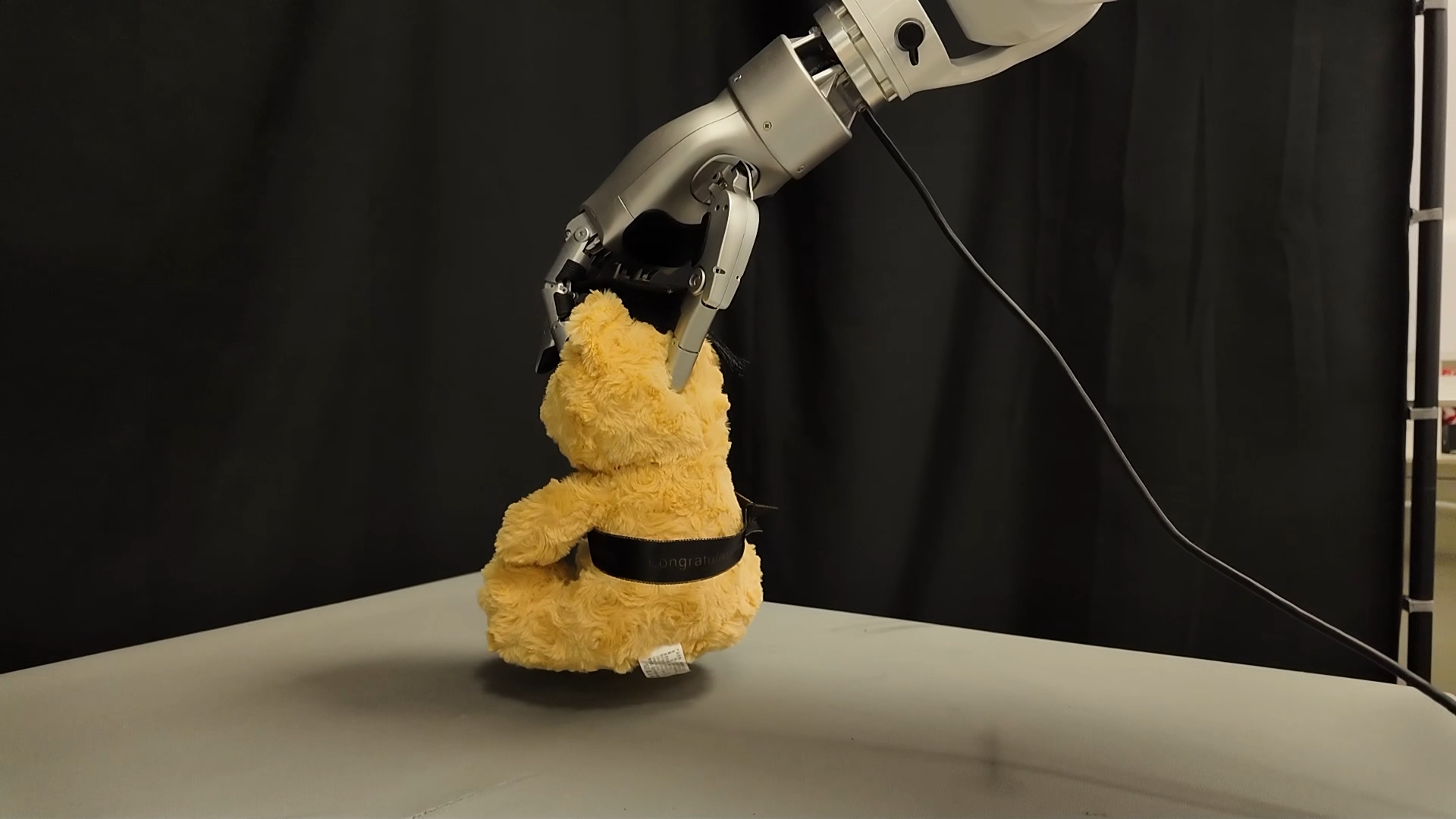}}\hfill
    \subfloat[Foam Box Lid]{%
        \includegraphics[width=0.19\textwidth]{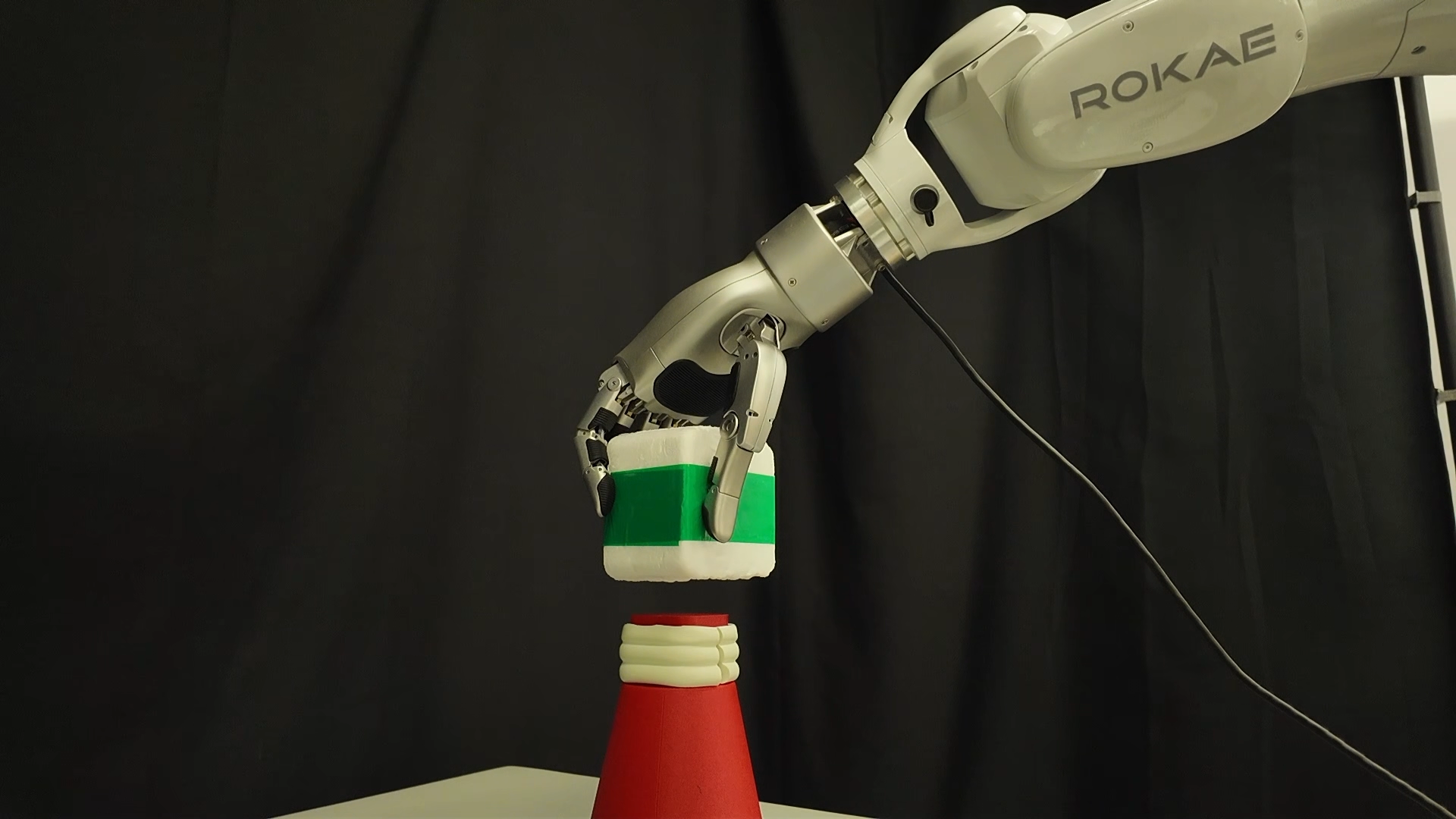}}\hfill
    \subfloat[McCafe]{%
        \includegraphics[width=0.19\textwidth]{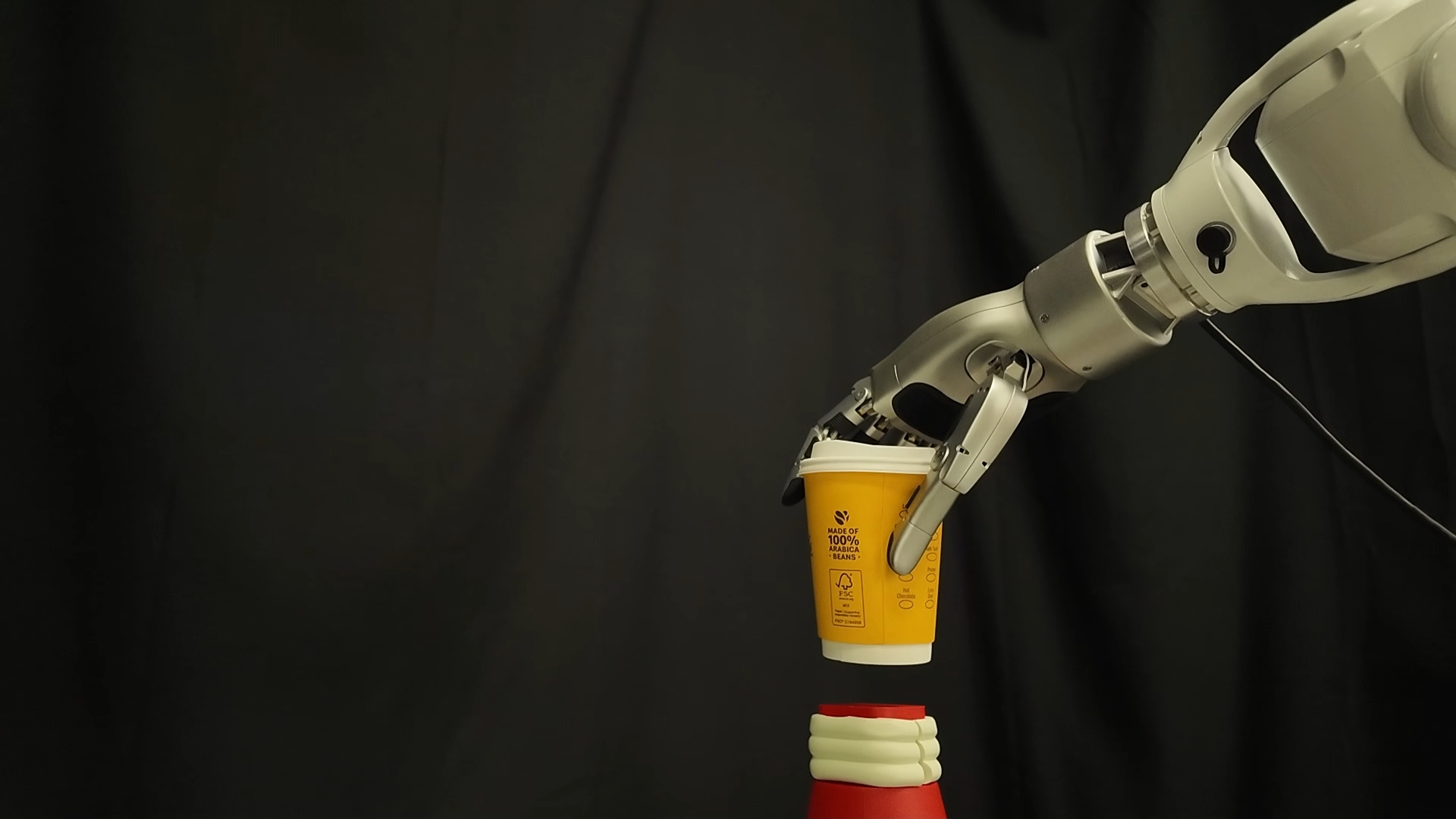}}\\[1mm]
    \subfloat[Orange]{%
        \includegraphics[width=0.19\textwidth]{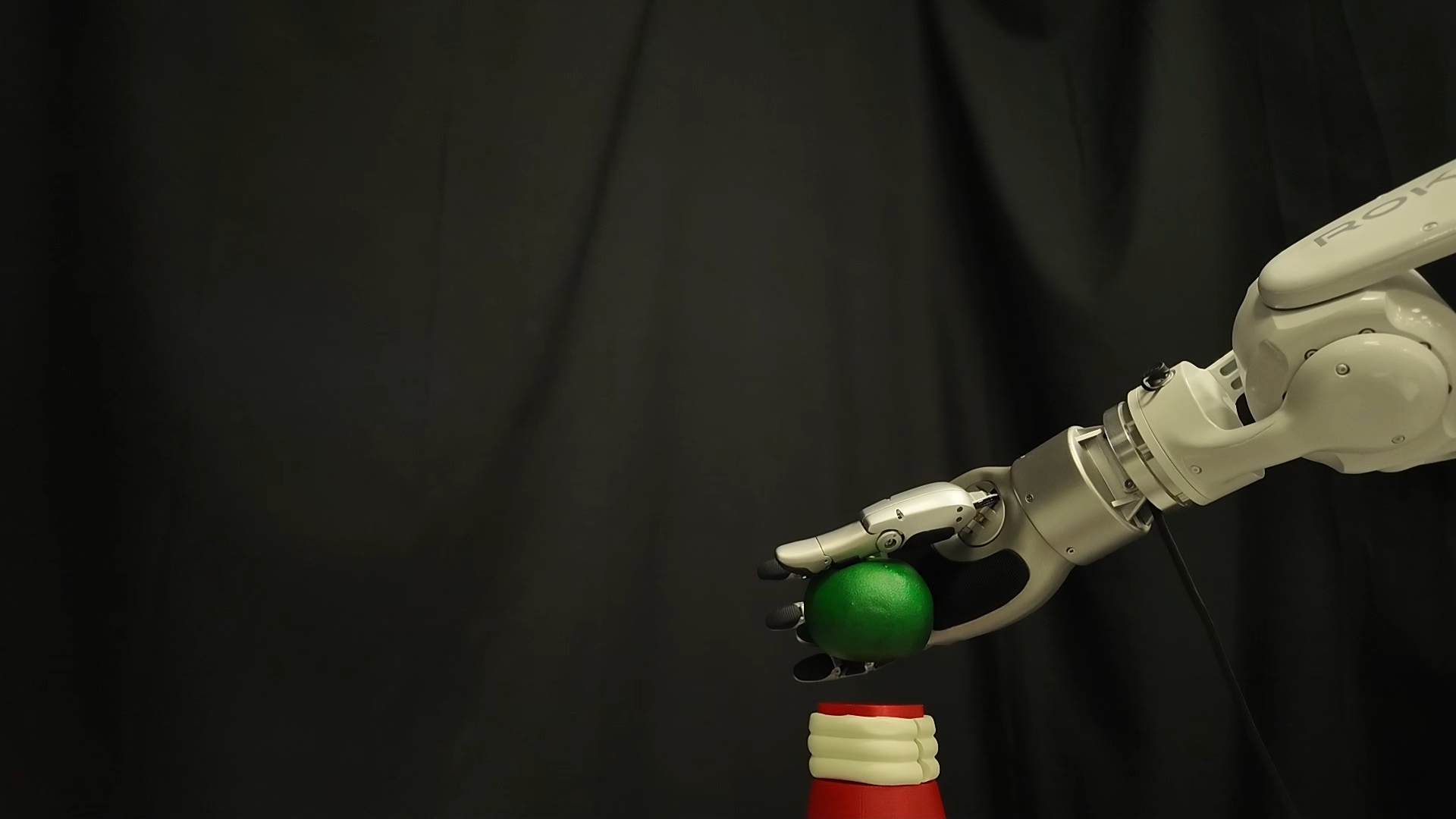}}\hfill
    \subfloat[Packaging Box]{%
        \includegraphics[width=0.19\textwidth]{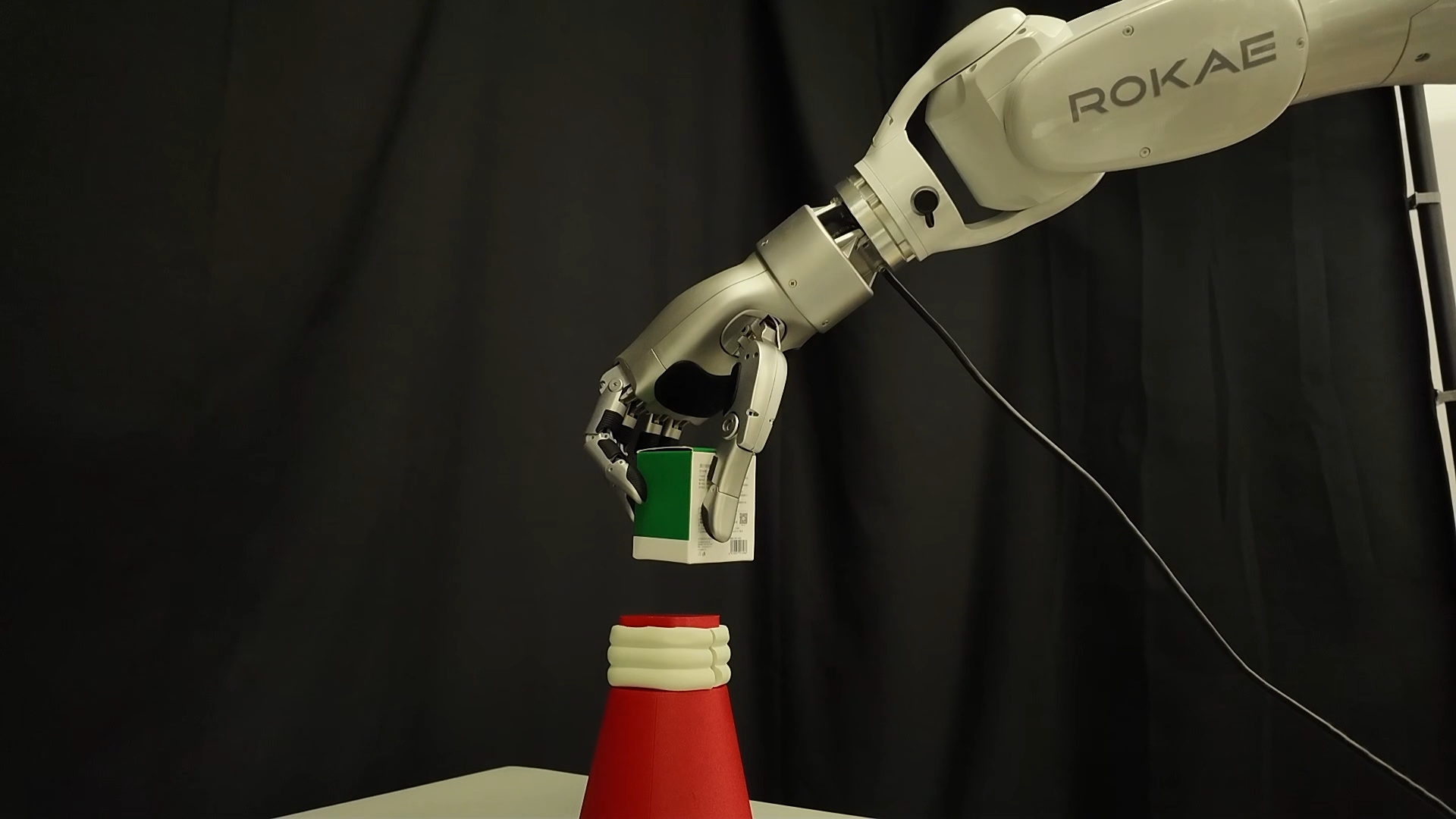}}\hfill
    \subfloat[Pill Bottle]{%
        \includegraphics[width=0.19\textwidth]{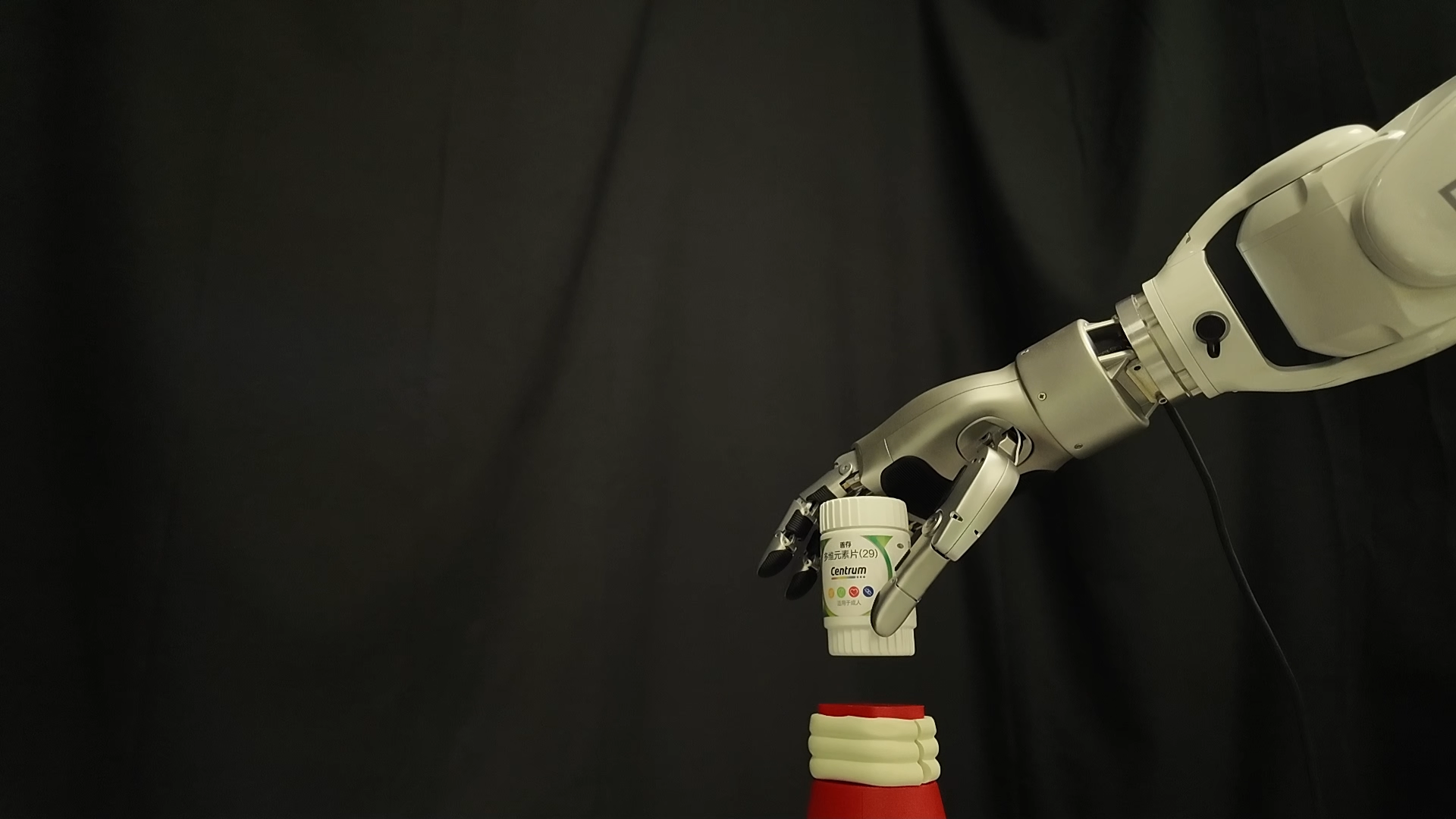}}\hfill
    \subfloat[Pomegranate]{%
        \includegraphics[width=0.19\textwidth]{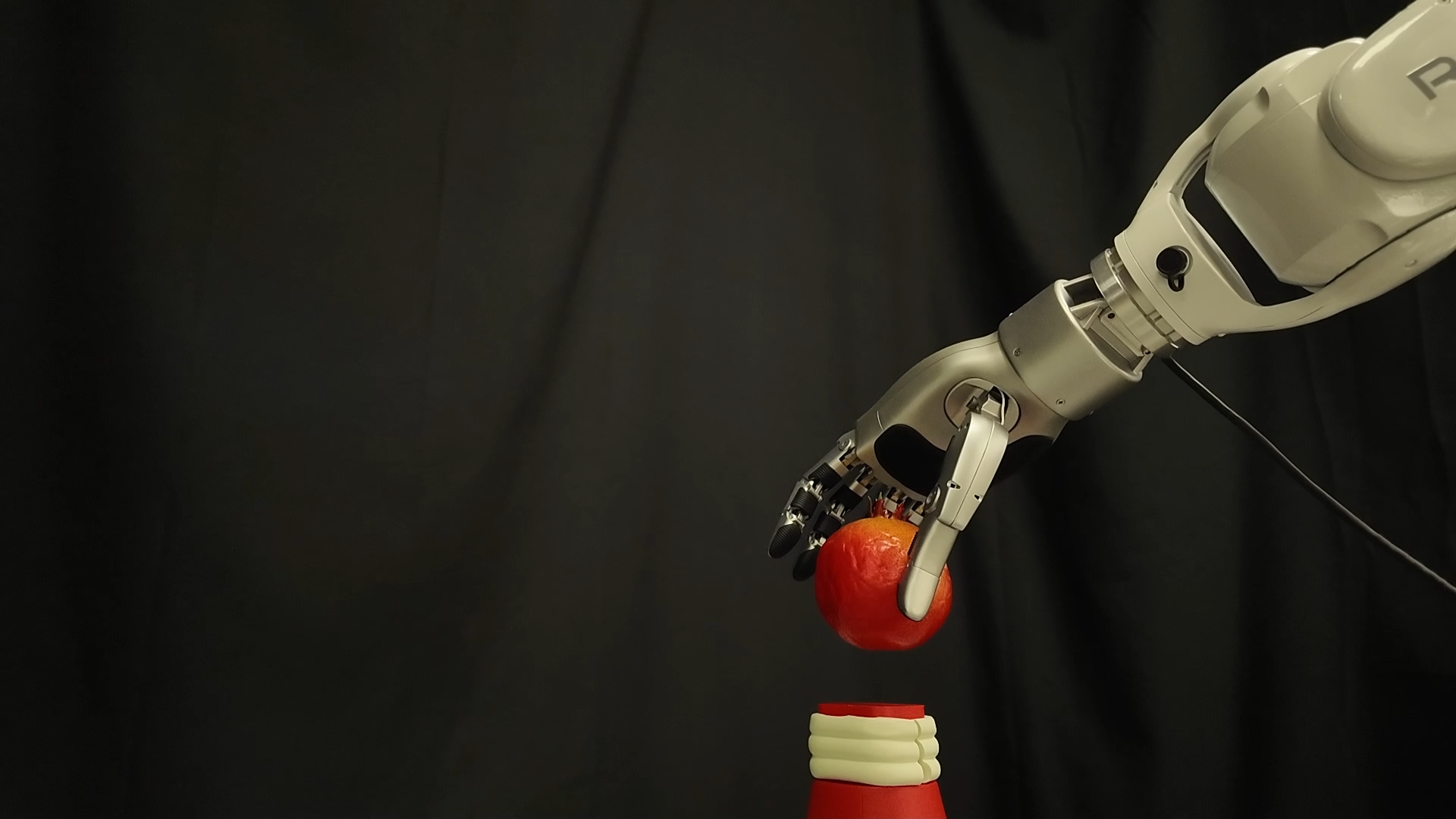}}\hfill
    \subfloat[Red Tape]{%
        \includegraphics[width=0.19\textwidth]{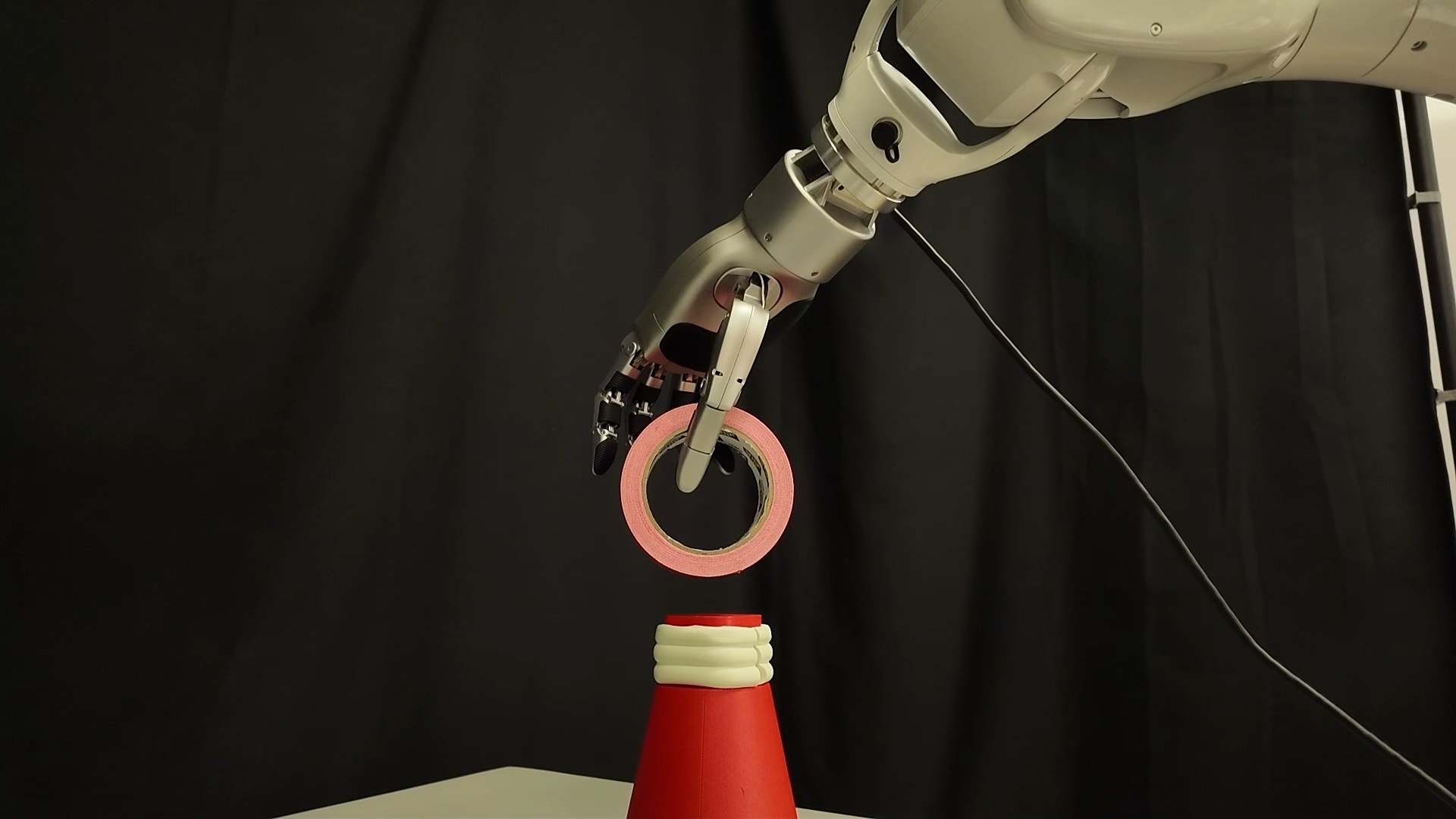}}
    \caption{Real-robot grasping examples generated by DexGraspFlow for ten objects unseen during training using the Linker Hand L20Lite.}
    \label{fig:real_robot_grasp_results}
\end{figure*}

Generated link poses are converted into joint configurations through Pyroki IK and are mapped to hand actuator commands. To ensure firm contact and reduce slippage on smooth surfaces, a fixed joint offset toward closure is uniformly applied to the finger flexion joints before lifting. We conduct 10 trials per object for a total of 100 attempts. A trial is successful if the robot lifts the object from the supporting surface and maintains a stable grasp throughout the lifting motion.

\begin{table}[!t]
    \centering
    \caption{Real-robot grasping results on 10 unseen objects.}
    \label{tab:real_robot_results}
    \resizebox{\columnwidth}{!}{%
    \footnotesize
    \setlength{\tabcolsep}{3.5pt}
    \begin{tabular}{lc|lc}
        \hline
        Object & Success/Trials & Object & Success/Trials \\
        \hline
        Can           & 7/10  & Coil         & 9/10  \\
        Toy          & 10/10 & Foam Box Lid & 10/10 \\
        McCafe        & 10/10 & Orange       & 9/10  \\
        Packaging Box & 10/10 & Pill Bottle  & 10/10 \\
        Pomegranate   & 8/10  & Red Tape     & 8/10  \\
        \hline
        \multicolumn{3}{l}{Overall} & \textbf{91/100} \\
        \hline
    \end{tabular}%
    }
\end{table}

\textbf{Results.}
DexGraspFlow achieves a 91\% overall success rate (91 of 100 trials) on unseen physical objects, with perfect 10/10 grasps on five objects (Table~\ref{tab:real_robot_results}, Fig.~\ref{fig:real_robot_grasp_results}). The few failed trials primarily stemmed from slight contact slippage on smooth metallic surfaces (e.g., the can). Overall, these trials demonstrate that the generated link poses can be converted through IK into joint configurations that support stable physical grasping.

\section{Conclusion and limitation}\label{sec:conclusion_limitations}
\subsection{Conclusion}

We presented DexGraspFlow, a cross-embodiment dexterous grasp generation framework that integrates object and hand geometry, link-level poses, and kinematic topology into a shared representation. Grasp generation is formulated through translational flow matching in Euclidean space and rotational flow matching on \(\mathrm{SO}(3)\), with both velocity components jointly predicted by a geometry- and topology-aware dual-stream DiT. This design combines evolving spatial interactions with the hand’s fixed kinematic structure throughout generation. Batched inverse kinematics converts the generated link poses into joint configurations for execution. Crucially, closed-form geodesic steering during flow integration enables training-free, orientation-controllable dexterous grasping, effectively bridging the gap between unconstrained pose synthesis and task-oriented robotic manipulation.

We evaluate DexGraspFlow in three settings: grasp generation on unseen objects across different hand embodiments, zero-shot grasp generation with unseen finger-removal variants of these hands, and real-robot execution using a model trained on grasp demonstrations for the deployed hand. Across 30 held-out object--hand pairs, it achieves a 95.33\% average simulation success rate with an end-to-end latency of 0.126--0.136~s per grasp, approximately 49\% lower on average than that of TRO-Grasp. Under six-direction palm-orientation conditioning, it achieves 86.69\% success with a mean palm-orientation error of 13.92~deg. The framework further achieves 88.74\% success on seven finger-removal variants without retraining. In real-robot experiments, a model trained on Linker Hand L20Lite grasp demonstrations succeeds in 91 of 100 grasping trials on unseen physical objects. Together, these results demonstrate the framework's effectiveness in grasp generation on unseen objects, adaptation to finger-removal topology changes, orientation guidance, and real-robot execution.

\subsection{Limitation}
Despite zero-shot generalization to finger-removal variants, DexGraspFlow still faces challenges in generalizing to completely novel hand embodiments whose morphology differs fundamentally from the training hands. In addition, the flow field generates link poses in $\mathrm{SE}(3)$ without explicitly enforcing joint limits or collision avoidance, leaving physical feasibility to the subsequent IK optimization. Future work will incorporate differentiable kinematic and penetration constraints directly into generative flow integration, expand training hand diversity, and explore contact- and task-oriented conditioning for versatile downstream manipulation.

\clearpage
\newpage
\bibliographystyle{assets/plainnat}
\bibliography{paper}

\end{document}